%% file: main.tex
\pdfoutput=1
\documentclass{article}

\usepackage[preprint]{neurips_data_2022}

\usepackage[utf8]{inputenc} 
\usepackage[T1]{fontenc}    
\usepackage{hyperref}       
\usepackage{url}            
\usepackage{booktabs}       
\usepackage{amsfonts}       
\usepackage{nicefrac}       
\usepackage{microtype}      
\usepackage{xcolor}         
\usepackage{graphicx}
\usepackage{longtable}
\usepackage[export]{adjustbox}
\usepackage{caption}
\usepackage{subcaption}
\usepackage{listings}       
\usepackage{float}

\usepackage{multirow}
\usepackage{multibib}
\newcites{supp}{Supplementary References}
\usepackage{scrextend}

\definecolor{verbgray}{gray}{0.9}
\lstnewenvironment{code}{%
  \lstset{backgroundcolor=\color{verbgray},
  frame=single,
  framesep=5pt,
  framerule=0.5pt,
  basicstyle=\ttfamily,
  columns=fullflexible}}{}
\definecolor{shadecolor}{rgb}{.9, .9, .9}

\usepackage{array}
\newcolumntype{P}[1]{>{\raggedright\arraybackslash}p{#1}}
\newcolumntype{L}[1]{>{\raggedright\let\newline\\\arraybackslash\hspace{0pt}}m{#1}}
\newcolumntype{C}[1]{>{\centering\let\newline\\\arraybackslash\hspace{0pt}}m{#1}}
\newcolumntype{R}[1]{>{\raggedleft\let\newline\\\arraybackslash\hspace{0pt}}m{#1}}

\newcommand{\datasets}{126}
\newcommand{\ours}{\textsc{BigBIO}}

\title{\ours: A Framework for Data-Centric \\ Biomedical Natural Language Processing}

\input{00_authors}
\begin{document}
\maketitle

\input{00_abstract}
\input{01_introduction}

\input{02_related_work}
\input{03_dataset_implementation}

\input{04_dataset_details}

\input{05_use_case_prompting}

\input{06_use_case_mtl}

\input{08_discussion}

\input{09_conclusion}

\begin{ack}
Leon Weber acknowledges the support of the Helmholtz Einstein International Berlin Research School in Data Science (HEIBRiDS). Samuele Garda is supported by the Deutsche Forschungsgemeinschaft as part of the research unit “Beyond the Exome”. 
We are grateful to CoreWeave and EleutherAI for providing the compute needed to evaluate the $6$ and $11$ billion parameter models on our benchmarks, and to Suzana Ilić, Clem Delangue, and others for helping to advertise our calls for participation in the biomedical hackathon. 
Special thanks to the entire BigScience team, including but not limited to Huu Nguyen, Vassilina Nikoulina, Aurélie Névéol, Yong Zheng-Xin, Victor Sanh, and many others, for their thoughtful discussions and contributions in support of the biomedical working group. 

\end{ack}

{
\small
\bibliography{anthology,custom}
\bibliographystyle{plain}
}

\section*{Checklist}

\begin{enumerate}

\item For all authors...
\begin{enumerate}
  \item Do the main claims made in the abstract and introduction accurately reflect the paper's contributions and scope?
    \answerYes{See \S3-6} 
  \item Did you describe the limitations of your work?
    \answerYes{See \S\ref{sec:discussion}}
  \item Did you discuss any potential negative societal impacts of your work?
    \answerYes{See \S \ref{sec:discussion}}
  \item Have you read the ethics review guidelines and ensured that your paper conforms to them?
  \answerYes{}
   
\end{enumerate}

\item If you are including theoretical results...
\begin{enumerate}
  \item Did you state the full set of assumptions of all theoretical results?
    \answerNA{The paper is largely empirical and does not claim new theoretical results}
	\item Did you include complete proofs of all theoretical results?
    \answerNA{The paper is largely empirical and does not claim new theoretical results}
\end{enumerate}

\item If you ran experiments (e.g. for benchmarks)...
\begin{enumerate}
  \item Did you include the code, data, and instructions needed to reproduce the main experimental results (either in the supplemental material or as a URL)?
    \answerYes{See Abstract and Appendix \S \ref{sec:mtlsupp}, \S \ref{sec:zeroshotsupp}}
  \item Did you specify all the training details (e.g., data splits, hyperparameters, how they were chosen)?
    \answerYes{See Appendix \S \ref{sec:mtlsupp}, \S \ref{sec:zeroshotsupp}}
	\item Did you report error bars (e.g., with respect to the random seed after running experiments multiple times)?
    \answerYes{See \S \ref{sec:usecases}. Full details on replicates are in Appendix \S \ref{sec:mtlsupp}, \S \ref{sec:zeroshotsupp}. We refrained from running the MTL experiments over multiple random seeds to save compute budget.}
	\item Did you include the total amount of compute and the type of resources used (e.g., type of GPUs, internal cluster, or cloud provider)?
    \answerYes{See \S \ref{sec:usecases} and the Appendix \S \ref{sec:mtlsupp}, \S \ref{sec:zeroshotsupp}.}
\end{enumerate}

\item If you are using existing assets (e.g., code, data, models) or curating/releasing new assets...
\begin{enumerate}
  \item If your work uses existing assets, did you cite the creators?
    \answerYes{In addition to the assets cited in this paper, \ours~builds on all included datasets. Full metadata, including citations and licensing, for each dataset are available in the data loading scripts that are part of the {\tt bigbio} Python package}
  \item Did you mention the license of the assets?
    \answerYes{See \S \ref{sec:thedataset} and previous answer}
  \item Did you include any new assets either in the supplemental material or as a URL?
    \answerYes{See Abstract and Appendix}
  \item Did you discuss whether and how consent was obtained from people whose data you're using/curating?
    \answerYes{See \S \ref{sec:theframework}}
  \item Did you discuss whether the data you are using/curating contains personally identifiable information or offensive content?
    \answerYes{See \S \ref{sec:curation} and \S \ref{sec:discussion}}
\end{enumerate}

\item If you used crowdsourcing or conducted research with human subjects...
\begin{enumerate}
  \item Did you include the full text of instructions given to participants and screenshots, if applicable?
    \answerYes{See \S \ref{sec:hackathon}}
  \item Did you describe any potential participant risks, with links to Institutional Review Board (IRB) approvals, if applicable?
    \answerNA{The paper did not involve research with human subjects.}
  \item Did you include the estimated hourly wage paid to participants and the total amount spent on participant compensation?
    \answerNo{Crowdsourcing was arranged through non-monetary voluntary participation(Hackathon). While the participants were not compensated financially, we informed participants that their contribution would be acknowledged through authorship in the resulting publication, based on the number of datasets they have contributed. See Appendix \S \ref{appx:authorship} for detailed authors contributions.}
\end{enumerate}

\end{enumerate}


\appendix

%
%
\input{10_appx_author_statement}

\input{11_appx_authorship}

\input{12_appx_schema}
\input{13_appx_metadata}
\input{14_appx_unit_tests}
\input{15_appx_pr_checklist}
\input{16_appx_hackathon}

\input{17_appx_data_dedup}

\input{18_appx_dataviz}

\input{19_appx_ml_zero_shot}

\input{20_appx_ml_mtl}

\input{21_appx_existing_benchmarks}

\input{22_appx_datasheets}
\input{23_appx_bigbio_datasheet}

{
\small
\bibliographystylesupp{plain}
\bibliographysupp{custom}
}

\end{document}

%% file: 00_authors.tex
\author{
Jason Alan Fries$^{{1}*}$ \quad Leon Weber$^{2,3*}$ \quad Natasha Seelam$^{{4}*}$ \quad Gabriel Altay$^{{5}*}$ \\
\textbf{Debajyoti Datta}$^{{6}\dagger}$ \quad \textbf{Ruisi Su}$^{{7}\dagger}$ \quad \textbf{Samuele Garda}$^{{2}\dagger}$ \quad \textbf{Sunny MS Kang}$^{{8}\dagger}$ \\
\textbf{Stella Biderman}$^{9,10\dagger}$ \quad \textbf{Matthias Samwald}$^{{11}\dagger}$ \quad \textbf{Stephen H. Bach}$^{{12}\dagger}$ \quad \textbf{Wojciech Kusa}$^{{13}\dagger}$ \\
\textbf{Samuel Cahyawijaya}$^{{14}\dagger}$ \quad \textbf{Fabio Barth}$^{{2}\dagger}$ \quad \textbf{Simon Ott}$^{{11}\dagger}$ \quad \textbf{Mario Sänger}$^{{2}\dagger}$ \quad \textbf{Bo Wang}$^{{15}}$ \\ 
\textbf{Alison Callahan}$^{{1}}$ \quad \textbf{Daniel León Periñán}$^{{16}}$ \quad \textbf{Théo Gigant}$^{{7}}$ \quad
\textbf{Patrick Haller}$^{{2}}$ \\ 
\textbf{Jenny Chim}$^{{17}}$ \quad \textbf{Jose Posada}$^{{18}}$ \quad \textbf{John Giorgi}$^{{19}}$ \quad \textbf{Karthik Rangasai Sivaraman}$^{{20}}$ \\ 
\textbf{Marc Pàmies}$^{{21}}$ \quad \textbf{Marianna Nezhurina}$^{{22}}$ \quad \textbf{Robert Martin}$^{{2}}$ \quad \textbf{Moritz Freidank}$^{{23}}$ \\
\textbf{Nathan Dahlberg}$^{{7}}$ \quad 
\textbf{Shubhanshu Mishra}$^{{24}}$ \quad \textbf{Shamik Bose}$^{{7}}$ \quad \textbf{Nicholas Broad}$^{{25}}$ \\
\textbf{Yanis Labrak}$^{{26}}$ \quad \textbf{Shlok S Deshmukh}$^{{27}}$ \quad \textbf{Sid Kiblawi}$^{{28}}$ \quad
\textbf{Ayush Singh}$^{{7}}$ \quad \textbf{Minh Chien Vu}$^{{29}}$ \\
\textbf{Trishala Neeraj}$^{{30}}$ \quad \textbf{Jonas Golde}$^{{2}}$ \quad \textbf{Albert Villanova del Moral}$^{{25}}$ \quad \textbf{Benjamin Beilharz}$^{{31}}$ \\
$^{1}$Stanford University \quad $^{2}$Humboldt-Universität zu Berlin \\ 
$^{3}$Max Delbrück Center for Molecular Medicine $^{4}$Sherlock Biosciences \quad $^{5}$Tempus Labs Inc. \\
$^{6}$University of Virginia \quad $^{7}$BigScience \quad $^{8}$Immuneering \quad $^{9}$EleutherAI \quad $^{10}$Booz Allen Hamilton \\ $^{11}$Institute of Artificial Intelligence, Medical University of Vienna \quad $^{12}$Brown University \\ $^{13}$TU Wien \quad $^{14}$The Hong Kong University of Science and Technology \\
$^{15-25}$See Appendix A \quad $^{*}$Equal Contribution \quad $^{\dagger}$Equal Contribution\\
Corresponding Authors: \texttt{jason-fries@stanford.edu} \quad \texttt{leonweber@posteo.de} \\
\texttt{nseelam1@gmail.com} \quad \texttt{gabriel.altay@gmail.com} \\ 
}

%% file: 00_abstract.tex
\begin{abstract}
Training and evaluating language models increasingly requires the construction of \emph{meta-datasets} -- diverse collections of curated data with clear provenance.
Natural language prompting has recently lead to improved zero-shot generalization by transforming existing, supervised datasets into a diversity of novel pretraining tasks, highlighting the benefits of meta-dataset curation.  
While successful in general-domain text, translating these data-centric approaches to biomedical language modeling remains challenging, as labeled biomedical datasets are significantly underrepresented in popular data hubs.  
To address this challenge, we introduce \ours~a community library of \datasets+ biomedical NLP datasets, currently covering 12 task categories and 10+ languages.
\ours~facilitates reproducible meta-dataset curation via programmatic access to datasets and their metadata, and is compatible with current platforms for prompt engineering and end-to-end few/zero shot language model evaluation.
We discuss our process for task schema harmonization, data auditing, contribution guidelines, and outline two illustrative use cases: zero-shot evaluation of biomedical prompts and large-scale, multi-task learning. 
\ours~is an ongoing community effort and is available at \href{https://github.com/bigscience-workshop/biomedical}{this URL}. 
\end{abstract}

%% file: 01_introduction.tex
\section{Introduction}


%
%
Large-scale language modeling has demonstrated exciting performance gains in zero-shot classification when combined with explicit, prompted supervision.
Here, existing labeled datasets are transformed into prompted training examples, which redefine classification tasks as generative, text completion tasks~\cite{raffel2019exploring}.
T0 and FLAN have demonstrated improvements in zero-shot generalization using this training approach \cite{sanh2022multitask,wei2022finetuned}.
Increasing the number of prompted training tasks can also lead to improved generalization even when the number of model parameters is fixed. 

%
%

The importance of carefully controlling the tasks a language model is exposed to during training highlights how \textit{meta-dataset} curation is critical for state-of-the-art language modeling.
Prompting offers new opportunities for constructing meta-datasets and aligns with the principles of data-centric machine learning, which focuses on training data curation to improve model performance.
In the general NLP domain, data-centric methods have benefited from community efforts such as Hugging Face’s datasets hub~\cite{lhoest-etal-2021-datasets}, which provides easy, programmatic access to datasets and their attributes.
However, biomedical datasets are significantly underrepresented in the datasets hub ~\cite{fries-etal-2022-dataset} creating challenges in reproducibly accessing, curating, and remixing biomedical NLP data for prompted training and zero/few-shot evaluation of language models. 

To help address these challenges, we introduce \ours, a community resource for programmatically accessing biomedical NLP datasets at scale and encouraging reproducibly when generating meta-datasets.
\ours~is, to the best of our knowledge, the largest public collection of curated and unit-tested biomedical NLP datasets. 
\ours~was developed as part of BigScience\footnote{https://bigscience.huggingface.co/}, a year-long workshop on large language modeling, and codifies many lessons of the biomedical working group as they developed dataset curation strategies.

A summary of our contributions:  
\begin{itemize}
    \item Programmatic access to \datasets+ unit-tested, biomedical datasets, covering 12 tasks, 10+ languages, and providing structured metadata for key attributes on provenance and licensing.
    \item Support for multiple lightweight schemata, which preserve the dataset as released and provide harmonized access for prompt engineering and cross-dataset integration.
    \item Community tools and guides for contributing new datasets.
    \item \ours~ is built upon Hugging Face’s datasets library, integrating with PromptSource~\cite{bach2022promptsource}, a prompt engineering system and repository, and the EleutherAI Language Model Evaluation Harness \cite{eval-harness} to support rapidly designing and evaluating prompts on biomedical tasks. 
\end{itemize}

We illustrate the utility of \ours~in two representative use cases: (1) zero-shot, prompted biomedical language model evaluation; and (2) large-scale multi-task learning (MTL) with 100+ tasks. 
In both use cases, we substantially lower the engineering costs required to construct the meta-datasets commonly utilized for language modeling and other machine learning applications.

%% file: 02_related_work.tex
\section{Related Work}

\ours~is a data-centric approach to natural language processing in the biomedical domain.
We briefly overview related work in these two areas.

\subsection{Data-Centric Machine Learning}

\emph{Data-centric machine learning} emphasizes the thoughtful curation of data as centrally important to the development of models.
Multiple arguments for this emphasis have been advanced.
Paullada et al.~\cite{paullada2021data} survey many aspects, including mitigating biases and annotation artifacts in training data that lead models to rely on spurious correlations that do not generalize to other datasets, and addressing representational harms in which certain people are under, over, or misrepresented.
Sambasivan et al.~\cite{sambasivan2021everyone} document prevalent ``data cascades,'' situations in AI and machine learning practice in which low-quality data causes downstream problems in high-stakes applications.
Biderman and Scheirer~\cite{pmlr-v137-biderman20a} make several recommendations for improved data practices, including auditing and documenting datasets.
Rogers~\cite{rogers-2021-changing} outlines issues with models that can be exacerbated by low-quality data. This encompasses for instance: learning spurious patterns, being vulnerable to basic input perturbations, and struggling with rare inputs.
\ours~is motivated by these same arguments, hence its emphasis on careful metadata curation and harmonized task schemata.

Data quality has a large impact on model performance.
Deduplicating data leads to more accurate and more robust models with faster convergence. \cite{cohen2013redundancy, lee2021deduplicating}.
For instance, cleaning up the consistency of answer response strings was reported to improve biomedical question answering \cite{yoon2021ku}.
Duplication contamination is a serious risk in biomedical datasets, which often iteratively build or extend prior annotations, introducing risk of test leakage in evaluation \cite{elangovan-etal-2021-memorization}.
As we describe in \S\ref{sec:theframework},
\ours's centralization of data in a unified format enables systematic data quality checks.

Data governance is also an important issue when curating biomedical language data.
Jernite et al.~\cite{jernite:faact22} survey many aspects of the governance of language data, and propose a framework for distributed governance of large language corpora.
Vayena et al.~\cite{vayena2017biomedical} describe models of data governance that enable biomedical research while respecting patient privacy.
Jones et al.~\cite{jones2020toward} propose data governance standards for clinical text data with personally identifiable information.
Some of these issues are not directly applicable to \ours, which currently only includes loaders for datasets that are compliant with the United States Health Insurance Portability and Accountability Act (HIPAA) as public research datasets.
Further, \ours~is not itself a repository of data, but a centralized repository of data loaders and metadata, meaning that future dataset creators can programmatically define how a dataset should be accessed and share this information with the community.

\subsection{Biomedical Benchmarks}
Task-specific benchmark datasets are common in biomedical workshops like BioNLP and BioCreative~\cite{kim2009overview,hirschman2005overview}. These datasets however typically assess a restricted set of skills learned by a model.
Several recent efforts have focused on curating larger collections of datasets and tasks to evaluate the performance of biomedical NLP models.
BLUE (Biomedical Language Understanding Evaluation) is a benchmark for 10 datasets representing 5 tasks \cite{peng-etal-2019-transfer}, which was extended by BLURB (Biomedical Language Understanding and Reasoning Benchmark) to include 13 datasets and 7 tasks \cite{DBLP:journals/health/GuTCLULNGP22}. HunFlair provides harmonized access to 23 NER datasets, but imposes assumptions on preprocessing choices (e.g., tokenization) \cite{weber2021hunflair}. Most benchmarks provide no multilingual data.
CBLUE is the only non-English benchmark consisting of 8 datasets and tasks for Chinese biomedical language \cite{zhang-etal-2022-cblue}.

Multiple biomedical prompt datasets have been released for few and zero-shot classification evaluation.
\textsc{Natural-Instructions$_{v2}$} provides 1600+ task instructions for a variety of domains, including ~30 tasks for medicine and healthcare~\cite{wang2022benchmarking}.
BoX provides natural language instructions for 32 datasets and 9 tasks, where instructions consist of an explanation, a prompt, and a collection of example input/outputs~\cite{parmar2022boxbart}.
Agrawal et al.~\cite{agrawal2022large} released 2 datasets for zero-shot clinical information extraction.

\ours~differs from previous efforts by focusing on the infrastructure and curation required to reproducibly generate meta-datasets.
Existing benchmarks provide consistent mechanisms for evaluating machine learning performance, however they do not support consistent tooling to access and ingest data into machine learning workflows.
This is a serious limitation in practice, especially as novel training and evaluation strategies increasingly require transforming input data.
We emphasize direct, easy and programmatic access to datasets with community curation to build open tools for data loading.
We have curated detailed metadata about tasks, e.g. languages, licensing and other aspects of dataset provenance.
We provide harmonized views of datasets by task schema, enabling easier integration into workflows, while also imposing minimal assumptions on NLP preprocessing decisions like sentence splitting and tokenization.
Existing benchmarks typically fix preprocessing choices, creating challenges when comparing end-to-end workflows common in prompting.



%% file: 03_dataset_implementation.tex
\section{The \ours~Framework}
\label{sec:theframework}

This research effort was initiated as part of BigScience, a year-long collaborative workshop on the creation of very large language models, comprised of over 1000 researchers from 60 countries and dozens of working groups.
The BigScience biomedical working group consisted of machine learning researchers and other stakeholders interested in the curation of biomedical data for large-scale language modeling.
\ours~ reflects the lessons and best practices we learned while developing a framework for more easily and reproducibly generating biomedical NLP meta-datasets.

\begin{figure}[h!]
\centering
\includegraphics[width=\linewidth]{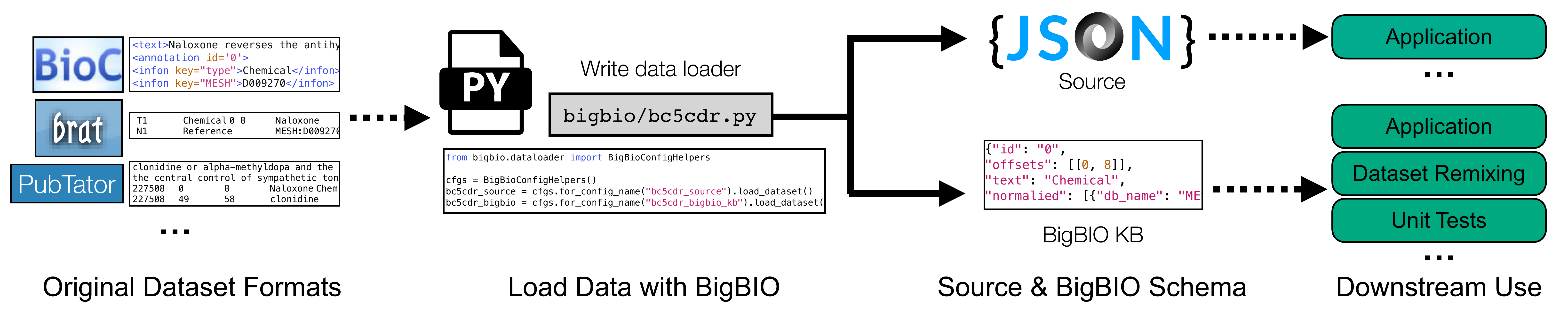}
\caption{The workflow for implementing, harmonizing, and unit testing datasets for inclusion in \ours. Harmonized schemata enable standardizing unit tests, cross-dataset integration, and easier dataset remixing, such as transforming supervised datasets into prompted tasks. }	
\end{figure}

\subsection{Dataset Curation}
\label{sec:curation}
\paragraph{Building the Dataset Catalog}
Our initial efforts in the BigScience working group produced a catalog of important biomedical datasets, key metadata, and other provenance \cite{fries-etal-2022-dataset}.
Selection criteria followed several principles: (1) relevance to biomedical research, (2) diversity of domains, tasks, and languages;  and (3) public availability. 
We used this open catalog as the starting point for \ours~.

\paragraph{Task Schema Harmonization} 
In biomedical NLP there are a proliferation of data formats (e.g., BioC, BRAT) but inconsistent adherence across those formats. 
Developing common data models for interoperability~\cite{bleiholder2009data}, while beneficial for cross-dataset integration, risks possible information loss when translating or \textit{harmonizing} information across schemata.
To develop shared infrastructure for data ingestion and minimize information loss, we designed data loaders to support 2 dataset views: (1) a source schema that preserves the original dataset format as faithfully as possible; and (2) task-specific, harmonized \ours~schema. 
We developed 6 lightweight schema supporting common NLP tasks including knowledge base construction (KB), question answering (QA), textual entailment (ENTAIL), text to text (T2T), textual pairs (PAIRS), and document/text classification (TEXT). 
Complete specifications are in the Appendix. 

\paragraph{Unit Tests and Dataset Cleaning}
To safeguard correctness of data loader implementations, we developed a testing suite of unit-tests for monitored quality issues. 
\ours~schema are designed to support key dataset integrity checks, such as enforcing unique IDs across elements, relational consistency, confirming text offsets are correctly aligned within document text, etc.
The unit testing suite is runnable as part of the dataset submission process, providing feedback on diagnosing implementation or dataset errors.
Where possible, we implemented tools for common data cleaning tasks, such as normalizing PubMed IDs (PMIDs). 

\paragraph{Acceptance Checklist} 
Submissions to \ours~ require completing a checklist of inclusion criteria before acceptance into the project GitHub repository.
First, correctly annotating all metadata relevant to the dataset (e.g., languages, task types, provenance).
Second appropriate schema and task pairing, and consistent materialization of data across all data subsets defined by the dataset authors.
Finally, submissions must demonstrate that code passes all unit-tests. 

All publicly accessible scripts were manually reviewed and accepted by a \ours~ admin. Local datasets that require a manual download of the data were also manually checked if an admin had appropriate authorization (e.g., several authors have PhysioNet credentials).
In absence of dataset access, data loaders were accepted contingent on showing the output of successful unit test logs.

\paragraph{Issue Tracking} Most biomedical datasets involve complex labeling tasks, so even in cases when datasets pass unit tests they may contain subtle bugs or misunderstandings that require revisiting. 
To identify and harden our dataset implementations, we implemented the 2 use cases outlined in \S\ref{sec:usecases}: zero-shot language model evaluation and  large-scale multi-task-learning. 
Implementing these realistic machine learning workflows resulted in identifying  non-obvious dataset-specific errors or limitations in our current schema.
For example, some datasets do not provide natural language class labels, such as labeling a relation with an internal code (CPR:6) instead language describing the underlying biological relationship (ANTAGONIST), which creates challenges when writing prompts.

\subsection{Prompting and Language Model Evaluation Harness}

To demonstrate the accessibility of the \ours~library, we integrated this package with several other frameworks as a proof of concept.
First, we integrated with PromptSource~\cite{bach2022promptsource} to enable the creation of prompted representations of the data.
PromptSource is a development environment for prompts, which requires datasets to be available for loading in a unified format.
All of \ours's datasets can be loaded into PromptSource, and then users can write prompts for them and materialize the prompted forms of those datasets locally for training and evaluation.

To further enable the evaluation of language models on datasets in \ours, we also connected \ours~with the EleutherAI Language Model Evaluation Harness \cite{eval-harness}.
The Evaluation Harness handles the loading, querying, and scoring of language models, with programmatic definitions of how evaluations are carried out.
Here, the unified task schema of \ours~are an advantage, enabling standard evaluation schemes to be automatically applied to a wide collection of datasets, while still allowing for additional definitions of specialized evaluations.

\subsection{Biomedical Hackathon}
\label{sec:hackathon}
After internally testing the elements outlined in \S\ref{sec:curation}, we drafted   instructional material and code tutorials for external collaborators. 
We then launched an international call for participation\footnote{https://hfbigbio.github.io/} in a biomedical hackathon to implement all 174 datasets in the \ours~catalog. 
Participants were recruited through Twitter.
We established formal participation guidelines and corresponding credit, including co-authorship on this manuscript, given implementation of 3 or more data loaders.
The hackathon officially ran for 2 weeks with an unofficial 2 week wrap-up period.
During the official period, we held daily office hours to help participants, running a Discord server to facilitate rapid communication and up-to-date FAQ.  
At the conclusion of the hackathon, 48 participants had implemented \datasets~total datasets with an additional 18 dataset still undergoing quality control.

%% file: 04_dataset_details.tex
\section{The \ours~Dataset}
\label{sec:thedataset}

\begin{figure}[ht!]
  \centering
  \includegraphics[scale=0.21]{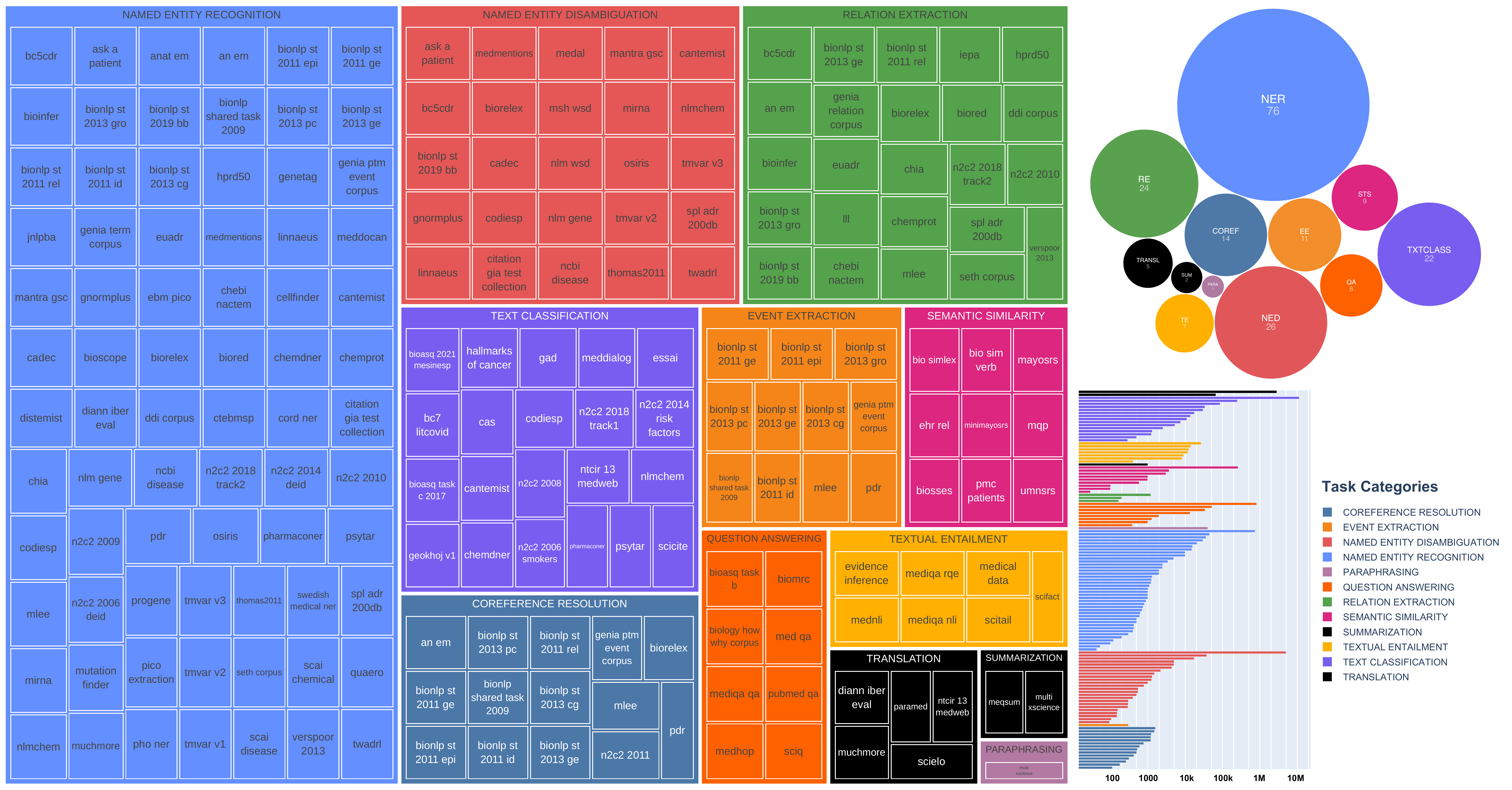}
  \caption{\label{fig:treemap-all-datasets}Treemap visualization of \ours's \datasets~datasets and 12 task categories, denoted by color (left); the distribution of dataset sizes measured by number of examples (bottom right); and a circle plot of task categories and their relative size (top right). }
\end{figure}

\begin{table}[htbp]
  \centering
  \caption{Summary statistics for \ours. Note datasets may contain multiple schema.}
    \begin{tabular}{rcccccc|c}
    \toprule
        &                   KB & TEXT & PAIRS & QA & ENTAIL & T2T & ALL \\
    \midrule
    Datasets               & 84   & 21   & 10   & 8    & 7    & 7   & 126 \\
    Public Datasets        & 73   & 9    & 10   & 7    & 4    & 6   & 105 \\
    Private Datasets       & 11   & 12   & 0    & 1    & 3    & 1   & 21  \\
    PubMed Datasets        & 64   & 7    & 3    & 4    & 1    & 1   & 77 \\
    Languages              & 7    & 4    & 1    & 1    & 1    & 4   & 10  \\ 
    Tasks                  & 5    & 1    & 1    & 1    & 1    & 3   & 12  \\

    \bottomrule
    \end{tabular}%
  \label{tab:bb-summary-tall-schema}%
\end{table}%

We provide a {\tt bigbio} Python package that supports streamlined loading of \datasets~biomedical datasets covering 12 tasks grouped into 6 schema types for a total of 24 million examples comprising 18 trillion characters. 
To the best of our knowledge, \ours~ is the largest single collection of curated and unit tested biomedical NLP datasets. 
Figure \ref{fig:treemap-all-datasets} visualizes the datasets and tasks in \ours~and Table 
\ref{tab:bb-summary-tall-schema} provides dataset counts by schema and key attributes. 
The publicly available datasets (105 of \datasets~datasets) can be automatically downloaded. 
We provide scripts to load the remaining 21 datasets that require further access approvals, where the user only needs to specify a path to their local copy of the datasets.
This restriction is common in clinical datasets, which require credentialing and training on how to handle protected health information.
    


\paragraph{Metadata Summary} Overall 10 languages are represented, with English being the majority (83\%) followed by Spanish (6.5\%), French (2.9\%), Chinese (2.2\%), and German (1.4\%).  Japanese, Dutch, Portuguese, Swedish, and Vietnamese are each present in one dataset. Creative Commons licenses are used more frequently than any other type covering 44 (35\%) of datasets with 8 (6.3\%) using the non commercial use (NC) option. The next most frequent type is an unknown license for 34 (27\%) of datasets. These are cases in which the dataset authors did not choose a license or one could not be located for the dataset.
The remaining licenses are a mixture of permissive open source licenses such as MIT and Apache and more restrictive licensing requiring written applications for use and custom data user agreements.
A complete list of structured metadata is available in Appendix \S\ref{sec:metadata}.

%% file: 05_use_case_prompting.tex
\section{Use Cases}
\label{sec:usecases}

We develop two downstream use cases of \ours, to showcase the utility of the library and identify any workflow issues.
In the first use case, we evaluate prompted language models in a zero-shot setting and in the second we train a large-scale MTL model.
Both use cases used a single 8x A40 compute node and MTL also used a 4x RTX 3090 node.
Expanded results and experimental details are available in Appendix \S\ref{sec:zeroshotsupp} (zero-shot evaluation) and \S\ref{sec:mtlsupp} (MTL).

\subsection{Zero-shot Evaluation of Prompted Language Models}
\label{sec:zseval}

\begin{figure}[ht!]
  \centering
  \includegraphics[scale=0.48]{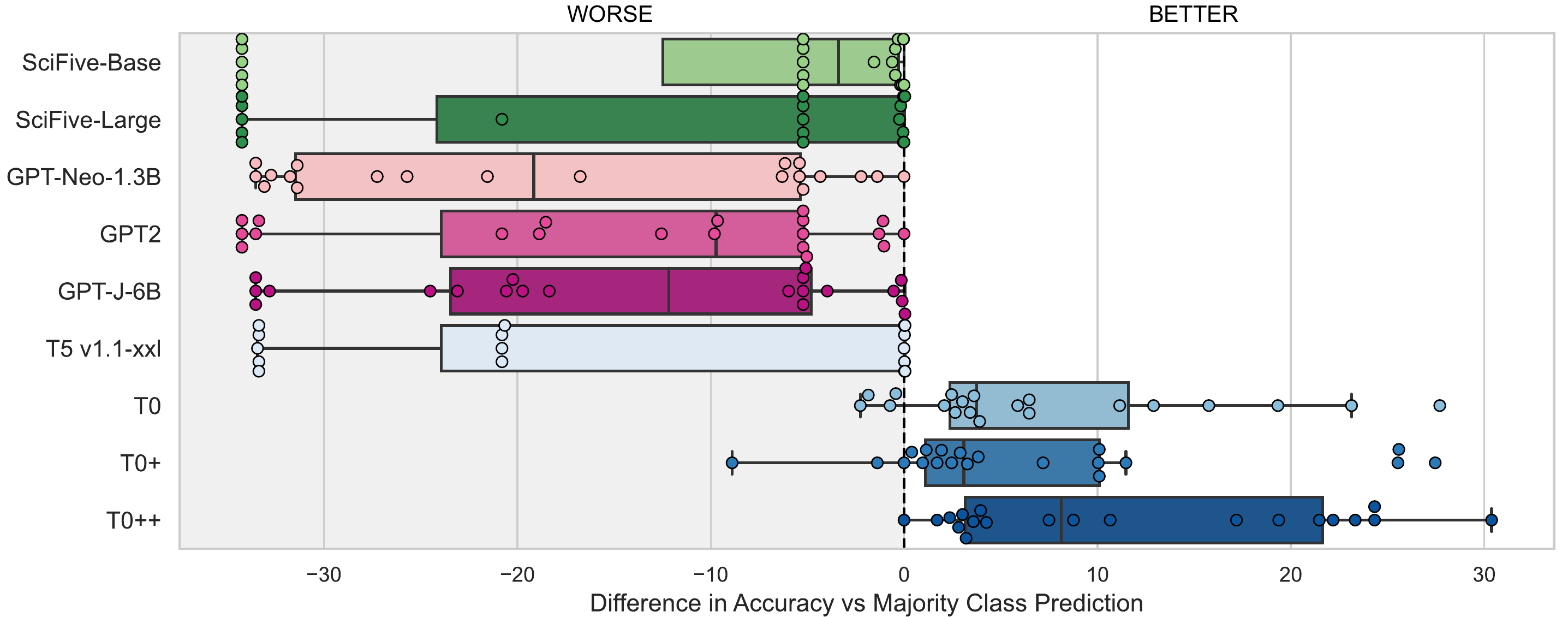}
  \caption{\label{fig:zsbaseline} Zero-shot generalization to biomedical tasks. Box plots show pooled accuracy differences between a majority class baseline and zero-shot prediction for all datasets excluding BIOSSES. Points are per-prompt scores. T0 is the only language model class to outperform the majority baseline.}
\end{figure}

\paragraph{Datasets and Prompts}
We selected 5 representative datasets from \ours: BIOSSES (semantic textual similarity), BioASQ (yes/no question answering), GAD (relation extraction), SciTail (textual entailment), and MedNLI (clinical textual entailment).
We exclude NER datasets due to challenges and computational costs of using discrete prompting for token classification tasks \cite{ma2021template}.
For each dataset, we wrote 5 prompts using PromptSource to reflect the original classification task. 

\paragraph{Evaluation Protocol} 
We evaluate 10 pretrained language models, ranging from 220 million to 11 billion parameters:
SciFive-base/large \cite{phan2021scifive}, 
GPT Neo-1.3B\cite{gpt-neo}, GPT-2\cite{radford2019language}, GPT-J-6B \cite{gpt-j}, the T0 family \cite{sanh2022multitask}, and the 11B parameter base T5 model used to build T0 \cite{raffel2019exploring}.
Models were evaluated using a BigScience prompted evaluation library\footnote{https://github.com/bigscience-workshop/lm-evaluation-harness} built on top of the language model evaluation harness from Gao et al. \cite{eval-harness}. 
All evaluations use the canonical test split where possible, otherwise we used BLURB's test set definitions.
All tasks are evaluated using accuracy except BIOSSES which uses Pearson's correlation after transforming outputs into numbers.
We evaluate all prompts and report the average and best performance for each dataset, as well as a baseline score based on the majority class. 
For contextualizing scores, we include prior state-the-art finetuned performance for all tasks \cite{tinn2021fine,yasunaga2022linkbert,phan2021scifive}. 

%
%
\begin{table}[ht!]
\setlength{\tabcolsep}{5.5pt}
\caption{\label{tbl:zs} Zero-shot performance of prompted language models}
\begin{tabular}{rcccllllllll}
\toprule
 & & \multicolumn{2}{c}{BIOSSES} & \multicolumn{2}{c}{BioASQ} & \multicolumn{2}{c}{SciTail} & \multicolumn{2}{c}{MedNLI} & \multicolumn{2}{c}{GAD}   \\
\cmidrule(r){3-12}
Model & PMC & Avg & Best &  Avg & Best &  Avg & Best &  Avg & Best &  Avg & Best \\
\midrule
SciFive-Base & \checkmark & 34.0 & 55.8 & 32.9 & 32.9 & 59.9 & 60.4 & 66.4 & 66.7 & 47.4 & 47.4 \\
SciFive-Large & \checkmark &  7.2 & 19.5 & 32.9 & 32.9 & 56.2 & 60.4 & 66.7 & 66.7 & 47.4 & 47.4 \\ \midrule
 GPT-Neo-1.3B & \checkmark & 36.4 & 36.4 & 40.9 & 65.7 & 50.6 & 60.4 & 36.6 & 41.0 & 47.7 & 50.4 \\
GPT-2         & & 12.5 & 19.5 & 36.1 & 48.6 & 50.3 & 60.4 & 55.1 & 65.6 & 47.4 & 47.6 \\
GPT-J-6B      & \checkmark &  0.2 & 32.1 & 40.4 & 67.1 & 51.6 & 60.3 & 48.3 & 62.7 & 48.2 & 52.1 \\ \midrule
T5 v1.1-xxl   & &  - &  - & 67.1 & 67.1 & 43.8 & 60.4 & 33.3 & 33.3 & 52.6 & 52.6 \\ 
T0            & & 23.3 & 49.5 & 76.1 & 82.9 & 73.9 & 88.1 & 72.0 & \textbf{77.8} & 53.7 & 55.6 \\
T0+           & & 37.8 & \textbf{66.7} & 73.1 & 78.6 & 74.3 & 87.9 & 72.5 & 76.8 & 53.9 & 55.1 \\
T0++.         & & \textbf{40.6} & 42.5 & \textbf{89.0} & \textbf{91.4} & \textbf{75.6} & \textbf{90.8} & \textbf{73.4} & 77.4 & \textbf{55.7} & \textbf{56.6} \\ \midrule
  Majority Class &  & \multicolumn{2}{c}{-} & \multicolumn{2}{c}{67.1} & \multicolumn{2}{c}{60.4} & \multicolumn{2}{c}{66.7} & \multicolumn{2}{c}{52.6} \\
  Finetuned SOTA & & \multicolumn{2}{c}{94.5} & \multicolumn{2}{c}{94.8} & \multicolumn{2}{c}{96.8} & \multicolumn{2}{c}{86.6} & \multicolumn{2}{c}{84.9} \\
\bottomrule
\end{tabular}
\end{table}


\paragraph{Results} 
Fig. \ref{fig:zsbaseline} shows that T5 and GPT models fail to generalize to biomedical text, regardless of parameter count or exposure to biomedical text during pretraining/finetuning. 
T0 class models do demonstrate task generalization, even though those models were not exposed to any biomedical tasks during prompted pretraining.
We replicate the finding in Sanh et al. that models using more prompted pretraining tasks demonstrate better generalization, finding that T0++ performed best overall.
Table \ref{tbl:zs} includes performance statistics for all language models and datasets and denotes if the model was trained or finetuned on biomedical data from PubMed Central (PMC).
All non-T0 perform worse than the simple majority class baseline.
For SciFive and T5 models, predictions were often pathological, i.e., emitting the same answer for all prompts.
For the T0 family, models consistently outperformed the majority class baseline.
On BioASQ and SciTail using T0++, the best prompts performed very well, falling 3.4 and 6.0 points short of state of the state-of-the-art supervised models.
MedNLI, GAD, and BIOSSES remained significantly challenging for all models.



%% file: 06_use_case_mtl.tex
\subsection{Large-scale Multi-task Learning}
\paragraph{Data Materialization} We train and evaluate a multi-task learning (MTL) model on 106 different BioNLP tasks using the MaChAmp MTL framework~\citep{van-der-goot-etal-2021-massive}. 
We generated training and evaluation splits using all datasets that were available in the \ours~repository version when we started the project.
From the 106 datasets, we filtered out datasets that: were non-English; had known implementation bugs; included silver-standard annotations; or were document-level or multilabel classification datasets.
For the 67 remaining datasets, we extracted data for 8 task types: Named Entity Recognition, Text Classification, Question Answering, Coreference Resolution, Event Detection, Event Argument Extraction, Relation Extraction and Semantic Textual Similarity, yielding 107 tasks (dataset/task type combinations) in total.
%

%
%
%

%
\paragraph{Training Protocol} We train a single encoder-only transformer model with a separate classification head for each of the 107 tasks.
We initialize the encoder with BioLinkBERT-base~\citep{yasunaga2022linkbert}.
We follow \cite{aghajanyan-etal-2021-muppet} in using a task-heterogeneous batching strategy.
Specifically, at each training step, we sample 32 different tasks and select 16 examples for each of them leading to a total batch size of 512.
We train the model to convergence, which takes less than 50 epochs and then select the best performing checkpoint based on validation performance.
%
%

%
\paragraph{Evaluation Protocol} We evaluate our model on a subset of dataset from the BLURB benchmark.
We select all four datasets that are contained in our MTL training data and have the same splits in the MTL data as in BLURB.
For all datasets, we use the version in the MaChAmp format, which differ in tokenization, sentence splitting and label space from the official BLURB versions.
After prediction, we postprocess the results to match the BLURB label space.
While this introduces confounders that makes direct comparison complicated, e.g., different choices in sentence splitting and tokenization, we include prior state-of-the-art results for the same model size~\citep{yasunaga2022linkbert} as a point of orientation.
We additionally compare with a version of our MTL model that we fine-tune on the training data of the evaluation dataset using the MaChAmp default hyperparameters. 
%
%

%
\paragraph{Results} MTL results are reported in  Table~\ref{tab:mtl-results}.
MTL+Finetuning results are reported as the mean and standard deviation of 3 different random seeds.
For contextualizing scores, we also include state-of-the-art LinkBERT-base results.
The MTL model performs markedly worse than the state-of-the-art LinkBERT model, with differences between 1.5 and 11.2 percentage points (pp) F1.
However, additional fine-tuning only on the evaluation dataset narrows the gap between LinkBERT and the MTL model significantly with a maximum difference of 3.2 pp F1.
This confirms the results of \cite{aghajanyan-etal-2021-muppet} that models trained in a large-scale MTL setting are a suitable basis for further fine-tuning.
However, the failure of the fine-tuned model to perform better than state-of-the-art indicates that more research on the conditions in which large-scale MTL pre-finetuning may improves results is required.



\begin{table}[ht!]
\centering
\caption{F1 scores of the MTL model evaluation}
\begin{tabular}{lcccc}
\toprule
Dataset         & Task                & MTL   & MTL+Finetuning & LinkBERT-base \\
\midrule
NCBI-Disease    & NER                 & 80.2 & 87.5 $\pm$ 0.9           & *88.2         \\
BC5CDR-Disease  & NER                 & 78.5 & 84.8 $\pm$ 0.3          & *86.1          \\
BC5CDR-Chemical & NER                 & 92.2 & 94.4 $\pm$ 0.3         & *93.8         \\
ChemProt        & RE                  & 66.4 & 74.3 $\pm$ 0.1          & *77.6     \\
\bottomrule
\multicolumn{5}{p{13cm}}{\footnotesize{* indicates that comparing results is complicated by different preprocessing choices across benchmarks.}} \\
\end{tabular}
\label{tab:mtl-results}
\end{table}

%% file: 08_discussion.tex
\section{Discussion}
\label{sec:discussion}


The focus of \ours~on providing a unified view over a large number of diverse NLP datasets has a number of benefits.
First, it could increase the robustness of data-centric machine learning because it allows end-to-end data generation workflows that trace data provenance and codify assumptions on data transformations, such as checking for duplicates.
Second, the unified view allows to programatically assure quality of both the source data and the transformed datasets, as exemplified by our suite of unit tests.
Finally, it drastically reduces the amount of work required for training or evaluating models on a large number of tasks, as can be seen in the MTL usecase, where we had to write only 8 data transformation scripts (one for each task type) as opposed to up to 67 (one for each dataset).
Crucially, \ours~achieves this without making strong assumptions about the downstream use case or type of model, e.g. by unifying tasks directly into a conditional text generation/prompting setting.
We believe that our work provides useful suggestions on how to write data loaders for a large number of datasets in a collaborative setting.
We found a uniform view of the datasets useful for quality assurance during implementation, because it allowed to have a uniform suite of unitests, identify common parsing and transformation components that were moved into a helper library and could be heavily tested. 
Furthermore, the categorization of datasets into schemas allowed code reviewers to specialize in a subset of schemas, which likely improved the quality of code reviews.
Finally, we found using \ours~in illustrative downstream use cases during library development immensely helpful, because this informed design decisions for the library such as a the need for a unified interface for filtering and loading a large number of datasets with a few lines of code. 
We also found a significant number of bugs in accepted data loaders when implementing the use cases, for instance because performance was much lower/higher than expected for certain datasets.
Our work has several limitations.
First, some data loaders likely contain implementation errors that were missed by our code review and unit tests. 
Second, our choice of schema makes assumptions on what structures are most useful for biomedical NLP research and thus will not represent all interesting tasks.
Third, \ours~reflects biases that are present in the included data sets, for instance a very strong focus on English text as only 23 of the 126 currently implemented datasets are in a language other than English.
We believe that these limiations will be mitigated over time as researchers continue to use and improve on the datasets and tooling. 






%% file: 09_conclusion.tex
\section{Conclusion and Future Work}
\label{sec:conclusion}

We introduce \ours~ a community library of 126+ biomedical NLP datasets currently covering 12 task categories and 10+ languages. 
\ours~enables reproducible data-centric machine learning workflows, by focusing on programmatic access to datasets and their metadata in a uniform format.
We discussed our process for task schema harmonization, data auditing, contribution guidelines and describe two illustrative use cases of~\ours: zero-shot evaluation of large language models for biomedical prompting and large-scale MTL.
We believe \ours~poses little-to-no negative societal impacts, as all datasets we support are public or governed by HIPAA protections as appropriate.
A chief motivation of this work is the belief that codifying dataset curation choices in code, tracking provenance of meta-dataset curation, and other decisions around transparent training set generation are critical to the ethical application of machine learning.
In the worst case, \ours~might amplify negative impacts already inherent to included datasets as it facilitates dataset access.
For future work, we plan to curate a library of prompted representations of \ours~tasks, including queries formulated like those used to train T0, as well as longer, self-contained instruction sets for novel biomedical tasks.  
Constructing such a library requires a framework for reproducible data ingestion which is provided by \ours.
%






%% file: 10_appx_author_statement.tex
\clearpage
\section{Appendix Overview}

This section summarizes the elements required by NeurIPS for inclusion in supplementary materials.

\begin{enumerate}
\item \textbf{Dataset documentation and intended uses. Recommended documentation frameworks include datasheets for datasets, dataset nutrition labels, data statements for NLP, and accountability frameworks.} We have provided datasheets for all datasets (see \S\ref{sec:datasheets}) in \ours~ as well as a datasheet for the meta-dataset itself (see \S\ref{sec:bigbiodatasheet}). The intended use of \ours~is to enable research on (biomedical) Natural Language Processing. Any usage for direct diagnostic use or medical decision making without review and supervision by medical professionals is out of scope. 
\item \textbf{URL to website/platform where the dataset/benchmark can be viewed and downloaded by the reviewers.} All code required to download datasets and run machine learning experiments outlined in this manuscript is available on the \ours~ GitHub code repository  \url{https://github.com/bigscience-workshop/biomedical}. We are in the process of creating a website that summarizes the aims and contributions of \ours.
\item \textbf{Author statement that they bear all responsibility in case of violation of rights, etc., and confirmation of the data license.} The authors of this manuscript bear all responsibility for any violation of rights caused by the development and release of \ours. All code for \ours~is released under Apache License 2.0. All dataset licensing remains the same as the source.
\item \textbf{Hosting, licensing, and maintenance plan. The choice of hosting platform is yours, as long as you ensure access to the data (possibly through a curated interface) and will provide the necessary maintenance.} All code is hosted on GitHub at the repository linked above. We have released all dataset-related software under an Apache License 2.0. \ours~is an active open source project that is maintained by an international community of volunteers and 4+ code administrators associated with the BigScience biomedical working group. See \S\ref{sec:units} and \S\ref{appx:authorship} for protocols for new dataset contributions and unit testing to ensure ongoing quality checks. Datasets are hosted by their original owners. In cases where the original license permits redistribution, we will mirror dataset releases on our community hub \url{https://huggingface.co/bigscience-biomedical}.
\item \textbf{Links to access the dataset and its metadata.} See our project GitHub for all dataset code and metadata. 
\item \textbf{The dataset itself should ideally use an open and widely used data format. Provide a detailed explanation on how the dataset can be read. For simulation environments, use existing frameworks or explain how they can be used.} \ours~is implemented using Hugging Face's datasets library to support easy integration into existing machine learning workflows. See \S\ref{sec:schema} for details on standardized schema to permit easier reuse. 
\item \textbf{Long-term preservation} For the subset of public datasets that can be redistributed, we intend to create regular snapshots on \ours~ on a data archiving website such as \url{https://zenodo.org/}. 
\item \textbf{Explicit license} All code for \ours~is released under Apache License 2.0. All dataset licensing remains the same as the source. See \S\ref{sec:metadata} and \S\ref{sec:bigbiodatasheet} for complete licensing information for all datasets in \ours.
\item \textbf{For benchmarks, the supplementary materials must ensure that all results are easily reproducible.} All machine learning experiments include instructions and code for reproducing results. See \S\ref{sec:zeroshotsupp} for zero-shot biomedical benchmarking and \S\ref{sec:mtlsupp} for multi-task learning experiments.
\end{enumerate}

%% file: 11_appx_authorship.tex
\clearpage
\section{Author Contributions}
\label{appx:authorship}

The core idea behind this manuscript emerged from discussions in the BigScience biomedical working group. 
We formalized the following criteria for determining authorship.
Joint first authorship required significant intellectual contribution shaping this project, including organization, contributing/reviewing code, writing documentation, and writing this manuscript.
Co-authorship required 3+ submitted dataset implementations that passed all unit tests and other quality control measures.
Co-second authorship required one or more significant contributions to the project beyond participation in the hackathon.

We also thank Giyaseddin Bayrak, Gully Burns, Antonio Miranda-Escalada, Abhinav Ramesh Kashyap and Tanmay Laud for their dataset contributions.

\begin{figure}[ht!]
  \centering
  \includegraphics[width=14cm]{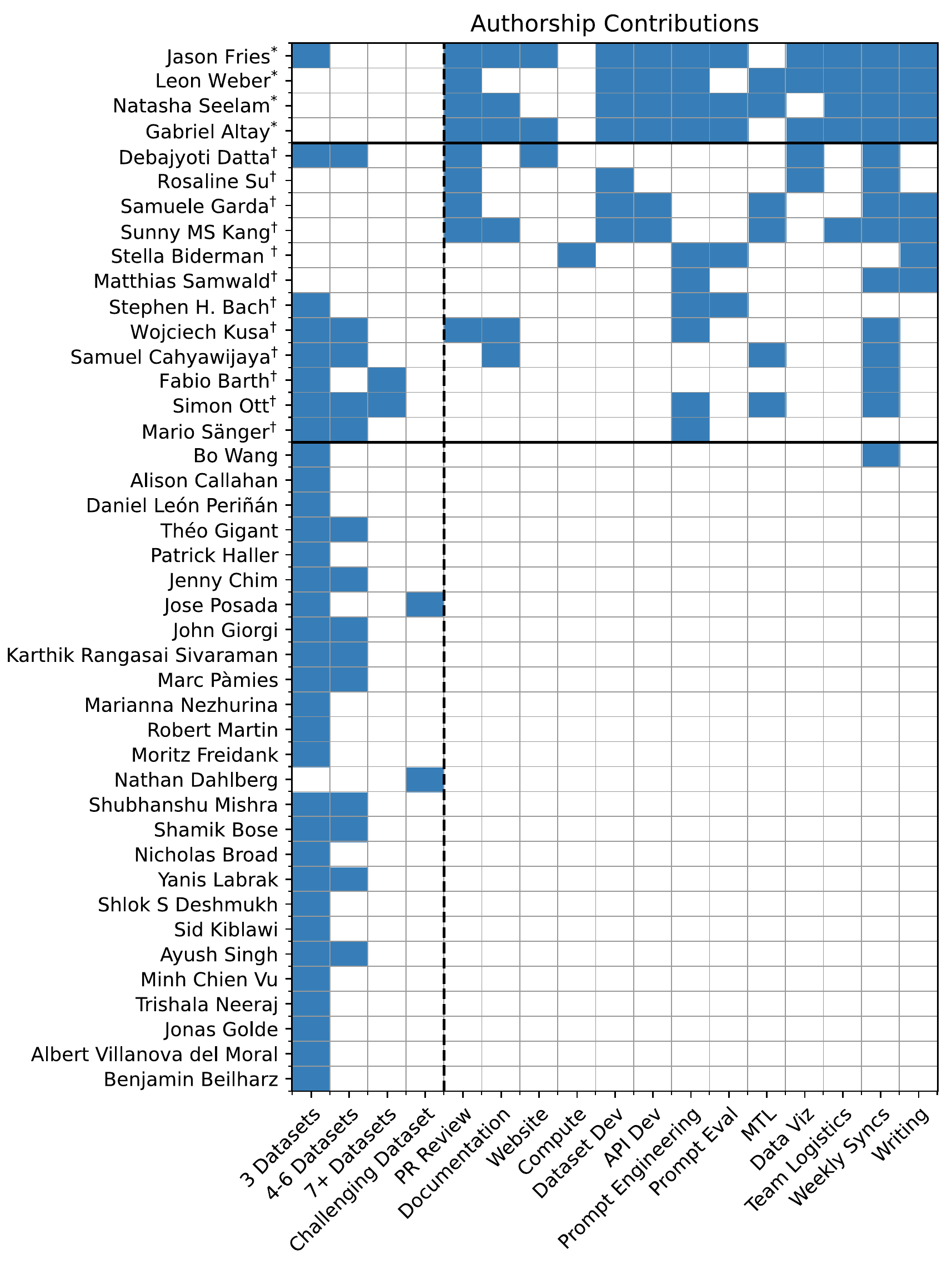} 
  \caption{\label{fig:authorship}Authorship contribution matrix. Cells to the left of the dotted black vertical line are hackathon dataset contributions, while the right are other paper contributions as part of the BigScience biomedical working group. For each author, $*$ denotes co-first author and $\dagger$ denotes co-second author, with equal contributions within category. }
\end{figure}

Specific contribution categories are listed below and visualized by author in Figure \ref{fig:authorship}.

\begin{itemize}
    \item \textbf{3 Datasets, 4-6 Datasets, 7+ Datasets}: Number of dataset loaders coded during the hackathon.
    \item \textbf{Challenging Dataset}: Implemented a difficult dataset loader (e.g., many label errors, poor documentation on structure). 
    \item \textbf{PR Review}: Managed PR process during hackathon, including code review, debugging, and other quality control measures. This includes llive QA sessions during hackathon office hours on the team Discord server. 
    \item \textbf{Documentation}: Wrote instructional material for participants on designing data loaders,  coding tutorials, and logistics material for hackathon participation
    \item \textbf{Website}: Contributed to the creation of the BigBIO hackathon website. 
    \item \textbf{Compute}: Provided computational resources for running machine learning experiments.
    \item \textbf{Dataset Dev}: Contributed to the design and implementation of task schema design, designing dataset loaders, data unit tests, and other dataset loader infrastructure.
    \item \textbf{API Dev}: Contributed to the design and development of the \ours~ API, including querying of metadata, programmatic access across datasets, and other infrastructure.
    \item \textbf{Prompt Engineering}: Designed biomedical dataset prompts in PromptSource
    \item \textbf{Prompt Eval}: Contributed to the infrastructure of connecting \ours~ data loaders with the language model evaluation harness and/or ran prompt evaluation experiments.
    \item \textbf{MTL}: Contributed to the multi-task learning experiments 
    \item \textbf{Data Viz}: Designed data visualizations
    \item \textbf{Team Logistics}: Organizational tracking of team goals and action items. 
    \item \textbf{Weekly Syncs}: Attended and contributed to weekly team meetings
    \item \textbf{Writing}: Contributed text or edited content within this manuscript 
\end{itemize}

%% file: 12_appx_schema.tex
\clearpage


\section{Task Schema and Harmonization}
\label{sec:schema}



We have defined a set of lightweight, task-specific schema to help simplify programmatic access to common biomedical datasets.

Each dataset loader implemented in \ours~ provides at least one \texttt{source} view of the dataset and at least one \texttt{bigbio} view of the dataset. The \texttt{source} view attempts to capture the original form of the dataset with as little change as possible. The \texttt{bigbio} view attempts to normalize the dataset into one of our \ours~ task-specific schemas. All schemas are defined by creating an instance of the \texttt{datasets.Features} class from the Hugging Face datasets package. 

Every element of the \ours~ schemas has an \texttt{id} attribute that is unique across the dataset. In some datasets, entities are represented as discontiguous spans. For example, the string "estrogen and progesterone receptor positive" could be labeled with two entities and two lists of character offsets,

\begin{verbatim}
    ["estrogen", "receptor"]; [(0,8), (26,34)]
    ["progesterone receptor"]; [(13, 34)]
\end{verbatim}

To support these types of annotations and maintain consistency, we represent all text-offset combinations this way. 

\subsection{Schema Definitions}

\paragraph{Knowledge Base (KB)}

The knowledge base schema covers entity based tasks and includes named entity recognition (NER), named entity disambiguation/normalization (NED), event extraction (EE), relation extraction (RE), and coreference resolution (COREF).
The schema is loosely based on the XML BioC format \cite{comeau2013bioc} and the brat annotation format \cite{stenetorp2012brat}.
The top level features are, 

\begin{verbatim}
{
    "id": datasets.Value("string"),
    "document_id": datasets.Value("string"),
    "passages": [],
    "entities": [],
    "events": [],
    "coreferences": [],
    "relations": [],
}
\end{verbatim}

The \texttt{id} attribute can be set to anything that makes it unique and the \texttt{document\_id} attribute represents any identifying value included in the original dataset. Passages capture the text content of a sample. A single sample can have one passage (such as a single abstract) or multiple elements (such as abstract and title). The character offsets in the rest of the KB schema elements index into the string that would be created by joining all the passage texts. 

\begin{verbatim}
"passages": [
    {
        "id": datasets.Value("string"),
        "type": datasets.Value("string"),
        "text": datasets.Sequence(datasets.Value("string")),
        "offsets": datasets.Sequence([datasets.Value("int32")]),
    }
]
\end{verbatim}

Entities can be associated with a type as well as multiple database entries. 

\begin{verbatim}
"entities": [
    {
        "id": datasets.Value("string"),
        "type": datasets.Value("string"),
        "text": datasets.Sequence(datasets.Value("string")),
        "offsets": datasets.Sequence([datasets.Value("int32")]),
        "normalized": [
            {
                "db_name": datasets.Value("string"),
                "db_id": datasets.Value("string"),
            }
        ],
    }
]
\end{verbatim}

Events are modeled in \ours~ as they are in the brat annotation tool. 

\begin{verbatim}
"events": [
    {
        "id": datasets.Value("string"),
        "type": datasets.Value("string"),
        "trigger": {
            "text": datasets.Sequence(datasets.Value("string")),
            "offsets": datasets.Sequence([datasets.Value("int32")]),
        },
        "arguments": [
            {
                "role": datasets.Value("string"),
                "ref_id": datasets.Value("string"),
            }
        ],
    }
]
\end{verbatim}

Coreference annotations can be specified using a sequence of entity IDs. 

\begin{verbatim}
"coreferences": [
    {
        "id": datasets.Value("string"),
        "entity_ids": datasets.Sequence(datasets.Value("string")),
    }
]
\end{verbatim}

Binary typed relations with multiple database normalizations are also supported.  

\begin{verbatim}
"relations": [
    {
        "id": datasets.Value("string"),
        "type": datasets.Value("string"),
        "arg1_id": datasets.Value("string"),
        "arg2_id": datasets.Value("string"),
        "normalized": [
            {
                "db_name": datasets.Value("string"),
                "db_id": datasets.Value("string"),
            }
        ],
    }
]
\end{verbatim}

\paragraph{Question Answering (QA)}
The QA schema supports several question answering tasks. 
The \texttt{type} attribute is not constrained but takes the values "factoid", "how", "list", "multiple\_choice", "summary", "why", and "yesno" in the current \ours~ datasets.
For "multiple\_choice" and "yesno" questions, the \texttt{choices} attribute is populated with valid answers. The \texttt{context} attribute is used for closed-domain QA.

\begin{verbatim}
{
    "id": datasets.Value("string"),
    "question_id": datasets.Value("string"),
    "document_id": datasets.Value("string"),
    "question": datasets.Value("string"),
    "type": datasets.Value("string"),
    "choices": [datasets.Value("string")],
    "context": datasets.Value("string"),
    "answer": datasets.Sequence(datasets.Value("string")),
}
\end{verbatim}

\paragraph{Textual Entailment (TE)}
The TE schema supports tasks in which two text spans can be mapped onto the triplet of entailment labels ("entailment", "neutral", "contradict"). 

\begin{verbatim}
{
    "id": datasets.Value("string"),
    "premise": datasets.Value("string"),
    "hypothesis": datasets.Value("string"),
    "label": datasets.Value("string"),
}
\end{verbatim}

\paragraph{Text (TEXT)}
The TEXT schema supports tasks with a single text span and one or more associated labels (TXTCLASS). 

\begin{verbatim}
{
    "id": datasets.Value("string"),
    "document_id": datasets.Value("string"),
    "text": datasets.Value("string"),
    "labels": [datasets.Value("string")],
}
\end{verbatim}

\paragraph{Text Pairs (PAIRS)}
The PAIRS schema supports tasks with two text spans and one label. In this initial release, the only task using this schema is semantic similarity (STS).  

\begin{verbatim}
{
    "id": datasets.Value("string"),
    "document_id": datasets.Value("string"),
    "text_1": datasets.Value("string"),
    "text_2": datasets.Value("string"),
    "label": datasets.Value("string"),
}
\end{verbatim}

\paragraph{Text to Text (T2T)}
The T2T schema supports sequence to sequence tasks such as paraphasing (PARA), 
translation (TRANSL), and summarization (SUM). 

\begin{verbatim}
{
    "id": datasets.Value("string"),
    "document_id": datasets.Value("string"),
    "text_1": datasets.Value("string"),
    "text_2": datasets.Value("string"),
    "text_1_name": datasets.Value("string"),
    "text_2_name": datasets.Value("string"),
}
\end{verbatim}

\subsection{Harmonization}

Harmonization efforts aimed for the simplest schema, per task, that was able to flexibly cover the majority of relevant features. We found in the majority of cases, the schema provided suited the task of the original dataset. Toward that end, we found that only 22$\%$ (29/129 datasets submitted) of the datasets required major refactors (defined by significant changes or fixes to the dataloader post submission). While the schema satisfied most cases, we noted some areas of improvement below:

\paragraph{Extension of question answering}
Question-answering supports multiple choice, binary choice, or span-based answers, but does not enable `long-form' responses that may provide greater context to the question asked.
This particular issue arose in PubMedQA, of which the source schema has a context key that provides framing for the answer.

\paragraph{Extension of text pairs classification}
The text-pairs schema enables a relationship between two input texts and their corresponding labels.
However, in at least one dataset (Scielo), a three-language translation was provided.
This can be handled be implementing the dataset twice, one for each translation, or omitting this feature altogether. 

\paragraph{Multi-label entities}
Several datasets had multiple labels associated to a single entity. While we have adapted the schema to associate multiple labels to a single entity. To resolve this concern, we duplicate the feature but change the label and provide a new unique id. This concern was particularly noted in the \texttt{MedMentions} dataset.

\paragraph{Diverse label representations}
For classification problems, the labels associated to a feature may be a string answer, or a numerical score.
To maintain a consistent format across all datasets, label keys across schemas in the \ours-view are always \texttt{str} types.
This limitation affected at least 4 datasets (UMNSRS, MayoSRS, BioSimVerb), particulary in the context of semantic similarity scores across text. For the user to appropriately cast the score type, they would need familiarity of the dataset. We opted to enable the source view to represent label information for scores as floats when present.

\paragraph{Unsupported task types}
In certain cases, tasks may extend beyond the descriptive capacity of the provided \ours-schemas. For example, tasks that explicitly required contextualization were unable to fit into a pre-existing schema.
For example, speech-based tasks, such as MedDialogue require a text, label, and potential context; the \ours-text classification schema does not enable a context key.
Additionally, Ask-a-Patient required a tuple-like structure to represent a text, a social media response, and a medical concept to be relevant to the task. 
In addition to tasks that require context, part-of-speech tagging or annotations on a per-token basis was not easily represented in our pre-existing schema. 

During the initiative, common themes of recurring problems in biomedical NLP processing occurred.
We denote them as follows:

\paragraph{Issues with offsets}
One of the unit-tests specifically monitored whether reported features matched offsets provided from the original dataset. We found a several datasets with slight offset errors, or inconsistencies. In several cases, offset errors included off-by-one or whitespacing considerations, discontiguous spans, and one case, entirely omitted from the original dataset. 

\paragraph{Large datasets}
Several datasets possessed corpora that were large in size (upwards of 20 GB).
In at least one instance, the initial implementation of the dataset yielded examples exceedingly slow.
While we standardized information content, we did not explicitly optimize for efficiency.

%% file: 13_appx_metadata.tex
\clearpage
\section{Dataset Metadata}
\label{sec:metadata}

\begin{figure}[ht!]
  \centering
  \includegraphics[scale=0.29]{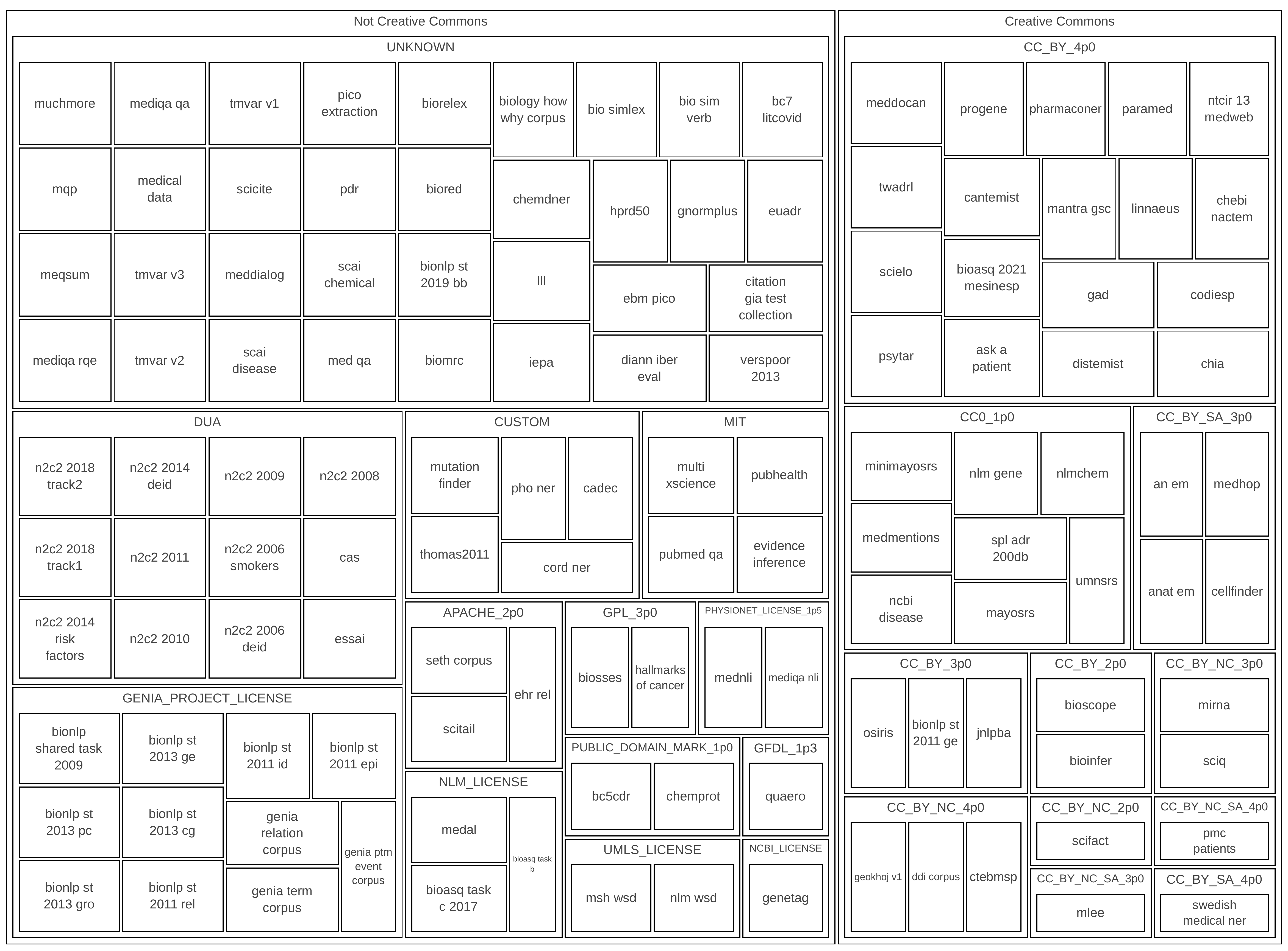}
  \caption{\label{fig:licensetree}Treemap visualization of datasets by license. }
\end{figure}

\begin{figure}[ht!]
  \centering
  \includegraphics[scale=0.29]{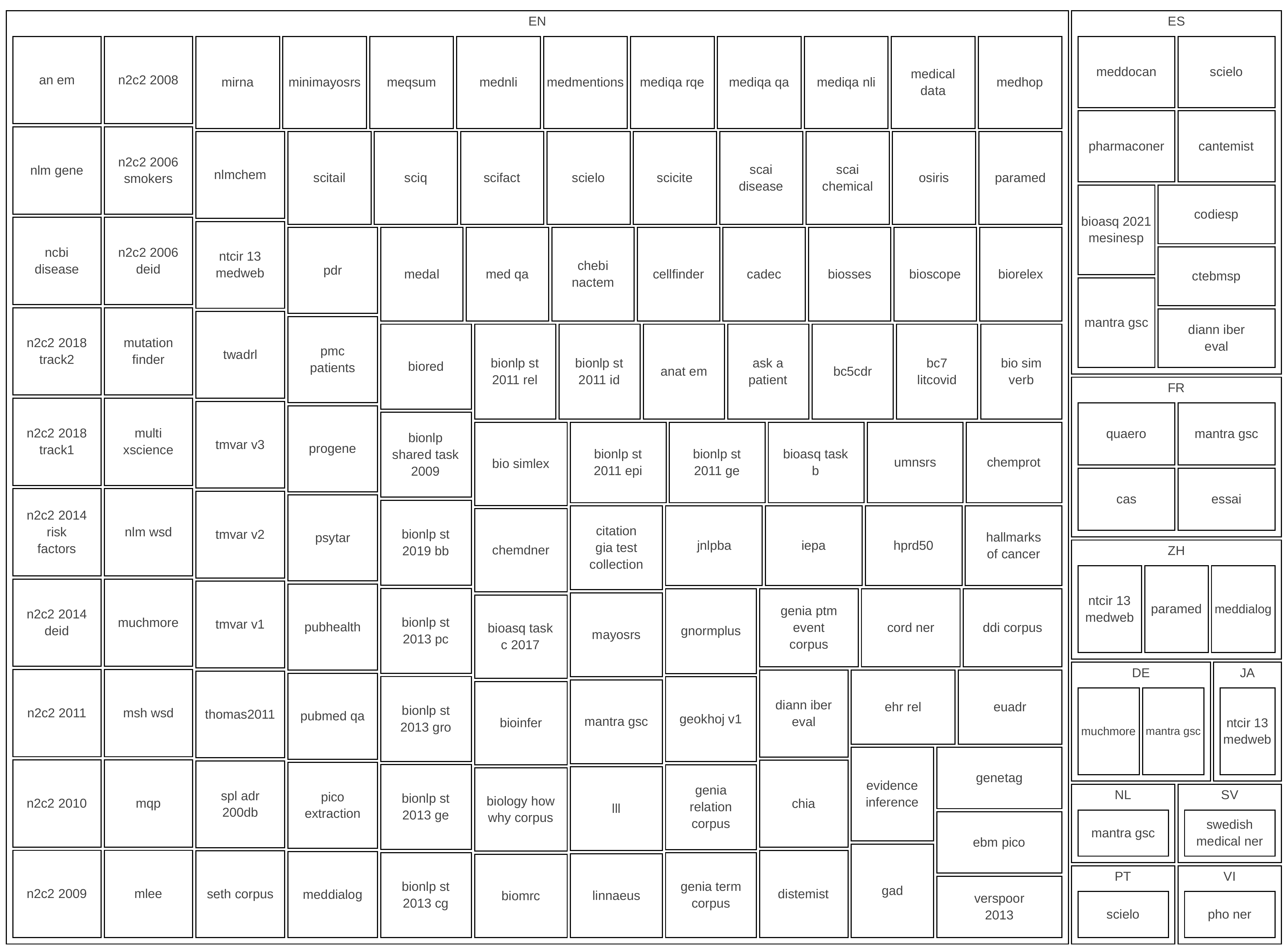}
  \caption{\label{fig:languagetree}Treemap visualization of datasets by language. }
\end{figure}

We collected the structured metadata outlined in Table \ref{tbl:metadata} for all datasets in the \ours~catalog. 
Required elements are written as code in the data loader.
Figures \ref{fig:licensetree} and \ref{fig:languagetree} show treemap visualizations of all datasets based on their license and language respectively. 

\begin{table}[ht!]
\centering
\caption{\label{tbl:metadata} Metadata collected for all datasets.}
\begin{tabular}{lll}
\toprule
Field    & Required & Description \\
\midrule
Name        & \checkmark   & Dataset name \\ 
Task Types  & \checkmark    & NER, question answering, coreference resolution, etc.    \\ 
Domain      & \checkmark    & Corpora domain: biomedical or clinical/health-related    \\ 
PubMed/PMC  & \checkmark    & Corpora are from PubMed/PubMed Central (PMC)    \\ 
Splits      & \checkmark    & Canonical definitions for training/validation/testing splits \\
Publication & \checkmark    & Manuscript describing dataset     \\ 
Year        &   & Publication year            \\ 
Homepage    & \checkmark    & Website describing dataset     \\ 
Public URL  & \checkmark     & Open URL (no authentication)     \\ 
Private     & \checkmark    & Requires authentication/credentialing     \\ 
License     & \checkmark    & Provided license type \\ 
Languages   & \checkmark    & Included languages \\ 
Multilingual &  & Parallel corpora \\ 
Annotation Source   & & Expert label provenance (e.g., hand labeled, silver labels) \\ 
\bottomrule
\end{tabular}
\end{table}

%% file: 14_appx_unit_tests.tex
\clearpage
\section{Unit Tests}
\label{sec:units}

We developed 11 unit tests to check the \ours versions of all implemented data loaders.
Unit tests run on all \ours~\textit{configurations} (i.e., a schema view of the dataset) found within a dataset, whether they represent different dataset subsets or different tasks. 

Among all implemented unit tests, we differentiate between \textbf{global} and \textbf{task-specific} tests. 
For datasets that support configurations with multiple schemas (each supporting different tasks), we run the task-specific tests using only the configuration supporting the task. 


Below, we describe each unit test found in \href{https://github.com/bigscience-workshop/biomedical/blob/master/tests/test_bigbio.py}{\ours}:

\subsection{Global Tests}
\begin{enumerate}

    \item \textbf{Metadata} Checks if the dataloader module provides relevant metadata attributes. Supported attributes include \verb|LANGUAGE| (language of the dataset), \verb|LOCAL| (whether the dataset is publicly accessible or requires local files), \verb|PUBMED| (is part of Pubmed), and
    \verb|LICENSE| (type of license). The \verb|LANGUAGE| and \verb|LICENSE| are standardized to common labels across datasets, whereas \verb|LOCAL| and \verb|PUBMED| are boolean.
    
    \item \textbf{Unique Global IDs} Each element within a dataset is assigned a string ID that is unique across the dataset split (such as train, validation or test). For example, all passages, entities, relations, questions, labels, and other attributes will be assigned a unique string. This ID can be used to reference a given element if it is being used in a new context without considering explicit text overlap or other heuristics. This unit-test confirms that a every element has an ID that is unique across the full dataset split.
    
    \item \textbf{Schema} This test checks whether the populated fields in the examples are consistent with the tasks supported by the dataset. 
    For instance, if a dataset is annotated to support NER but there is not a single entity field populated across a full dataset split, the test will fail. 
    Additionally, the test will provide a warning if fields are populated that would support a task missing from the annotated supported tasks.
    The loading procedure in Hugging Face's datasets fails if a dataloader does not adhere to its defined schema.
    Thus, we implicitly check for consistency between data and schema by loading the dataset.
    
    \item \textbf{Feature Statistics} This test prints statistics of populated fields in the dataset to allow the user to manually check their plausibility.
    For each data split, it collects the number of elements (e.g. number of entities, relations, text pairs, etc.).
    We use these statistics for quality control by manually comparing to the dataset statistics reported in the publication describing the respective dataset.
\end{enumerate}


\subsection{Task-specific Tests: Knowledge Base}

\begin{enumerate}
    \item \textbf{Referenced ids} Certain fields may be referenced by other elements (for example, a relation usually references two entities). References in the \ours-schema will use the unique ID assigned to them. This unit test checks if all referenced IDs exist, and have an appropriate type. For instance, it makes sure that the arguments of a relation are indeed entities (and not relations or events).  
    
    \item \textbf{Passage Offsets} This test checks whether the start and end indices of all passages are correct. This is achieved by comparing the text span defined by the indices to the text field assigned to the passage. Additionally, the unit test will make sure that each passage is contiguous and does not overlap.
    
    \item \textbf{Entity Offsets} This test makes sure that the start and end indices of entities are correct. Analogous to the \textit{Passage Offsets} test, we compare the reported feature text for entities versus the extracted text from the start/ending index provided from the data. This test does not provide an explicit failure, but instead warns the user of all entities that do not explicitly match their offset-extracted text. We chose a warning over failure because some datasets contain faulty offsets in the original formats due to annotation errors.
    
    \item \textbf{Event Offsets} Similar to the passage-offsets and entities-offset check, we compare the reported event text feature to the extracted text from provided offsets. We warn the user of any instances of discordance between the reported and extracted text.
    
    \item \textbf{Multi-label Entities} The current \ours~schema does not support multiple types for entities. This test flags instances where an entity is assigned multiple types by concatenating the types with common connector symbols (such as `|' or `;'). 
    
    \item \textbf{Multi-label Types} This unit-test performs the same check as \verb|Multi-label Entities| for other features with the \verb|type| attribute (passages, relations, events). This test is distinct from the multi-label entities test, because the envisioned \ours~schema revision to support multiple labels is different in this case.

    
    
\end{enumerate}

\subsection{Task-specific Tests: Question Answering}

\begin{enumerate}
    \item \textbf{Multiple Choice} This test checks whether the answers of a question-answering schema are either multiple choice or binary (yes/no). It verifies that the answer provided exists in the choices available for each example.
\end{enumerate}

All accepted data-loading scripts must pass code review, unit-tests, and implement explicit fixes for warnings that indicated destructive transformations of the original dataset (such as introducing faulty offsets).

In general, participants who implemented data-loading scripts were asked to refrain from resolving dataset issues in the dataloader for the original dataset but were free to fix the issues for the \ours~versions.
Any data quality changes were explicitly annotated within the review process, and the data loading script itself. 
 
Certain datasets may require specific keys to be ignored.
We implemented functions that allow a user to bypass a specific key (e.g., skip all events), a data split (e.g., skip the validation set), or a specific key within a dataset (e.g., skip relation labels in the test set).
These functions were used to check the \verb|BioNLP shared task| datasets, as the test splits of these datasets omitted annotations for some supported tasks.
These bypass functions allow a user to test if all other aspects of the dataset implementation work as intended.

%% file: 15_appx_pr_checklist.tex
\clearpage
\section{Dataset Submission Checklist}
\label{sec:appx:checklist}


\begin{itemize}
\item[$\square$] Confirm that this PR is linked to the dataset issue.
\item[$\square$] Create the dataloader script {\tt biodatasets/my\_dataset/my\_dataset.py} (please use only lowercase and underscore for dataset naming). 
\item[$\square$] Provide values for
\begin{itemize}
\item[$\square$] {\tt \_CITATION}
\item[$\square$] {\tt \_DATASETNAME}
\item[$\square$]{\tt \_DESCRIPTION}
\item[$\square$]{\tt \_HOMEPAGE}
\item[$\square$]{\tt \_LICENSE}
\item[$\square$]{\tt \_URLs}
\item[$\square$]{\tt \_SUPPORTED\_TASKS}
\item[$\square$]{\tt \_SOURCE\_VERSION}
\item[$\square$]{\tt \_BIGBIO\_VERSION}
\end{itemize}
\item[$\square$] Data loader implementations for
\begin{itemize}
\item[$\square$]{\tt \_info()}
\item[$\square$]{\tt \_split\_generators()} 
\item[$\square$]{\tt \_generate\_examples()}
\end{itemize}
\item[$\square$] Make sure that the {\tt BUILDER\_CONFIGS} class attribute is a list with at least one `BigBioConfig` for the source schema and one for a bigbio schema.
\item[$\square$] Confirm dataloader script works with {\tt datasets.load\_dataset} function.
\item[$\square$] Confirm that your dataloader script passes the test suite run with \\ {\tt python -m tests.test\_bigbio biodatasets/my\_dataset/my\_dataset.py}.
\item[$\square$] If my dataset is local, I have provided an output of the unit-tests in the PR (please copy paste). This is OPTIONAL for public datasets, as we can test these without access to the data files.
\end{itemize}

%% file: 16_appx_hackathon.tex
\clearpage
\section{BigScience Biomedical Hackathon}
\label{sec:appx:hackathon}

\begin{figure}[ht!]
  \centering 
  \includegraphics[width=385px,height=206px]{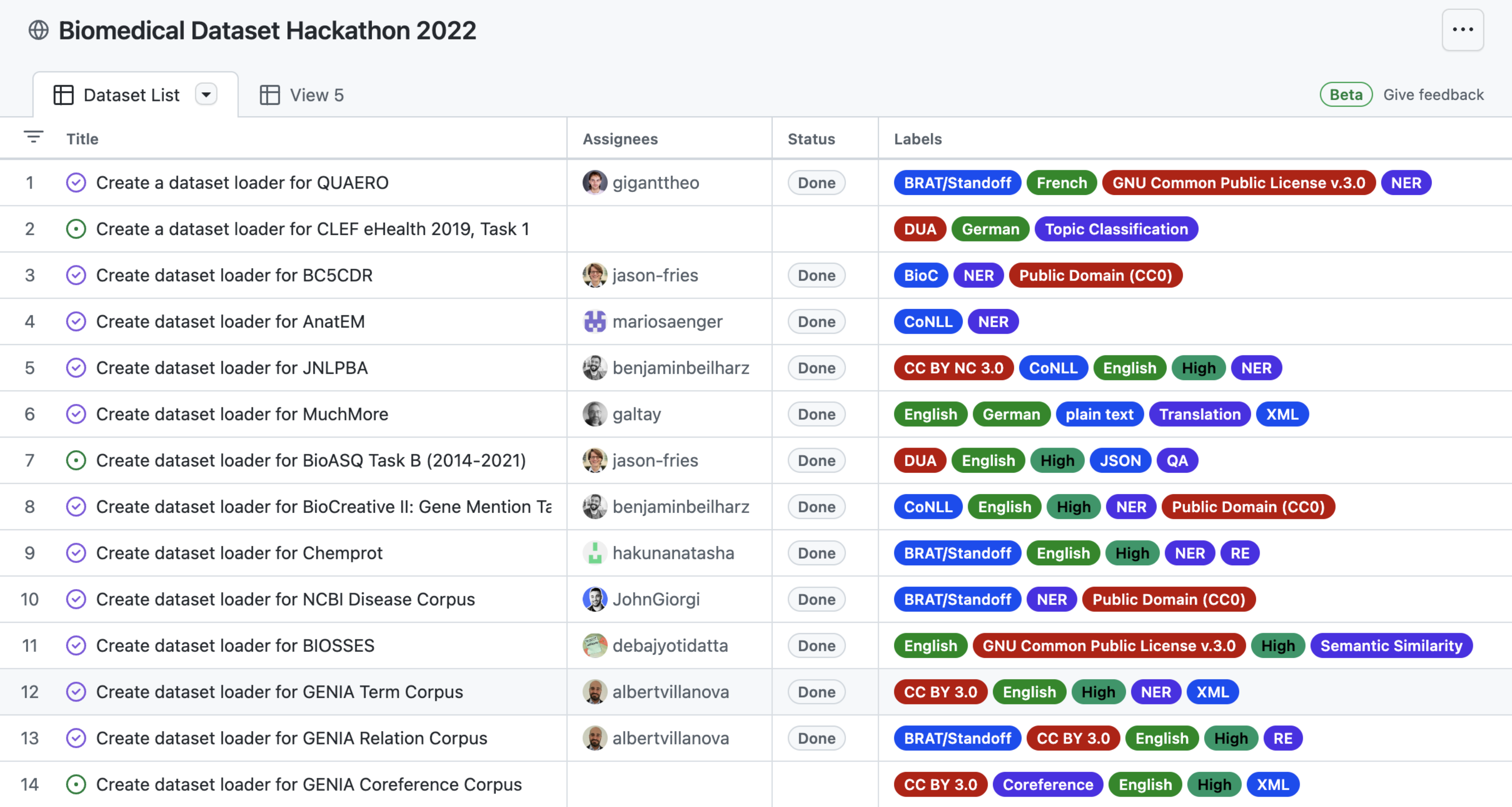} 
  \caption{\label{fig:projectboard}Participants volunteered to implement dataset loaders using GitHub project tracking tools.}
\end{figure}

We catalogued an initial set of 174 datasets and prior to launching the hackathon, we provided users with a \href{https://github.com/orgs/bigscience-workshop/projects/6/views/1?visibleFields=%5B%22Title%22%2C%22Assignees%22%2C%22Status%22%2C%22Labels%22%2C1774350%2C1774506%2C1774652%2C2832331%5D}{project board} that tagged each dataset as a new issue within our GitHub repository.
For all datasets, we provided meta-data tags such as language, license, and associated task (e.g., NER, question answering).
Participants could assign themselves to a dataset via issues and status would be reflected in the project board (see Figure 
\ref{fig:projectboard}).
Admins could change the status of the issue based on progress of the data loading script.

Participants were asked to create a fork of the repository, and implement their data-loading script.
We provided a \href{https://github.com/bigscience-workshop/biomedical/blob/master/templates/template.py}{template} of a dataloading script, where explicit comments were left to indicate key functions and attributes the participant must complete. 
For datasets in common formats like BRAT or BioC, we provided utility functions to improve standardization across formats.
At minimum, participants implemented an {\tt \_info\_} function that instantiated the {\tt source} and {\tt bigbio} configs. 
A {\tt \_split\_generators} function that identified how to access each data split in the dataset, and the \verb|_generate_examples|  that extracted relevant information from each data split according to the specifications of the configs. 

Dataloader scripts were submitted through pull-requests (PRs) on GitHub.
Prior to submitting code for review, we asked participants to check if the code passed unit-tests and style guidelines.
Accepted PRs required at least 1 admin approval to merge to the library.
To respect data governance, we did not accept any submissions that provided explicit dataset files. Dataloading scripts must access datasets via URLs, or expect a filepath to the local dataset. 

If a dataset had multiple tasks, we asked the participant to implement tasks based on the number of unique schemas, if possible. Some datasets possess different views based on the different tasks that can be performed on them.
Participants were told to handle multiple annotations/harmonization per the original dataset's recommendations. If none were given, participants were asked to choose what seemed reasonable, and iterate with an admin.

All contribution instructions may be found \href{https://github.com/bigscience-workshop/biomedical/blob/master/CONTRIBUTING.md}{here}.

Of the 174 datasets identified, 126 datasets satisfied the acceptance criteria, including the checklist in \S\ref{sec:appx:checklist}, code-review, and passing unit-tests.
Exceptions were made on a case-by-case basis for datasets with unique challenges that extended beyond the scope of the schema provided.

\subsection{Frequently Asked Questions (FAQ)}

During the hackthon, we developed the following list of frequently asked questions (FAQ). 

\textbf{How can I find the appropriate license for my dataset?}
The license for a dataset is not always obvious. Here are some strategies to try in your search:
\begin{enumerate}
    \item Check the \href{https://docs.google.com/spreadsheets/d/1eOa9NhNmgGLByWKZ9ioKmNErq824dGA-nV5WpRWZ4a8/edit?usp=sharing}{Experiment A: Annotated Datasets} sheet of the we used while planning the hackathon %
    \item Check for files such as README or LICENSE that may be distributed with the dataset itself
    \item Check the dataset webpage
    \item Check publications that announce the release of the dataset
    \item Check the website of the organization providing the dataset
\end{enumerate}

If no official license is listed anywhere, but you find a webpage that describes general data usage policies for the dataset, you can fall back to providing that URL in the {\tt\_LICENSE} variable.
If you can't find any license information, please make a note in your PR and put {\tt\_LICENSE = "Unknown"} in your dataset script.

\textbf{What if my dataset is not publicly available?}
We understand that some biomedical datasets are not publicly available due to data usage agreements or licensing.
For these datasets, we recommend implementing a dataloader script that references a local directory containing the dataset.
You can find examples in the \href{https://github.com/bigscience-workshop/biomedical/blob/master/examples/n2c2_2011.py}{{\tt n2c2\_2011}} and \href{https://github.com/bigscience-workshop/biomedical/blob/master/examples/bioasq.py}{{\tt bioasq}} implementations. There are also local dataset specific instructions in \href{https://github.com/bigscience-workshop/biomedical/blob/master/templates/template.py}{template}. 

\textbf{What types of libraries can we import?}
Eventually, your dataloader script will need to run using only the packages supplied by the \href{https://github.com/huggingface/datasets}{datasets} package. If you find a well supported package that makes your implementation easier (e.g. \href{https://github.com/bionlplab/bioc}{bioc}), then feel free to use it.

We will address the specifics during review of your PR to the \href{https://github.com/bigscience-workshop/biomedical}{BigScience biomedical repo} and find a way to make it usable in the final submission to \href{https://huggingface.co/bigscience-biomedical}{huggingface bigscience-biomedical}

\textbf{Can I upload my dataset anywhere?}
No. Please don't upload the dataset you're working on to the huggingface hub or anywhere else. This is not the goal of the hackathon and some datasets have licensing agreements that prevent redistribution. If the dataset is public, include a downloading component in your dataset loader script. Otherwise, include only an "extraction from local files" component in your dataset loader script. If you have a custom dataset you would like to submit, please \href{https://github.com/bigscience-workshop/biomedical/issues/new}{make an issue} and an admin will get back to you.

\textbf{My dataset supports multiple tasks with different bigbio schemas. What should I do?}
In some cases, a single dataset will support multiple tasks with different bigbio schemas.
For example, the muchmore dataset can be used for a translation task (supported by the Text to Text (T2T) schema) and a named entity recognition task (supported by the Knowledge Base (KB) schema).
In this case, please implement one config for each supported schema and name the config {\tt <datasetname>\_bigbio\_<schema>}.
In the muchmore example, this would mean one config called {\tt muchmore\_bigbio\_t2t } and one config called {\tt muchmore\_bigbio\_kb}.

\textbf{My dataset comes with multiple annotations per text and no/multiple harmonizations. How should I proceed?}
Please implement all different annotations and harmonizations as source versions (see \href{https://github.com/bigscience-workshop/biomedical/blob/master/examples/bioasq.py}{{\tt examples/bioasq.py}} for an example). 
If the authors suggest a preferred harmonization, use that for the bigbio version.
Otherwise use the harmonization that you think is best.

\textbf{How should I handle offsets and text in the bigbio schema?}
Full details on how to handle offsets and text in the bigbio kb schema can be found in the \href{https://github.com/bigscience-workshop/biomedical/blob/master/task_schemas.md}{schema documentation}.

\textbf{My dataset is complicated, can you help me?}
Yes! Please feel free to leave a question in questions or ping the admins directly with @admins.
We will be hosting office hours round the clock to be able to answer you in a timely manner! 

\textbf{My dataset is too complicated, can I switch?}
Yes! Some datasets are easier to write dataloader scripts for than others.
If you find yourself working on a dataset that you can not make progress on, please make a comment in the associated issue, asked to be un-assigned from the issue, and start the search for a new unclaimed dataset.
You are also welcome to ping the admins - we are happy to help you!

\textbf{Can I change the Big-Bio schema?}
No, please do not modify the Big-Bio Schema.
The goal of this hackathon is to enable simple, programmatic access to a large variety of biomedical datasets. Part of this requires having a dependable interface.
We developed our schema to address the most salient types of questions to ask of the datasets.
We would be more than happy to discuss your suggestions, and you are welcome to implement it as a new config.

\textbf{My dataset has multiple labels to a span of text - what do I do?}
In many of our schemas, we have a 1:1 mapping between a key and its label (i.e. in KB, entity and label). In some datasets, we've noticed that there are multiple labels assigned to a text entity. Generally speaking, if a big-bio key has multiple labels associated with it, please populate the list with multiple instances of (key, label) according to each label that correspond to it. 

So for instance if the dataset has an entity "copper" with the  types "Pharmacologic Substance" and "Biologically Active", please create one entity with type "Pharmacologic Substance" and an associated unique id and another entity with type "Biologically Active" with a different unique id. The rest of the inputs (text, offsets, and normalization) of both entities will be identical.

\textbf{What happens after I claim a dataset?}
In order to keep turnaround time reasonable, and ensure datasets are being completed, we propose a few notes on claiming a dataset:

\begin{enumerate}
\item Please claim a dataset only if you intend to work on it. We'll try to check in within 3 days to ensure you have the help you need. Don't hesitate to contact the admins! We are ready to help!
\item If you have already claimed a dataset prior to (2022/04/05), we will check in on Friday (2022/04/08). If we do not hear back via GitHub issues OR a message to the Discord admins on general, we will make the dataset open for other participants by Saturday (2022/04/09).
\item If things are taking longer than expected - that is totally ok! Please let us know via GitHub issues (preferred) or by pinging the @admins channel on Discord.
\end{enumerate}

%% file: 17_appx_data_dedup.tex
\clearpage
\section{Assessing Dataset Overlap}

\begin{figure}[ht!]
  \centering
  \includegraphics[scale=0.40]{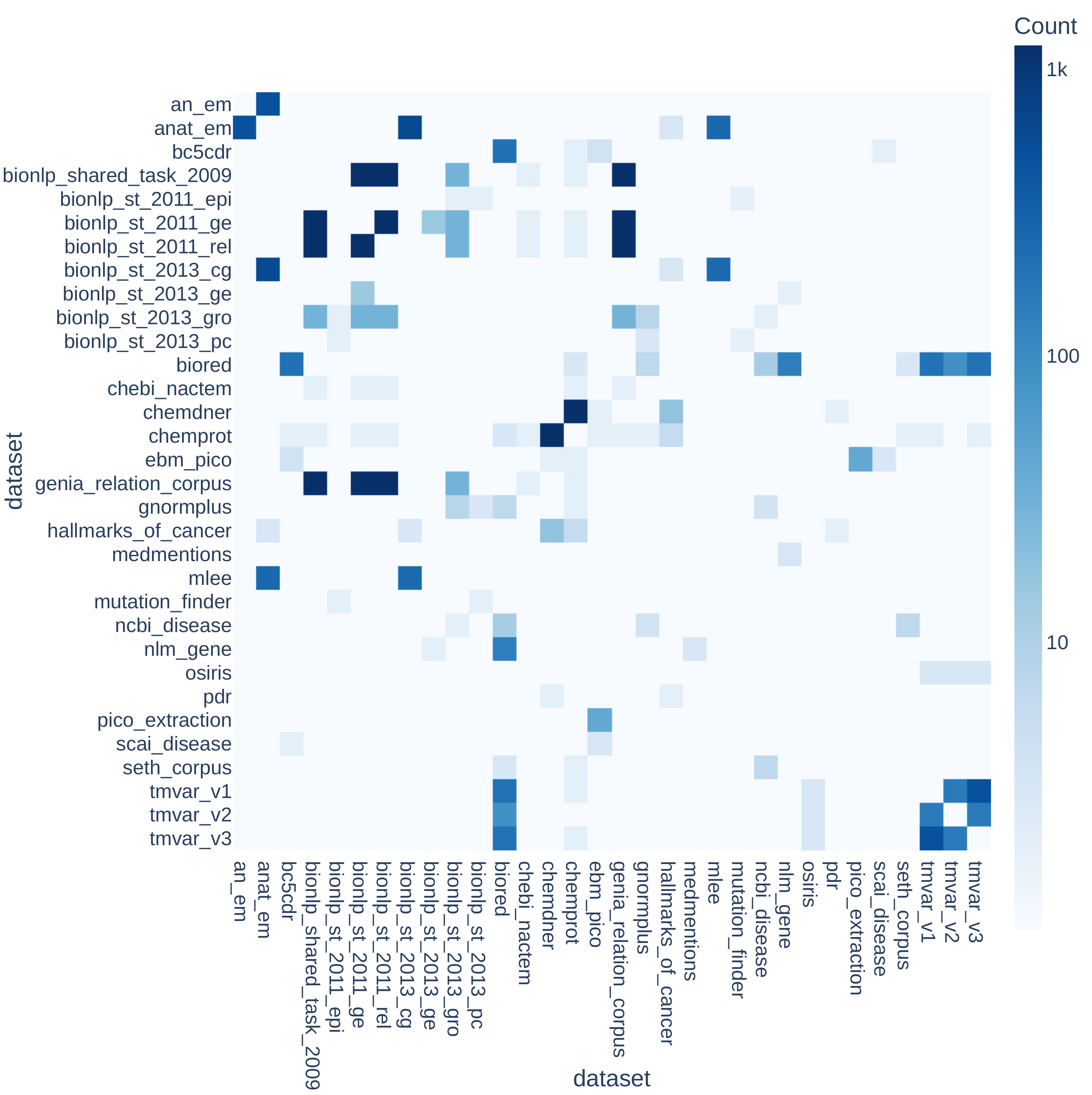}
  \caption{\label{fig:pubmed-heatmap}A heatmap representation of PubMed overlap between public datasets in \ours. Each cell is shaded using the log count of PMIDs shared by the pair of datasets it represents. 
  }
\end{figure}

\begin{table}[htbp]
  \centering
  \caption{Example document IDs as they appear in the original source datasets and their corresponding \ours~ normalization to PubMed PMIDs, Pubmed Central PMCIDs, and journal titles.}
    \begin{tabular}{llll}
    \toprule
Original Document ID & PMID & PMCID & Journal \\
    \midrule
    PMID-12604762 & 12604762 & PMC1497507 & Public Health Rep \\
    BB-kb+ner-F-25496341-000 & 25496341 & PMC4320590 & BMC Genomics \\
    17389645\_04\_discussion & 17389645 & PMC1885650 & Nucleic Acids Res \\
    pmcA2538543 & 2538543 & PMC2189270 & J Exp Med \\
    10747015-3 & 10747015 & PMC310216 & EMBO J \\
    6421395:4 & 6421395 & PMC1444356 & Br Med J (Clin Res Ed) \\
    PMC2885601-03-RESULTS-01 & 20556207 & PMC2885601 & Open Microbiol J \\
    PMC-2626671-01-INTRODUCTION & 19139168 & PMC2626671 & J Exp Med \\
    \bottomrule
    \end{tabular}
  \label{tab:pubmed-norm-sample}
\end{table}

%
%
\begin{table}[htbp]
  \centering
  \caption{Dataset clusters of document (PMID) overlap. }
    \begin{tabular}{p{9.25cm}cc}
    \toprule
Dataset Names & Count & PMID Overlap \\
\midrule
BioRED, NCBI Disease & 2 & 11 \\
\midrule
MLEE, AnatEM & 2 & 12 \\
\midrule
Hallmarks of Cancer, CHEMDNER & 2 & 12 \\
\midrule
BioNLP ST 2013 GE, BioNLP ST 2011 GE & 2 & 14 \\
\midrule
BioNLP ST 2011 REL, BioNLP ST 2013 GRO, GENIA Relation Corpus, BioNLP Shared Task 2009, BioNLP ST 2011 GE & 5 & 29 \\
\midrule
PICO Extraction, EBM PICO & 2 & 41 \\
\midrule
tmVar v1, tmVar v2, tmVar v3 & 3 & 69 \\
\midrule
BioRED, tmVar v1, tmVar v2, tmVar v3 & 4 & 87 \\
\midrule
BioRED, tmVar v1, tmVar v3 & 3 & 109 \\
\midrule
NLM Gene, BioRED & 2 & 140 \\
\midrule
BC5CDR, BioRED & 2 & 203 \\
\midrule
tmVar v1, tmVar v3 & 2 & 232 \\
\midrule
MLEE, BioNLP ST 2013 CG, AnatEM & 3 & 250 \\
\midrule
BioNLP ST 2013 CG, AnatEM & 2 & 348 \\
\midrule
AnatEM, AnEM & 2 & 492 \\
\midrule
GENIA Relation Corpus, BioNLP Shared Task 2009, BioNLP ST 2011 REL, BioNLP ST 2011 GE & 4 & 1179 \\
\midrule
ChemProt, CHEMDNER & 2 & 1199 \\
    \bottomrule
    \end{tabular}
  \label{tab:pubmed-cluster-counts}
\end{table}

As biomedical models are trained and evaluated on ever larger meta-datasets, it is important to characterize duplication within and between datasets.
This can take the form of direct train/test leakage \citesupp{elangovan-etal-2021-memorization} or more subtle issues of near-duplicates and repeated substrings which can negatively impact performance and training time of language models \citesupp{lee2021deduplicating}.
In biomedical NLP, annotation efforts often build upon existing datasets meaning meta-dataset curation needs to take additional steps to mitigate possible train/test leakage.
To assess the magnitude of this phenomena across the \ours~ corpus, we conducted a preliminary analysis counting the number of shared documents across all annotated datasets sourced from PubMed or PubMed Central (PMC). 

\paragraph{PubMed Document ID Normalization} PubMed/PMC provides uniform identifiers for documents: PubMed PMID and PubMed Central PMCID. 
However, many datasets encode this document information using inconsistent formats as shown in Table \ref{tab:pubmed-norm-sample}. 
We wrote a normalization function to standardize all document identifiers to facilitate joins with other PubMed/PMC datsets.
We then joined this data with the \texttt{PMC-ids.csv.gz} file available from the National Library of Medicine\footnote{\url{https://ftp.ncbi.nlm.nih.gov/pub/pmc} accessed May 29, 2022}.

\paragraph{PubMed Dataset Overlap Analysis}
Our normalizations of PMIDs allowed us to calculate which PubMed articles were used in multiple datasets.
In Table \ref{tab:pubmed-cluster-counts} we show the largest PMID clusters, i.e., sets of datasets that contain the same documents.
In Figure \ref{fig:pubmed-heatmap} we visualize this overlap as a heatmap.
We observe several cases of clear dataset iteration (e.g., tmVar v1-v3, AnEM to AnatEM) and NLP challenges building on the same source datasets (BioNLP shared tasks 2009 and 2011 build on the GENIA Relation Corpus). 
BioRED illustrates another common pattern, where documents were sampled from 5 existing biomedical datasets before annotating \citesupp{luo2022biored}. 



%% file: 18_appx_dataviz.tex
\clearpage
\section{Data Visualization and Exploration} 

To highlight the efficiency of using consistent schema across datasets, we created a Streamlit\footnote{https://streamlit.io/} web application to allow anyone to browse through any schema-specific details and visualization for all supported datasets. The web application enables task sorting at the level of task schema (e.g., NER), which supports downstream approaches to use groups of datasets with minimal effort. Such as prompt based methods or multi-task learning (MTL). 

\begin{figure}[ht!]
  \centering
  \includegraphics[scale=0.50]{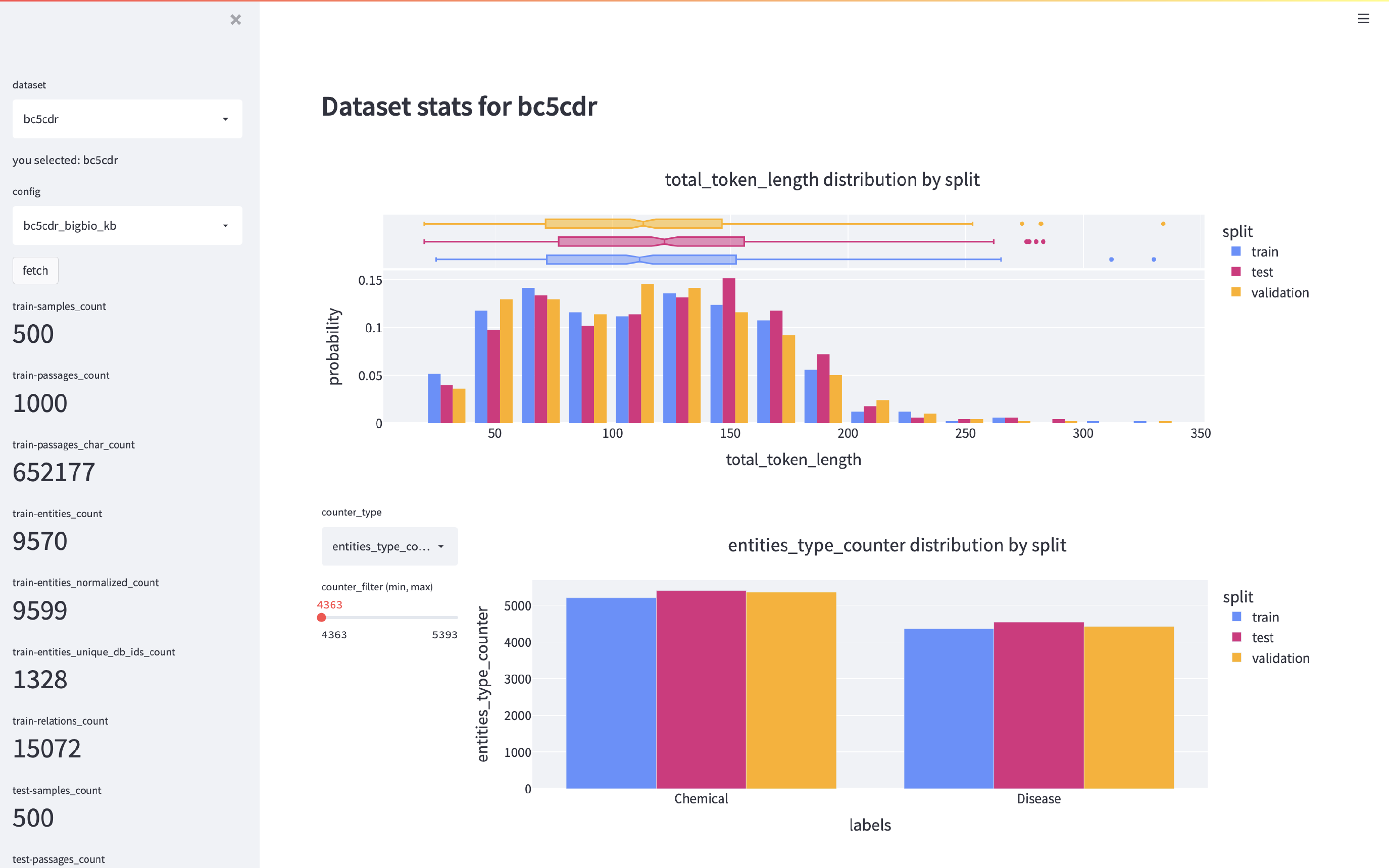}
  \caption{\label{fig:streamlit}Streamlit web application for visualizing dataset-specific details and textual analysis at the span-level. Here we show plots for the BioCreative V Chemical Disease Relation (CDR) dataset. }
\end{figure}

For each split, we provide basic dataset details (like number of training samples, character counts, word counts, number of unique labels, etc.).
Further, we also present distributions of token lengths and labels (or sub-component types) within each dataset to compare across splits.
We used periods and new lines to break the text block into sentences, and tokenized each sentence by white space to count the token lengths.
For datasets of tasks that do not have labels, which is the case for most common knowledge base construction and information extraction tasks, we analyze the data distribution across the sub-component types.
For instance, our task schema for the BioCreative V Chemical Disease Relation (CDR) dataset \citesupp{DBLP:journals/biodb/LiSJSWLDMWL16} provides an efficient way to compare the distribution of chemical and disease entities across splits (See Figure \ref{fig:streamlit}).


%% file: 19_appx_ml_zero_shot.tex
\clearpage
\section{Zero-shot Language Model Evaluation}
\label{sec:zeroshotsupp}

\subsection{Expanded Results}
Table \ref{tbl:allresults} contains complete zero-shot language model results pooled across all prompts (n=5) by dataset.

\begin{table}[ht!]
\small
\centering
\caption{\label{tbl:allresults}Summary results across all datasets and language models.}
\begin{tabular}{llccccc}
	\toprule
	Model                            & Dataset & Metric   & Mean & SE   & Min   & Max  \\
	\midrule
	SciFive-Base  & BIOSSES & pearson  & 34.0 & 14.6 & 12.8  & 55.8 \\
	SciFive-Large & BIOSSES & pearson  & 7.2  & 12.2 & -17.2 & 19.5 \\
	GPT-Neo-1.3B          & BIOSSES & pearson  & 36.4 & n/a  & 36.4  & 36.4 \\
	GPT-2                             & BIOSSES & pearson  & 12.5 & 7.2  & 5.3   & 19.5 \\
	GPT-J-6B              & BIOSSES & pearson  & 0.2  & 31.9 & -31.8 & 32.1 \\
	T5 v1.1-xxl              & BIOSSES & pearson  & n/a  & n/a  & n/a   & n/a  \\
	T0\_3B                & BIOSSES & pearson  & n/a  & n/a  & n/a   & n/a  \\
	T0                    & BIOSSES & pearson  & 23.3 & 17.6 & -7.2  & 49.5 \\
	T0+                   & BIOSSES & pearson  & 37.8 & 20.1 & 18.7  & \textbf{66.7} \\
	T0++                  & BIOSSES & pearson  & \textbf{40.6} & 1.3  & \textbf{38.7}  & 42.5 \\
	\midrule
	SciFive-Base  & BioASQ  & accuracy & 32.9 & 0.0  & 32.9  & 32.9 \\
	SciFive-Large & BioASQ  & accuracy & 32.9 & 0.0  & 32.9  & 32.9 \\
	GPT-Neo-1.3B          & BioASQ  & accuracy & 40.9 & 6.2  & 33.6  & 65.7 \\
	GPT-2                             & BioASQ  & accuracy & 36.1 & 3.1  & 32.9  & 48.6 \\
	GPT-J-6B              & BioASQ  & accuracy & 40.4 & 6.7  & 33.6  & 67.1 \\
	T5 v1.1-xxl              & BioASQ  & accuracy & 67.1 & 0.0  & 67.1  & 67.1 \\
	T0\_3B                & BioASQ  & accuracy & 40.1 & 0.6  & 38.6  & 42.1 \\
	T0                    & BioASQ  & accuracy & 76.1 & 2.3  & 70.7  & 82.9 \\
	T0+                   & BioASQ  & accuracy & 73.1 & 2.4  & 65.7  & 78.6 \\
	T0++                  & BioASQ  & accuracy & \textbf{89.0} & 1.3  & \textbf{84.3}  & \textbf{91.4} \\
	\midrule
	SciFive-Base  & SciTail & accuracy & 59.9 & 0.3  & 58.8  & 60.4 \\
	SciFive-Large & SciTail & accuracy & 56.2 & 4.1  & 39.6  & 60.4 \\
	GPT-Neo-1.3B          & SciTail & accuracy & 50.6 & 4.0  & 38.9  & 60.4 \\
	GPT-2                             & SciTail & accuracy & 50.3 & 4.3  & 39.6  & 60.4 \\
	GPT-J-6B              & SciTail & accuracy & 51.6 & 4.4  & 40.2  & 60.3 \\
	T5 v1.1-xxl              & SciTail & accuracy & 43.8 & 4.2  & 39.6  & 60.4 \\
	T0\_3B                & SciTail & accuracy & 68.9 & 4.7  & 55.0  & 77.6 \\
	T0                    & SciTail & accuracy & 73.9 & 6.2  & 58.1  & 88.1 \\
	T0+                   & SciTail & accuracy & 74.3 & 7.6  & 51.5  & 87.9 \\
	T0++                  & SciTail & accuracy & \textbf{75.6} & 5.9  & \textbf{60.4}  & \textbf{90.8} \\
	\midrule
	SciFive-Base  & MedNLI  & accuracy & 66.4 & 0.1  & 66.2  & 66.7 \\
	SciFive-Large & MedNLI  & accuracy & 66.7 & 0.0  & 66.5  & 66.7 \\
	GPT-Neo-1.3B          & MedNLI  & accuracy & 36.6 & 1.5  & 33.6  & 41.0 \\
	GPT-2                             & MedNLI  & accuracy & 55.1 & 5.9  & 33.3  & 65.6 \\
	GPT-J-6B              & MedNLI  & accuracy & 48.3 & 3.7  & 42.2  & 62.7 \\
	T5 v1.1-xxl              & MedNLI  & accuracy & 33.3 & 0.0  & 33.3  & 33.3 \\
	T0\_3B                & MedNLI  & accuracy & 67.6 & 0.3  & 66.6  & 68.3 \\
	T0                    & MedNLI  & accuracy & 72.0 & 1.6  & 68.8  & \textbf{77.8} \\
	T0+                   & MedNLI  & accuracy & 72.5 & 1.8  & 68.6  & 76.8 \\
	T0++                  & MedNLI  & accuracy & \textbf{73.4} & 1.5  & \textbf{69.1}  & 77.4 \\
	\midrule
	SciFive-Base  & GAD     & accuracy & 47.4 & 0.0  & 47.4  & 47.4 \\
	SciFive-Large & GAD     & accuracy & 47.4 & 0.0  & 47.4  & 47.4 \\
	GPT-Neo-1.3B          & GAD     & accuracy & 47.7 & 0.7  & 46.4  & 50.4 \\
	GPT-2                             & GAD     & accuracy & 47.4 & 0.0  & 47.4  & 47.6 \\
	GPT-J-6B              & GAD     & accuracy & 48.2 & 1.0  & 46.6  & 52.1 \\
	T5 v1.1-xxl              & GAD     & accuracy & 52.6 & 0.0  & 52.6  & 52.6 \\
	T0\_3B                & GAD     & accuracy & 47.5 & 0.1  & 47.4  & 47.8 \\
	T0                    & GAD     & accuracy & 53.7 & 1.0  & 50.7  & 55.6 \\
	T0+                   & GAD     & accuracy & 53.9 & 0.4  & 53.0  & 55.1 \\
	T0++                  & GAD     & accuracy & \textbf{55.7} & 0.4  & \textbf{54.3 } & \textbf{56.6} \\
	\bottomrule
\end{tabular}
\end{table}

\begin{figure}[ht!]
  \centering
  \includegraphics[scale=0.38]{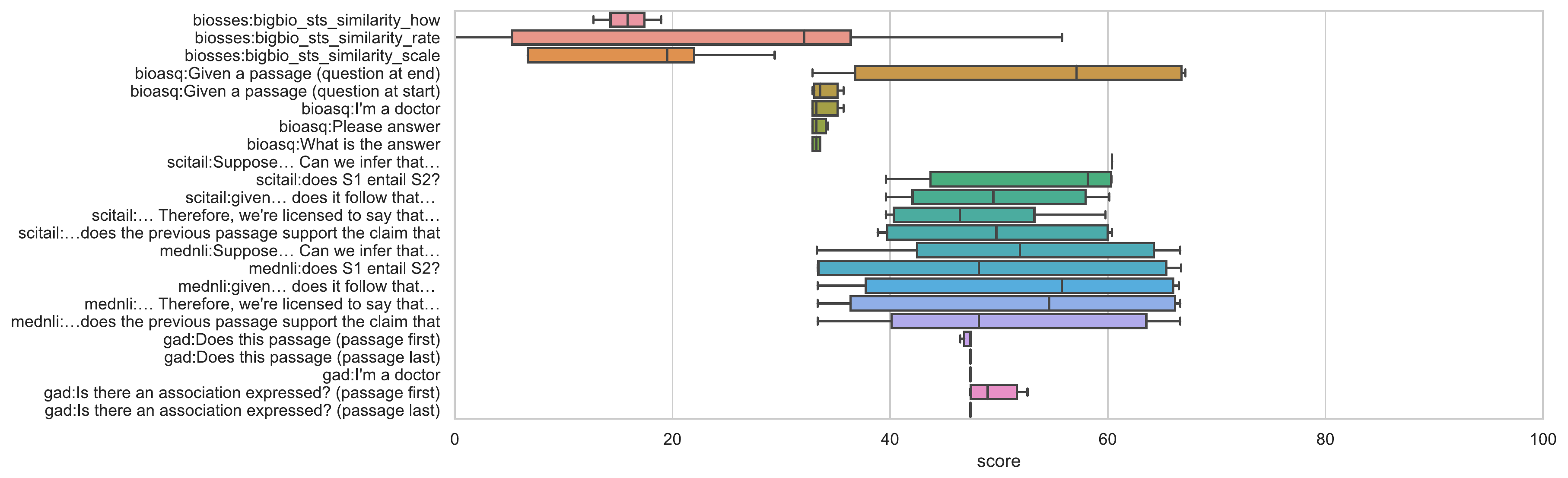}
  \caption{\label{fig:nott0prompts}Per-prompt scores (x-axis) for all non-T0 family language models (SciFive, GPT-Neo-1.3B, GPT-2, GPT-J-6B, T5 v1.1-xxl). Prompt template names are on the y-axis.}
\end{figure}

\begin{figure}[ht!]
  \centering
  \includegraphics[scale=0.38]{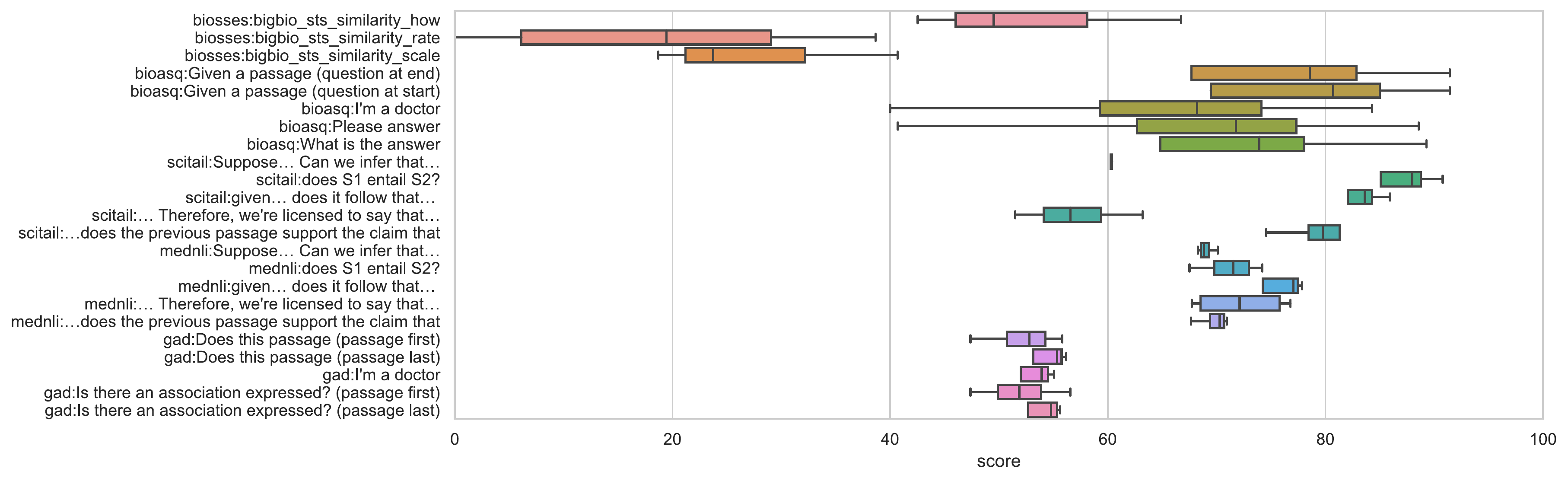}
  \caption{\label{fig:t0prompts}Per-prompt scores (x-axis)for all T0 family language models (T0\_3B, T0, T0+, T0++). Prompt template names are on the y-axis. Performance for non semantic similarity tasks is more varied and higher performing compared to current GPT-2 style pretrained models or T5 models with standard pretraining and in-domain finetuning. }
\end{figure}

\subsection{Evaluation}
All language models summary statistics are calculated using n=5 samples (1 score per prompt). 
Standard error is calculated using the sample standard deviation.
Pearson's Correlation was calculated using SciPy v1.7.3.
All other metrics are calculated calculated using Scikit-learn v1.0.2.
All models are evaluated using fp32 precision on a single 8x A40 compute node running CUDA 11.2.

\subsection{Code}

All experiment were run using the most up-to-date version of \ours~ before paper submssion. \url{https://github.com/bigscience-workshop/biomedical} commit {\tt 0ff295b25bb1be813e64f13246090bff6168cb5a }

Complete language model evaluation harness code and instructions for running \ours~ experiments: \url{https://github.com/bigscience-workshop/lm-evaluation-harness/tree/bigbio}

For these experiments, we used a modified version of PromptSource: \url{https://github.com/bigscience-workshop/promptsource/tree/eval-hackathon}
Prompt templates are available at \url{https://github.com/OpenBioLink/promptsource} and are outlined below.

All pretrained language models were downloaded from Hugging Face's datasets hub.

\subsection{Prompt Templates}

The following prompt templates were developed using PromptSource.
A prompt consists of a set of answer choices, an input template, and an output template. 

\subsubsection{BIOSSES}

\begin{table}[ht!]
\caption{BIOSSES example instance.}
\begin{tabular}{ll}
\toprule
         Key &                                                             Value \\
\midrule
          id &                                                           {\tt 6} \\
document\_id &                                                           {\tt 7} \\
     text\_1 & {\tt Recently, it was reported that expression of IDH1R132H su...} \\
     text\_2 & {\tt the mechanism was clarified by yet another genomic survey...} \\
       label &                                                         {\tt 1.6} \\
\bottomrule
\end{tabular}
\end{table}

\paragraph{Prompt 1: ``bigbio\_sts\_similarity\_scale"}
\paragraph{Answer Choices:} \
\begin{code}
0 ||| 1 ||| 2 ||| 3 ||| 4
\end{code}
\paragraph{Input Template:} \
\begin{code}
from {{"0"}} to {{"4"}}, how similar are "{{text_1}}" and "{{text_2}}"?
\end{code}
\paragraph{Output Template:} \
\begin{code}
{{label}}
\end{code}
\par\noindent\rule{\textwidth}{0.4pt}

\paragraph{Prompt 2: ``bigbio\_sts\_similarity\_how"}
\paragraph{Answer Choices:} \
\begin{code}
0 ||| 1 ||| 2 ||| 3 ||| 4
\end{code}
\paragraph{Input Template:} \
\begin{code}
How similar are "{{text_1}}" and "{{text_2}}"? Give a score \ 
between {{"0"}} and {{"4"}}.
\end{code}
\paragraph{Output Template:} \
\begin{code}
{{label}}
\end{code}
\par\noindent\rule{\textwidth}{0.4pt}

\paragraph{Prompt 3: ``bigbio\_sts\_similarity\_rate"}
\paragraph{Answer Choices:} \
\begin{code}
0 ||| 1 ||| 2 ||| 3 ||| 4
\end{code}
\paragraph{Input Template:} \
\begin{code}
Rate the similarity of these two sentences ({{"0"}} being the lowest \
and {{"4"}} the highest):
"{{text_1}}" and "{{text_2}}"
\end{code}
\paragraph{Output Template:} \
\begin{code}
{{label}}
\end{code}
\par\noindent\rule{\textwidth}{0.4pt}

\paragraph{Prompt 4: ``bigbio\_sts\_similarity\_on\_a\_scale"}
\paragraph{Answer Choices:} \
\begin{code}
0 ||| 1 ||| 2 ||| 3 ||| 4
\end{code}
\paragraph{Input Template:} \
\begin{code}
On a scale of {{"0"}} (completely unrelated) to {{"4"}} (exactly same) \ 
score these sentences: 
"{{text_1}}" and "{{text_2}}"
\end{code}
\paragraph{Output Template:} \
\begin{code}
{{label}}
\end{code}
\par\noindent\rule{\textwidth}{0.4pt}

\paragraph{Prompt 5: ``bigbio\_sts\_similarity\_what\_is"}
\paragraph{Answer Choices:} \
\begin{code}
0 ||| 1 ||| 2 ||| 3 ||| 4
\end{code}
\paragraph{Input Template:} \
\begin{code}
What is the similarity of these two sentences on a scale of {{"0"}} (low) \ 
to {{"4"}} (high): "{{text_1}}" and "{{text_2}}"
\end{code}
\paragraph{Output Template:} \
\begin{code}
{{label}}
\end{code}

\subsubsection{BioASQ}

\begin{table}[ht!]
\caption{BioASQ example instance.}
\begin{tabular}{ll}
\toprule
         Key &                                                             Value \\
\midrule
          id &                                 {\tt 5c58a74e86df2b917400000d\_0} \\
question\_id &                                    {\tt 5c58a74e86df2b917400000d} \\
document\_id &                 {\tt http://www.ncbi.nlm.nih.gov/pubmed/29623652} \\
    question &                       {\tt Is Baloxavir effective for influenza?} \\
        type &                                                       {\tt yesno} \\
     choices &                                                          {\tt []} \\
     context & {\tt Baloxavir marboxil (Xofluza™; baloxavir) is an oral cap-d...} \\
      answer &                                                     {\tt [`yes']} \\
\bottomrule
\end{tabular}
\end{table}

\paragraph{Prompt 1: ``Given a passage (question at end)"}
\paragraph{Answer Choices:} \
\begin{code}
no ||| yes
\end{code}
\paragraph{Input Template:} \
\begin{code}
Given a passage: {{ context }}

Answer the question: "{{question}}"

\end{code}
\paragraph{Output Template:} \
\begin{code}
 
{{answer[0]}}
\end{code}
\par\noindent\rule{\textwidth}{0.4pt}

\paragraph{Prompt 2: ``I'm a doctor"}
\paragraph{Answer Choices:} \
\begin{code}
no ||| yes
\end{code}
\paragraph{Input Template:} \
\begin{code}
I'm a doctor and I need to answer the question "{{ question }}" using \ 
the following passage:

{{ context }}

\end{code}
\paragraph{Output Template:} \
\begin{code}
 
{{answer[0]}}
\end{code}
\par\noindent\rule{\textwidth}{0.4pt}

\paragraph{Prompt 3: ``What is the answer"}
\paragraph{Answer Choices:} \
\begin{code}
no ||| yes
\end{code}
\paragraph{Input Template:} \
\begin{code}
What is the answer to the question "{{ question }}" based on \ 
the following passage:

{{ context }}

\end{code}
\paragraph{Output Template:} \
\begin{code}
 
{{answer[0]}}
\end{code}
\par\noindent\rule{\textwidth}{0.4pt}

\paragraph{Prompt 4: ``Please answer"}
\paragraph{Answer Choices:} \
\begin{code}
no ||| yes
\end{code}
\paragraph{Input Template:} \
\begin{code}
Please answer the question "{{ question }}" using \ 
the following passage:

{{ context }}

\end{code}
\paragraph{Output Template:} \
\begin{code}
 
{{answer[0]}}
\end{code}
\par\noindent\rule{\textwidth}{0.4pt}

\paragraph{Prompt 5: ``Given a passage (question at start)"}
\paragraph{Answer Choices:} \
\begin{code}
no ||| yes
\end{code}
\paragraph{Input Template:} \
\begin{code}
Given the following passage, answer the question: "{{question}}"

Passage: {{ context }}

\end{code}
\paragraph{Output Template:} \
\begin{code}
 
{{answer[0]}}
\end{code}

\subsubsection{SciTail}

\begin{table}[ht!]
\caption{SciTail example instance.}
\begin{tabular}{ll}
\toprule
       Key &                                                             Value \\
\midrule
        id &                                                           {\tt 0} \\
   premise & {\tt Based on the list provided of the uses of substances 1-7,...} \\
hypothesis & {\tt If a substance has a ph value greater than 7,that indicat...} \\
     label &                                                     {\tt neutral} \\
\bottomrule
\end{tabular}
\end{table}

\paragraph{Prompt 1: ``... Therefore, we're licensed to say that..."}
\paragraph{Answer Choices:} \
\begin{code}
true ||| false
\end{code}
\paragraph{Input Template:} \
\begin{code}
{{premise}} Therefore, we are licensed to say that {{hypothesis}}  {{ answer_choices | join(' or ') }}
\end{code}
\paragraph{Output Template:} \
\begin{code}

{
{{answer_choices[0]}}
{
{{answer_choices[1]}}
{
\end{code}
\par\noindent\rule{\textwidth}{0.4pt}

\paragraph{Prompt 2: ``Suppose... Can we infer that..."}
\paragraph{Answer Choices:} \
\begin{code}
neutral ||| entailment
\end{code}
\paragraph{Input Template:} \
\begin{code}
Suppose {{premise}} Can we infer that {{hypothesis}}? 
\end{code}
\paragraph{Output Template:} \
\begin{code}
 {{label}}
\end{code}
\par\noindent\rule{\textwidth}{0.4pt}

\paragraph{Prompt 3: ``...does the previous passage support the claim that"}
\paragraph{Answer Choices:} \
\begin{code}
yes ||| no
\end{code}
\paragraph{Input Template:} \
\begin{code}
{{premise}} Does the previous passage support the claim that {{hypothesis}}? 
\end{code}
\paragraph{Output Template:} \
\begin{code}
{
{{answer_choices[0]}}
{
{{answer_choices[1]}}
{
\end{code}
\par\noindent\rule{\textwidth}{0.4pt}

\paragraph{Prompt 4: ``given... does it follow that..."}
\paragraph{Answer Choices:} \
\begin{code}
yes ||| no
\end{code}
\paragraph{Input Template:} \
\begin{code}
Given that {{premise}} Does it follow that {{hypothesis}}  {{ answer_choices | join(' or ') }} 
\end{code}
\paragraph{Output Template:} \
\begin{code}

{
{{answer_choices[0]}}
{
{{answer_choices[1]}}
{
\end{code}
\par\noindent\rule{\textwidth}{0.4pt}

\paragraph{Prompt 5: ``does S1 entail S2?"}
\paragraph{Answer Choices:} \
\begin{code}
yes ||| no
\end{code}
\paragraph{Input Template:} \
\begin{code}
Sentence 1: {{premise}}

Sentence 2: {{hypothesis}}

Question: Does Sentence 1 entail Sentence 2?  \
{{ answer_choices | join(' or ') }} 
\end{code}
\paragraph{Output Template:} \
\begin{code}

{
{{answer_choices[0]}}
{
{{answer_choices[1]}}
{
\end{code}

\subsubsection{MedNLI}

\begin{table}[ht!]
\caption{MedNLI example instance.}
\begin{tabular}{ll}
\toprule
       Key &                                                             Value \\
\midrule
        id &                        {\tt 1f2a8146-66c7-11e7-b4f2-f45c89b91419} \\
   premise & {\tt In the ED, initial VS revealed T 98.9, HR 73, BP 121/90, ...} \\
hypothesis &                     {\tt  The patient is hemodynamically stable } \\
     label &                                                  {\tt entailment} \\
\bottomrule
\end{tabular}
\end{table}

\paragraph{Prompt 1: ``... Therefore, we're licensed to say that..."}
\paragraph{Answer Choices:} \
\begin{code}
yes ||| no
\end{code}
\paragraph{Input Template:} \
\begin{code}
{{premise}} Therefore, we are licensed to say that {{hypothesis}}  {{ answer_choices | join(' or ') }}
\end{code}
\paragraph{Output Template:} \
\begin{code}

{
{{answer_choices[0]}}
{
{{answer_choices[1]}}
{
\end{code}
\par\noindent\rule{\textwidth}{0.4pt}

\paragraph{Prompt 2: ``Suppose... Can we infer that..."}
\paragraph{Answer Choices:} \
\begin{code}
yes ||| no
\end{code}
\paragraph{Input Template:} \
\begin{code}
Suppose {{premise}} Can we infer that {{hypothesis}}? 
\end{code}
\paragraph{Output Template:} \
\begin{code}

{
{{answer_choices[0]}}
{
{{answer_choices[1]}}
{
\end{code}
\par\noindent\rule{\textwidth}{0.4pt}

\paragraph{Prompt 3: ``...does the previous passage support the claim that"}
\paragraph{Answer Choices:} \
\begin{code}
yes ||| no
\end{code}
\paragraph{Input Template:} \
\begin{code}
{{premise}} Does the previous passage support the claim that {{hypothesis}}?

\end{code}
\paragraph{Output Template:} \
\begin{code}
{
{{answer_choices[0]}}
{
{{answer_choices[1]}}
{
\end{code}
\par\noindent\rule{\textwidth}{0.4pt}

\paragraph{Prompt 4: ``given... does it follow that..."}
\paragraph{Answer Choices:} \
\begin{code}
yes ||| no
\end{code}
\paragraph{Input Template:} \
\begin{code}
Given that {{premise}} Does it follow that {{hypothesis}} \ 
{{ answer_choices | join(' or ') }} 
\end{code}
\paragraph{Output Template:} \
\begin{code}

{
{{answer_choices[0]}}
{
{{answer_choices[1]}}
{
\end{code}
\par\noindent\rule{\textwidth}{0.4pt}

\paragraph{Prompt 5: ``does S1 entail S2?"}
\paragraph{Answer Choices:} \
\begin{code}
yes ||| no
\end{code}
\paragraph{Input Template:} \
\begin{code}
Sentence 1: {{premise}}

Sentence 2: {{hypothesis}}

Question: Does Sentence 1 entail Sentence 2? \ 
{{ answer_choices | join(' or ') }} 
\end{code}
\paragraph{Output Template:} \
\begin{code}

{
{{answer_choices[0]}}
{
{{answer_choices[1]}}
{
\end{code}

\subsubsection{GAD}

\begin{table}[ht!]
\caption{GAD example instance.}
\begin{tabular}{ll}
\toprule
         Key &                                                             Value \\
\midrule
          id &                                                           {\tt 0} \\
document\_id &                                                           {\tt 0} \\
        text & {\tt These results suggest that the C1772T polymorphism in @GE...} \\
      labels &                                                       {\tt [`1']} \\
\bottomrule
\end{tabular}
\end{table}

\paragraph{Prompt 1: ``Does this passage (passage last)"}
\paragraph{Answer Choices:} \
\begin{code}
No ||| Yes
\end{code}
\paragraph{Input Template:} \
\begin{code}
Does the following passage indicate that there is an association \ 
between the gene @GENE$ and the disease @DISEASE$ ?

{{ text }}

\end{code}
\paragraph{Output Template:} \
\begin{code}
{{ answer_choices[labels[0] | int] }}
\end{code}
\par\noindent\rule{\textwidth}{0.4pt}

\paragraph{Prompt 2: ``Does this passage (passage first)"}
\paragraph{Answer Choices:} \
\begin{code}
No ||| Yes
\end{code}
\paragraph{Input Template:} \
\begin{code}
{{ text }}

Does this passage indicate that there is an association between the \ 
gene @GENE$ and the disease @DISEASE$ ?

\end{code}
\paragraph{Output Template:} \
\begin{code}
{{ answer_choices[labels[0] | int] }}
\end{code}
\par\noindent\rule{\textwidth}{0.4pt}

\paragraph{Prompt 3: ``Is there an association expressed? (passage last)"}
\paragraph{Answer Choices:} \
\begin{code}
No ||| Yes
\end{code}
\paragraph{Input Template:} \
\begin{code}
Is there an association between the gene @GENE$ and the disease \ 
@DISEASE$ expressed in this passage?

{{ text }}

\end{code}
\paragraph{Output Template:} \
\begin{code}
{{ answer_choices[labels[0] | int] }}
\end{code}
\par\noindent\rule{\textwidth}{0.4pt}

\paragraph{Prompt 4: ``I'm a doctor"}
\paragraph{Answer Choices:} \
\begin{code}
No ||| Yes
\end{code}
\paragraph{Input Template:} \
\begin{code}
I'm a doctor. Can you tell me, is there an association between the \ 
gene @GENE$ and the disease @DISEASE$ expressed in this passage?

{{ text }}

\end{code}
\paragraph{Output Template:} \
\begin{code}

{{ answer_choices[labels[0] | int] }}
\end{code}
\par\noindent\rule{\textwidth}{0.4pt}

\paragraph{Prompt 5: ``Is there an association expressed? (passage first)"}
\paragraph{Answer Choices:} \
\begin{code}
No ||| Yes
\end{code}
\paragraph{Input Template:} \
\begin{code}
{{ text }}

Is there an association between the gene @GENE\$ and the disease \ 
@DISEASE$ expressed in this passage?

\end{code}
\paragraph{Output Template:} \
\begin{code}

{{ answer_choices[labels[0] | int] }}
\end{code}

%% file: 20_appx_ml_mtl.tex
\clearpage
\section{Large-scale Multi-Task Learning}
\label{sec:mtlsupp}
We make the MTL model available at \url{https://huggingface.co/bigscience-biomedical/bigbio-mtl}.
Code and instructions to reproduce our results can be found in \url{https://github.com/leonweber/biomuppet}.

\begin{table}[htbp]
	\centering
	\caption{MTL dataset statistics}
	\begin{tabular}{llrrr}
		\toprule
		Task                        & Abbrev. & \# Train Examples & \# Valid Examples & \# Datasets \\
		\midrule
		Relation Extraction         & RE      & 656,171          & 106,519           & 14         \\
		Coreference Resolution      & COREF   & 113,137          & 35,030            & 9          \\
		Event Argument Extraction   & EAE     & 294,129          & 119,033           & 10         \\
		Text Classification         & CLASS   & 30,743           & 3,416             & 2          \\
		Semantic Textual Similarity & STS     & 7,215            & 804               & 6          \\
		Question Answering          & QA      & 6,490            & 561               & 2          \\
		Named Entity Recognition    & NER     & 287,582          & 89,135            & 53         \\
		Event Detection             & ED      & 28,388           & 9,883             & 10         \\
		\midrule
		Total                       &         & 1,423,855        & 364,381           & 106        \\
		\bottomrule
	\end{tabular}
	\label{tab:mtl-data-statistics}
\end{table}

\subsection{Conversion to MaChAmp}
We generated training and evaluation data for 106 datasets that were available when we started to develop the MTL project source code (\ours~version found  \href{https://github.com/bigscience-workshop/biomedical/commit/dbd7f1646280e7ebb5adb201c504bb433d9067c2}{here}).
If a dataset within this collective set did not have a predefined validation split, we reserved 10\% of its training data as the validation set. Each dataset also had one \ours-to-MaChAmp transformation script per \ours~task. The purpose of this transformation script is to convert the data represented in the \ours-schema in a MaChAmp-compatible input for simple extension to the ML library.
For statistics of the resulting data set Table~\ref{tab:mtl-data-statistics} and for examples of the transformed task data see Tables~\ref{tab:mtl-examples-class} and \ref{tab:mtl-examples-seq}. 
%

%
We model \textbf{Relation Extraction} (RE) as relation classification. Each sentence in an input passage is split; subsequently, we construct on example per entity-pair by introducing special marker tokens to mark the start and end of each head and tail entity. We consider each example as a text classification problem in MaChAmp, where the goal is to predict the type of relation between the marked head/ail entities, including a `None' type relation.
We follow the BLURB preprocessing strategy for RE and replace the strings of the marked head and tail entity with their respective entity type.
For multi-label datasets where an entity or relation may possess multiple labels, we transform such cases to a multiclass dataset by concatenating all labels.
We use this multilabel-to-multiclass transformation for all task types, if required.
%

%
We treat \textbf{Coreference Resolution} (COREF) in a similar fashion as RE, with the only difference that we have only two relation types: `coref' denoting a coreference relation between two token spans and `None'.
We transform the \textbf{Event Argument Extraction} (EAE) data in exactly the same way as RE, with the trigger span acting as the head entity and all possible event arguments (entities and triggers) acting as tail entities. 
For \textbf{Text Classification} (CLASS), we adapt the \ours~version to the MaChAmp format without any further modification apart from the multilabel-to-multiclass transformation. 
We transform the \textbf{Semantic Textual Similarity} (STS) task from a regression task to classification by replacing the STS score with the decantile into which it falls. 
We use the template `Text1 [SEP] Text2` where `Text1' and `Text2' are either words, sentences or paragraphs depending on the dataset.
We model \textbf{Named Entity Recognition} (NER) and Event Detection (ED) as sequence labelling tasks using an IOB-tagging scheme after sentence splitting.
For \textbf{Question Answering} (QA), we experimented with two formulations.
In the classification formulation, we construct one example per answer candidate by using the template `Context [SEP] Question [SEP] AnswerCandidate` and the two labels `True' (if `AnswerCandidate' is the correct answer) and `False' (if `AnswerCandidate' is the wrong answer).
In the sequence labelling setting, we use the template `Context. Question` and mark all tokens in occurrences of the answer in `Context' with `answer' and the rest with `O'.
We use Flair's~\citesupp{akbik-etal-2019-flair} `SegtokSentenceSplitter` for sentence splitting and `SpaceTokenizer` for tokenization.

\begin{table}[htbp]
  \centering
  \caption{Examples for the classification task formulation}
    \begin{tabular}{p{2cm}p{6cm}p{4.5cm}}
    \toprule
    Task Type  & Input & Label \\
    \midrule
    RE    & Taken together, these results make it clear that @chemical\$-bound forms of ORC and @protein\$ are likely to be required for productive interactions and pre-RC formation. & {\tt bind} \\
    & & \\
    COREF &  We investigated the potential of the @aryl hydrocarbon receptor\$ (@AHR\$) to suppress NF-kappaB regulated-gene expression, especially acute-phase genes, such as serum amyloid A (Saa). & {\tt coref} \\
    & & \\
    EAE   &  v-erbA @Gene\_expression\$ is required to @Negative\_regulation\$ c-erbA function in erythroid cell differentiation and regulation of the erbA target gene CAII. & cause \\
    & & \\
    CLASS &  These results are in contrast with the findings of Santos et al.(16), who reported a significant association between low sedentary time and healthy CVF among Portuguese & {\tt result\&supportive} \\
    & & \\
    STS   &  Renal failure [SEP] Kidney failure & {\tt 8} \\
    & & \\
    QA (class) & Cytokeratin 7/20 staining has been reported to be helpful [...] [SEP] Is cytokeratin immunoreactivity useful in the diagnosis of short-segment Barrett's oesophagus in Korea? & {\tt True} \\
   \bottomrule
    \end{tabular}
  \label{tab:mtl-examples-class}%
\end{table}%
    
\begin{table}[htbp]
  \centering
  \caption{Examples for the sequence labeling task formulation}
    \begin{tabular}{p{2cm}p{6cm}p{4.5cm}}
    \toprule
    Task Type  & Input & Label \\
    \midrule
    NER    & Tricuspid valve regurgitation and lithium carbonate toxicity in a newborn infant. & {\tt B-Disease I-Disease I-Disease O B-Chemical I-Chemical B-Disease O O O O} \\
    & & \\
    ED    & Coexpression of  NF-kappa B/Rel and Sp1 transcription factors in human immunodeficiency virus 1-induced, dendritic cell-T-cell syncytia. & {\tt B-Gene\_expression O O O O O O O O O O O O O O O} \\
    & & \\
    QA (seq) & a frameshift mutation is a deletion or insertion of one or more nucleotides [...] a frameshift mutation is a deletion or insertion of one or more of what that changes the reading frame of the base sequence ? & {\tt O O O O O [...] answer [...] O} \\
    \bottomrule
    \end{tabular}%
  \label{tab:mtl-examples-seq}%
\end{table}%


\subsection{Hyperparameters}
For hyperparameter choices, we use a mixture of the MaChAmp default hyperparameters and the suggestions from \citesupp{aghajanyan-etal-2021-muppet}.
We use AdamW~\citesupp{loshchilov2018decoupled} with a polynomial decay learning rate schedule with 50,000 warmup steps with a maximum learning rate of 1e-4.
We set weight decay to 0.01, dropout to 0.1 and the maximum length of the transformer to 512. 
We use an effective batch size of 32 tasks and 16 examples per task, train the model with Automated Mixed Precision set to fp16 using apex~(\url{https://github.com/NVIDIA/apex}) and clip the gradient norm to 5.
Finally, we downsample large datasets by using MaChAmp's multinomial sampling with alpha set to 0.5.

For model selection we evaluate the model after each epoch on all validation sets and select the model with the highest average accuracy.

\subsection{Results on Validation Sets}
\begin{figure}[ht!]
    \centering
    \includegraphics[width=\linewidth]{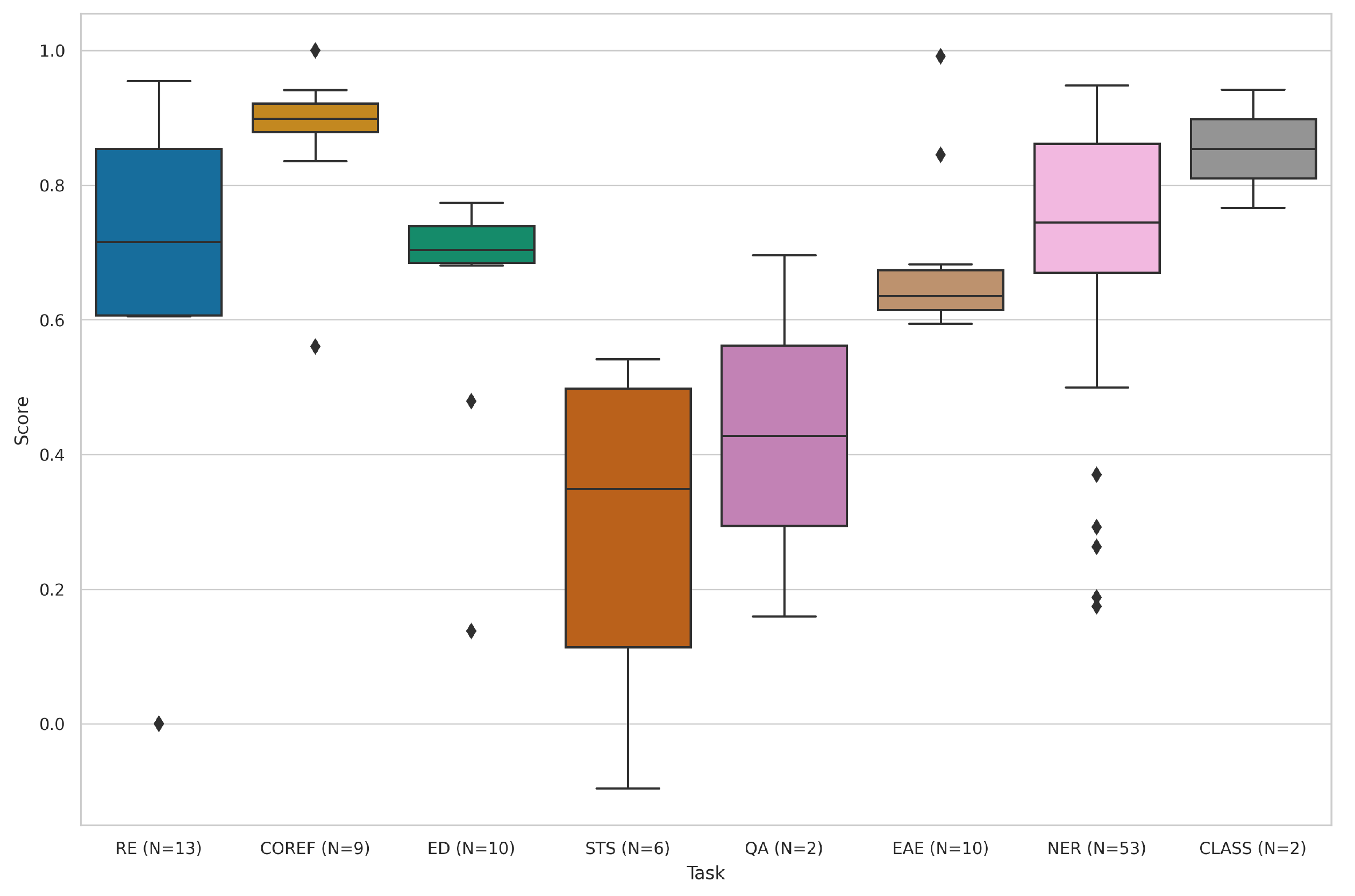}
    \caption{Validation set results of the MTL model by task. `RE' denotes Relation Extraction, `COREF' Coreference Resolution, `ED' event detection, `STS' Semantic Textual Similarity, `QA' Question Answering, `EAE' Event Argument Extraction, `NER' Named Entity Recognition, and `CLASS` Text Classification. Score is accuracy for QA and CLASS, Pearson's r for STS and F1 for the rest.}
    \label{fig:mtl-results}
\end{figure}

We evaluate our MTL model on all validation sets and deliberately refrain from evaluating on the test sets, because we did not rule out train/test overlap.
The validation results can be found in Figure~\ref{fig:mtl-results}.
Results vary strongly across task types, with the model performing well on COREF (mean 86.9\% F1), CLASS (mean 85.4 acc), and NER (mean 72.2\% F1).
Performance on STS (mean 28.1 Pearson's r) and QA (mean 42.8 acc) is surprisingly low. 
We attribute the weak performance on both STS and QA to the small amount of data per task (7,215 and 6,490 training examples respectively), which might prevent the model from allocating parameters for these tasks.

\subsection{Resources Used for Training}
We trained the MTL model on a local machine on four RTX 3090 GPUs.
Training for 50 epochs allowed the model to converge in all tested configurations and took roughly 33 hours.

%% file: 21_appx_existing_benchmarks.tex
\clearpage
\section{\ours~vs. Existing Benchmarks}

\paragraph{Biomedical Meta-dataset Benchmarks} Table~\ref{tbl:benchmarkftrs} compares \ours~against attributes of other popular English language biomedical meta-dataset benchmarks. 
To-date, our framework is the only one that supports API-based dataset access, providing access to 4x more datasets than the largest comparable meta-dataset. 
BLUE and BLURB do not provide dataset access via an API and require manual downloading and preprocessing.
Depending on the dataset, these preprocessing choices may not be easily reproducible. 
For example, in 4/5 NER tasks BLURB uses the IOB transformed datasets generated by Crichton et al.~\citesupp{crichton2017neural}.
These datasets rely on regular expression-based tokenization and sentence boundary detection methods developed by Crichton et al. and can vary by dataset, making it difficult to systemically the impact of different tokenization and sentence splitting choices.

End-to-end few and zero-shot evaluation of datasets, prompts, and pretrained language models is emerging as a standardized way to measure the performance of pretrained language models. 
BLUE and BLURB do not directly support prompt evaluation. 
BoX provides prompts for 32 biomedical datasets and Python tools for evaluating BART~\citesupp{lewis2019bart}-based language models, however BoX does not provide any access to the original datasets themselves. 
\ours~ integrates with the prompt engineering framework \textsc{PromptSource} to support users more easily designing prompts and running evaluations using the EleutherAI Language Model Evaluation Harness \cite{eval-harness}.
We currently support several seq2seq and causal language models (e.g., T5, T0, GPT families) available in Hugging Face's model hub. 
Currently \ours~ implements 25 prompts (5 datasets, 5 prompts), with future work focusing on constructing a library of task and dataset-specific biomedical prompts.

\begin{table}[ht!]
\centering
\caption{Attributes of existing English biomedical meta-dataset benchmarks}
\begin{tabular}{lccccccc}
\toprule
\multirow{2}{*}{Name} & \multirow{2}{*}{Datasets} & \multirow{2}{*}{Tasks} & \multirow{2}{*}{Langs} & \multirow{2}{*}{Data API} & \multirow{2}{2cm}{Reproducible Preprocessing} & \multirow{2}{*}{Prompts} & \multirow{2}{1.25cm}{\centering{Evaluation Harness}} \\ 
& & & & & & & \\
\toprule
\ours~                                        & 127      & 12    & 10        & \checkmark  & \checkmark                 & \textit{partial}  & \checkmark  \\
BLUE~\citesupp{peng-etal-2019-transfer}            & 10       & 5     & 1         &             & \textit{partial}                    &      &      \\
BLURB~\citesupp{DBLP:journals/health/GuTCLULNGP22} & 13       & 7     & 1         &             & \textit{partial}                    & &            \\
BoX~\citesupp{parmar2022boxbart}                   & 32       & 9     & 1         &             &                            & \checkmark & \checkmark \\
\bottomrule
\label{tbl:benchmarkftrs}
\end{tabular}
\end{table}

\begin{table}[ht!]
\centering
\caption{\ours~support of datasets used in popular meta-dataset benchmarks.}
\begin{tabular}{llccccc}
\toprule
Task Type & Dataset       & \ours & BLUE  & BLURB & BoX & DUA \\
\midrule
NER       & BC2GM         & \checkmark          &   & \checkmark  & \checkmark       &             \\
NER       & BC5-chem      & \checkmark          & \checkmark  & \checkmark  & \checkmark       &          \\
NER       & BC5-disease   & \checkmark          & \checkmark  & \checkmark  & \checkmark       &          \\
NER       & EBM PICO      & \checkmark          &   & \checkmark  &        &             \\
NER       & JNLPBA        & \checkmark          &   & \checkmark  & \checkmark       &             \\
NER       & NCBI-disease  & \checkmark          &   & \checkmark  & \checkmark       &          \\
RE        & ChemProt      & \checkmark          & \checkmark  & \checkmark  & \checkmark       &          \\
RE        & DDI           & \checkmark          & \checkmark  & \checkmark  & \checkmark       &          \\
RE        & GAD           & \checkmark          &   & \checkmark  &        &             \\
QA        & PubMedQA      & \checkmark          &   & \checkmark  &    \checkmark    &          \\
QA        & BioASQ        & \checkmark          &   & \checkmark  &  \checkmark       & \checkmark         \\
DC        & HoC           & \checkmark          & \checkmark  &   \checkmark  & \checkmark       &          \\
STS       & BIOSSES       & \checkmark          & \checkmark  &   \checkmark  &        &          \\
STS       & MedSTS        & $\ast$                & \checkmark  &   &        &   \checkmark          \\
NER       & n2c2 2010     & \checkmark          & \checkmark  &   &  \checkmark      & \checkmark         \\
NER       & ShARe/CLEF 2013   & $\ast$          & \checkmark  &   &        &   \checkmark          \\
NLI       & MedNLI        & \checkmark          & \checkmark  &   &        &    \checkmark         \\ 
NER        & n2c2 deid 2006  & \checkmark          &   &   & \checkmark       &    \checkmark           \\
DC       & n2c2 RFHD 2014     & \checkmark       &   &   & \checkmark       &   \checkmark           \\
NER       & AnatEM        & \checkmark          &   &   & \checkmark       &             \\
NER       & BC4CHEMD      & \checkmark          &   &   & \checkmark       &             \\
NER       & BioNLP09      & \checkmark          &   &   & \checkmark       &             \\
NER       & BioNLP11EPI   & \checkmark          &   &   & \checkmark       &             \\
NER       & BioNLP11ID    & \checkmark          &   &   & \checkmark       &             \\
NER       & BioNLP13CG    & \checkmark          &   &   & \checkmark       &             \\
NER       & BioNLP13GE    & \checkmark          &   &   & \checkmark       &             \\
NER       & BioNLP13PC    & \checkmark          &   &   & \checkmark       &             \\
NER       & CRAFT         & $\ast$                &   &   & \checkmark       &             \\
NER       & Ex-PTM        & \checkmark          &   &   & \checkmark       &             \\
NER       & Linnaeus      & \checkmark          &   &   & \checkmark       &             \\
POS       & GENIA         & $\ast$                &   &   & \checkmark       &             \\
SA        & Medical Drugs & \checkmark          &   &   & \checkmark       &  \\
\midrule
SR        & COVID         &          &   &   & \textit{private}       &             \\
SR        & Cooking       &          &   &   & \textit{private}      &             \\
SR        & HRT           &          &   &   & \textit{private}      &             \\
SR        & Accelerometer &          &   &   & \textit{private}       &             \\
SR        & Acromegaly    &          &   &   & \textit{private}      &             \\
\bottomrule
\multicolumn{3}{l}{\footnotesize{* denotes dataset implementation in-progress}} \\
\end{tabular}
\label{tbl:benchmarks}
\end{table}

\paragraph{Dataset Coverage} Table \ref{tbl:benchmarks} enumerates the list of datasets currently used by \ours~, BLUE, BLURB, and BoX.
Abbreviations are as follows: Named Entity Recognition (NER); Relation Extraction (RE); 
Question Answering (QA); Part-of-Speech Tagging (POS); Sentiment Analysis (SA) ; Natural Language Inference (NLI); and Systematic Review (SR). 
For the 32 public datasets \ours~ provides data loaders for the majority (28/32), while the remaining 4 are still being implemented by volunteers as of 06/16/2022.
Note that \textit{private} indicates that datasets are not available publicly or via DUA and thus cannot currently be included in \ours.

%% file: 22_appx_datasheets.tex
\clearpage
\section{Example Data Cards}
\label{sec:datasheets}
We generated data cards for all \ours~ datasets.
We include an example dataset from each schema type to illustrate data cards for different tasks. 
A PDF of all content is available on our project homepage.

\input{cantemist_cantemist_bigbio_text} 
\input{mediqa_qa_mediqa_qa_bigbio_qa} 
\input{an_em_an_em_bigbio_kb} 
\input{paramed_paramed_bigbio_t2t} 
\input{scitail_scitail_bigbio_te} 
\input{mqp_mqp_bigbio_pairs} 

%% file: cantemist_cantemist_bigbio_text.tex
\clearpage
\section*{Cantemist Data Card}
\begin{figure}[ht!]
\centering
\includegraphics[width=\linewidth]{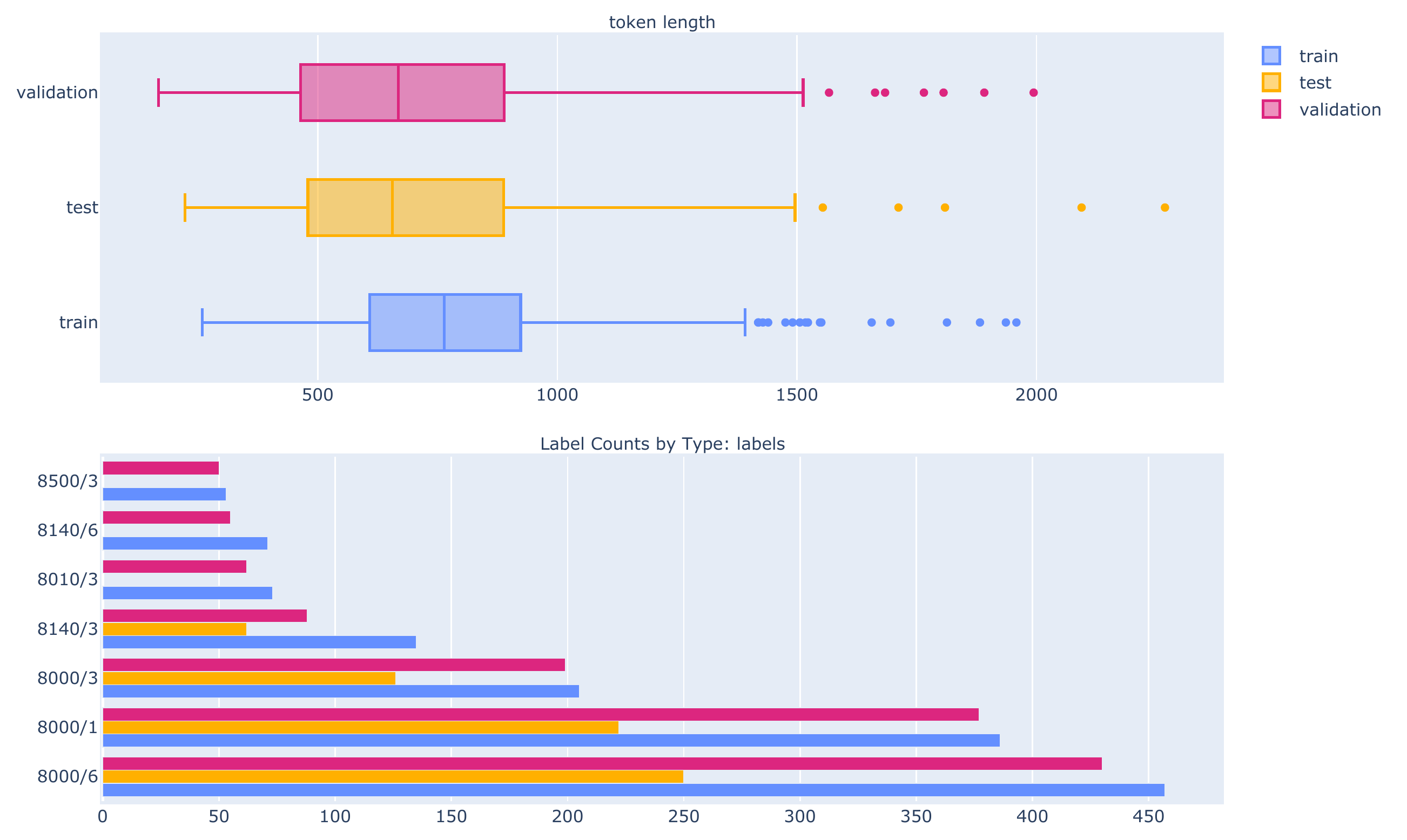}
\caption{\label{fig:cantemist}Token frequency distribution by split (top) and frequency of different kind of instances (bottom).}
\end{figure}

\textbf{Dataset Description:} Collection of 1301 oncological clinical case reports written in Spanish, with tumor morphology mentions manually annotated and mapped by clinical experts to a controlled terminology. Every tumor morphology mention is linked to an eCIE-O code (the Spanish equivalent of ICD-O).
The original dataset is distributed in BRAT format, and was randomly sampled into 3 subsets. The training, development and test sets contain 501, 500 and 300 documents each, respectively. This dataset was designed for the CANcer TExt Mining Shared Task, sponsored by Plan-TL. The task is divided in 3 subtasks: CANTEMIST-NER, CANTEMIST-NORM and CANTEMIST-CODING.

CANTEMIST-NER track: requires finding automatically tumor morphology mentions. All tumor morphology mentions are defined by their corresponding character offsets in UTF-8 plain text medical documents.

CANTEMIST-NORM track: clinical concept normalization or named entity normalization task that requires to return all tumor morphology entity mentions together with their corresponding eCIE-O-3.1 codes i.e. finding and normalizing tumor morphology mentions.

CANTEMIST-CODING track: requires returning for each of document a ranked list of its corresponding ICD-O-3 codes. This it is essentially a sort of indexing or multi-label classification task or oncology clinical coding. 

For further information, please visit \url{https://temu.bsc.es/cantemist} or send an email to encargo-pln-life@bsc.es

\textbf{Homepage:} \url{https://temu.bsc.es/cantemist/?p=4338}

\textbf{URL:} \url{https://zenodo.org/record/3978041/files/cantemist.zip?download=1}

\textbf{Licensing:} Creative Commons Attribution 4.0 International

\textbf{Languages:} Spanish

\textbf{Tasks:} NER, NED, Text Classification

\textbf{Schemas:} {\tt TEXT}, {\tt KB}, {\tt source}

\textbf{Splits:} train, validation, test

%% file: mediqa_qa_mediqa_qa_bigbio_qa.tex
\clearpage
\section*{MEDIQA Data Card}
\begin{figure}[ht!]
\centering
\includegraphics[width=\linewidth]{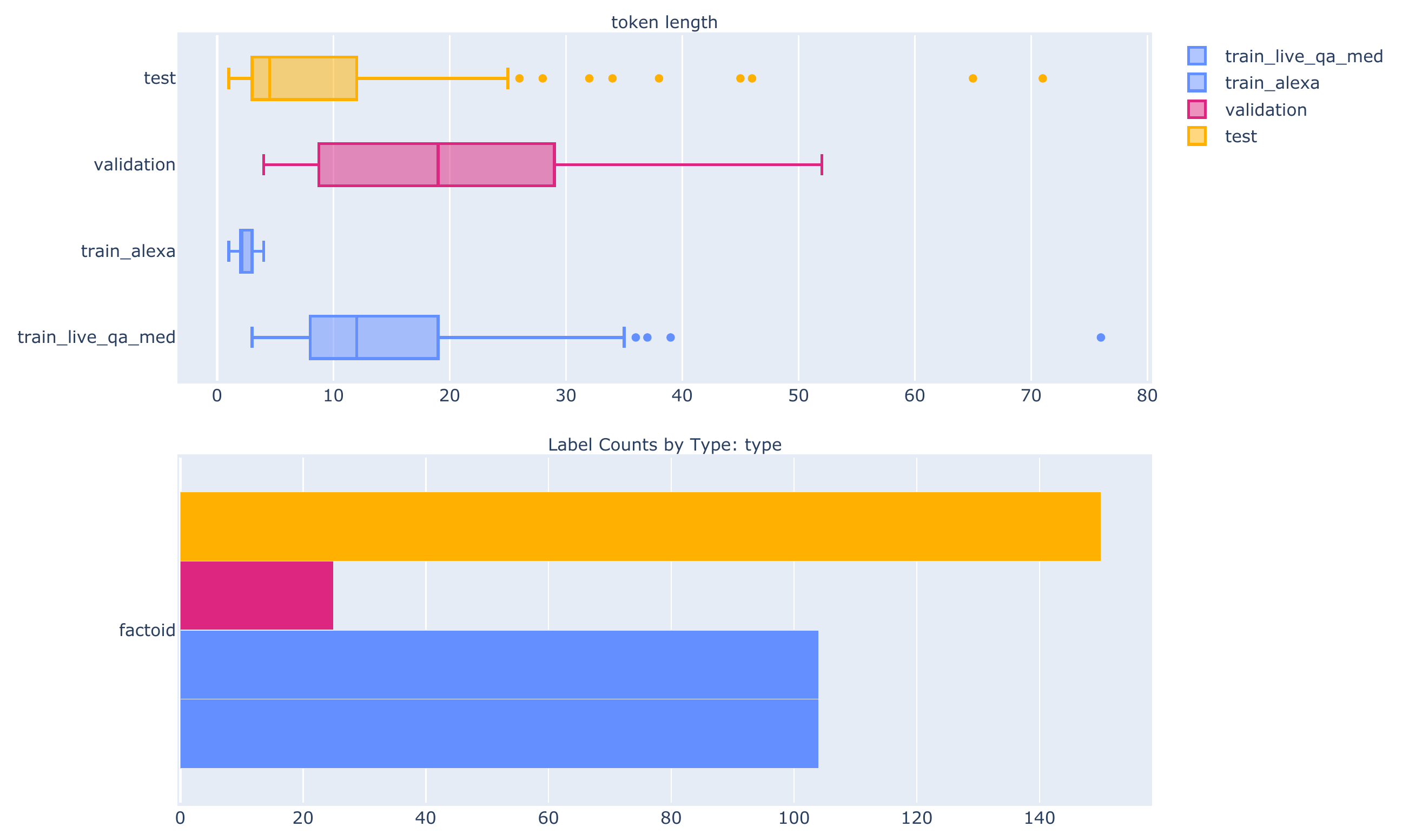}
\caption{\label{fig:mediqa_qa}Token frequency distribution by split (top) and frequency of different kind of instances (bottom).}
\end{figure}

\textbf{Dataset Description:} The MEDIQA challenge is an ACL-BioNLP 2019 shared task aiming to attract further research efforts in Natural Language Inference (NLI), Recognizing Question Entailment (RQE), and their applications in medical Question Answering (QA). Mailing List: \url{https://groups.google.com/forum/#!forum/bionlp-mediqa}

In the QA task, participants are tasked to:- filter/classify the provided answers (1: correct, 0: incorrect).- re-rank the answers.

\textbf{Homepage:} \url{https://sites.google.com/view/mediqa2019}

\textbf{URL:} \url{https://github.com/abachaa/MEDIQA2019/archive/refs/heads/master.zip}

\textbf{Licensing:} License information unavailable

\textbf{Languages:} English

\textbf{Tasks:} Question Answering

\textbf{Schemas:} {\tt QA}, {\tt source}

\textbf{Splits:} train-1-liveQAMed, train-2-Alexa, validation, test

%% file: an_em_an_em_bigbio_kb.tex
\clearpage
\section*{AnEM Data Card}
\begin{figure}[ht!]
\centering
\includegraphics[width=\linewidth]{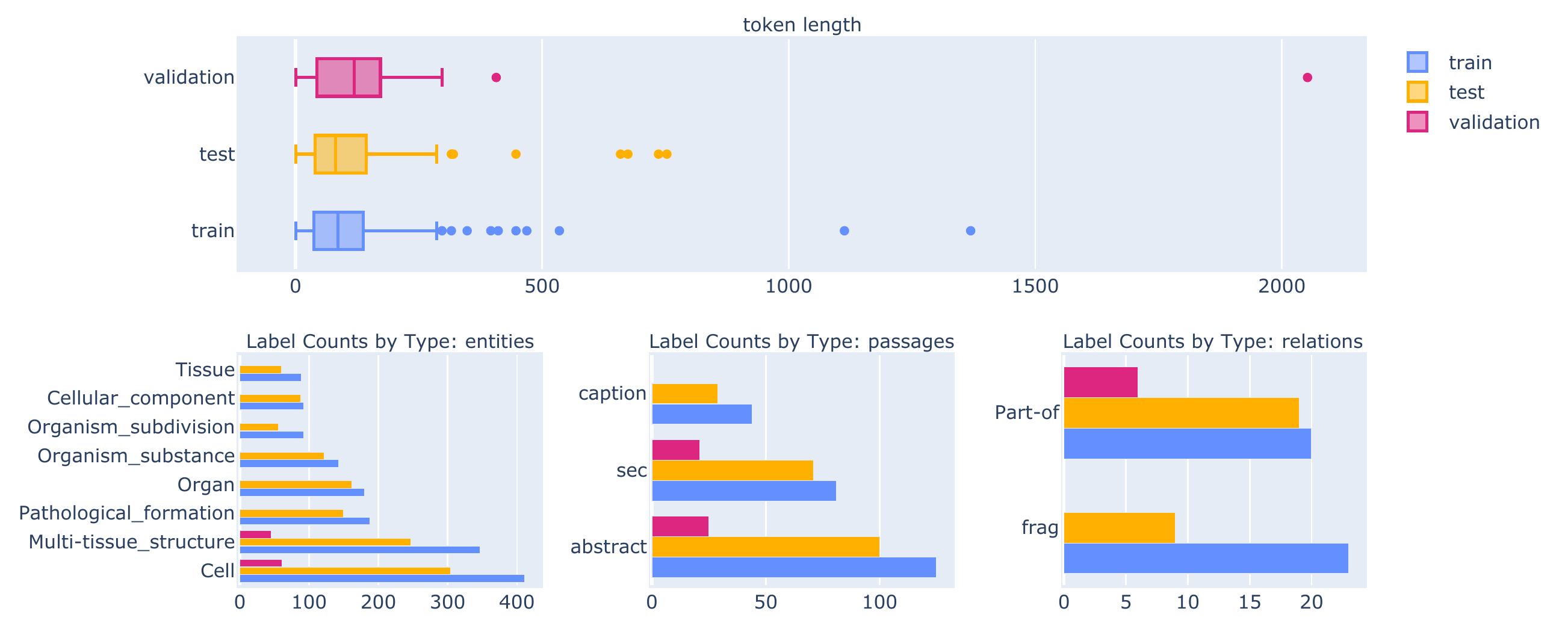}
\caption{\label{fig:an_em}Token frequency distribution by split (top) and frequency of different kind of instances (bottom).}
\end{figure}

\textbf{Dataset Description} AnEM corpus is a domain- and species-independent resource manually annotated for anatomical entity mentions using a fine-grained classification system. The corpus consists of 500 documents (over 90,000 words) selected randomly from citation abstracts and full-text papers with the aim of making the corpus representative of the entire available biomedical scientific literature. The corpus annotation covers mentions of both healthy and pathological anatomical entities and contains over 3,000 annotated mentions.

\textbf{Homepage:} \url{http://www.nactem.ac.uk/anatomy/}

\textbf{URL:} \url{http://www.nactem.ac.uk/anatomy/data/AnEM-1.0.4.tar.gz}

\textbf{Licensing:} Creative Commons Attribution Share Alike 3.0 Unported

\textbf{Languages:} English

\textbf{Tasks:} NER, Coreference Resolution, Relation Extraction

\textbf{Schemas:} {\tt KB}, {\tt source}

\textbf{Splits:} train, validation, test

%% file: paramed_paramed_bigbio_t2t.tex
\clearpage
\section*{ParaMed Data Card}
\begin{figure}[ht!]
\centering
\includegraphics[width=\linewidth]{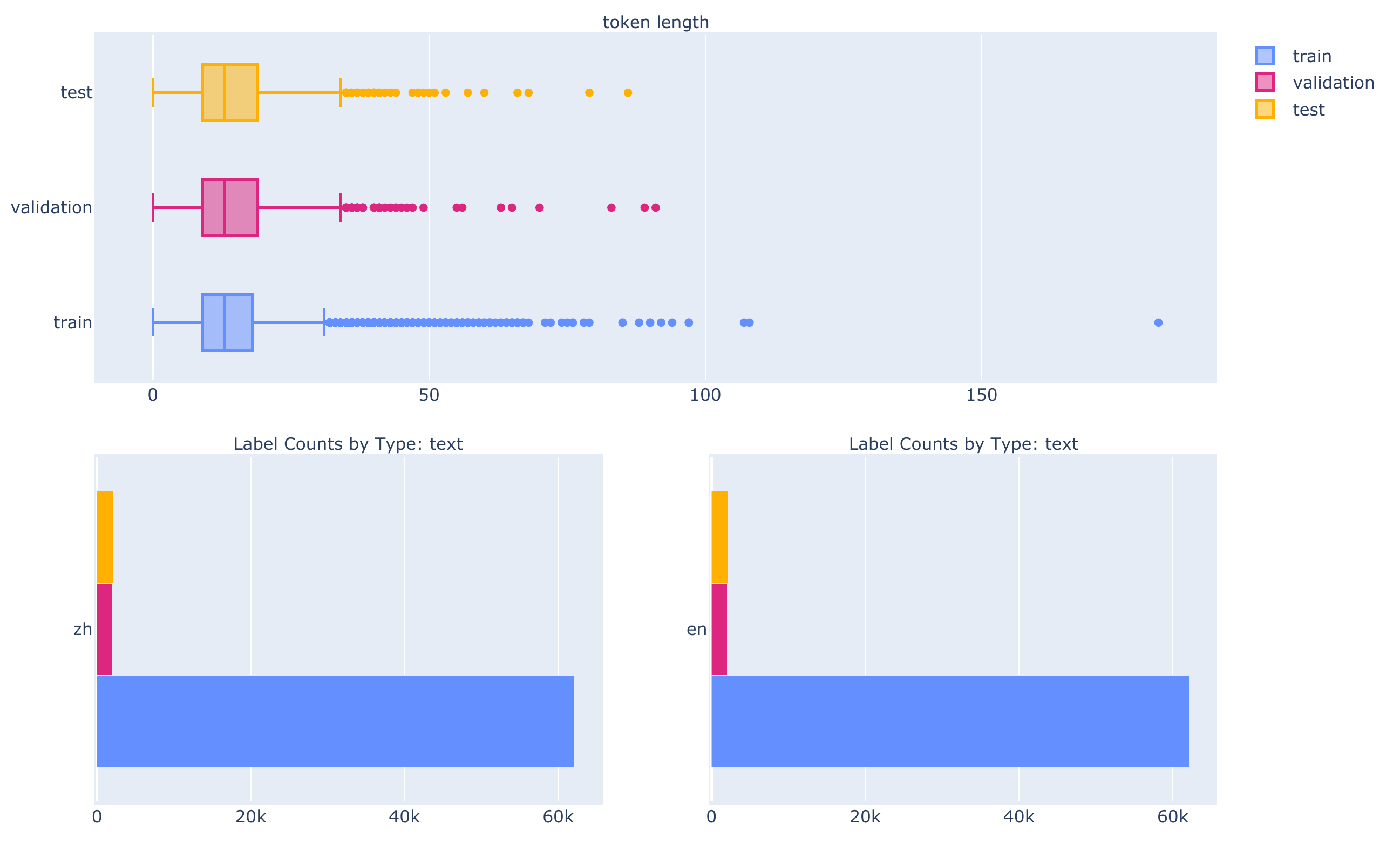}
\caption{\label{fig:paramed}Token frequency distribution by split (top) and frequency of different kind of instances (bottom).}
\end{figure}

\textbf{Dataset Description:} NEJM is a Chinese-English parallel corpus crawled from the New England Journal of Medicine website. English articles are distributed through https://www.nejm.org/ and Chinese articles are distributed through http://nejmqianyan.cn/. The corpus contains all article pairs (around 2000 pairs) since 2011.

\textbf{Homepage:} \url{https://github.com/boxiangliu/ParaMed}

\textbf{URL:}~\url{https://github.com/boxiangliu/ParaMed/blob/master/data/nejm-open-access.tar.gz?raw=true}

\textbf{Licensing:} Creative Commons Attribution 4.0 International

\textbf{Languages:} English, Chinese

\textbf{Tasks} Translation

\textbf{Schemas:} {\tt t2t}, {\tt source}

\textbf{Splits:} train, validation, test

%% file: scitail_scitail_bigbio_te.tex
\clearpage
\section*{SciTail Data Card}
\begin{figure}[ht!]
\centering
\includegraphics[width=\linewidth]{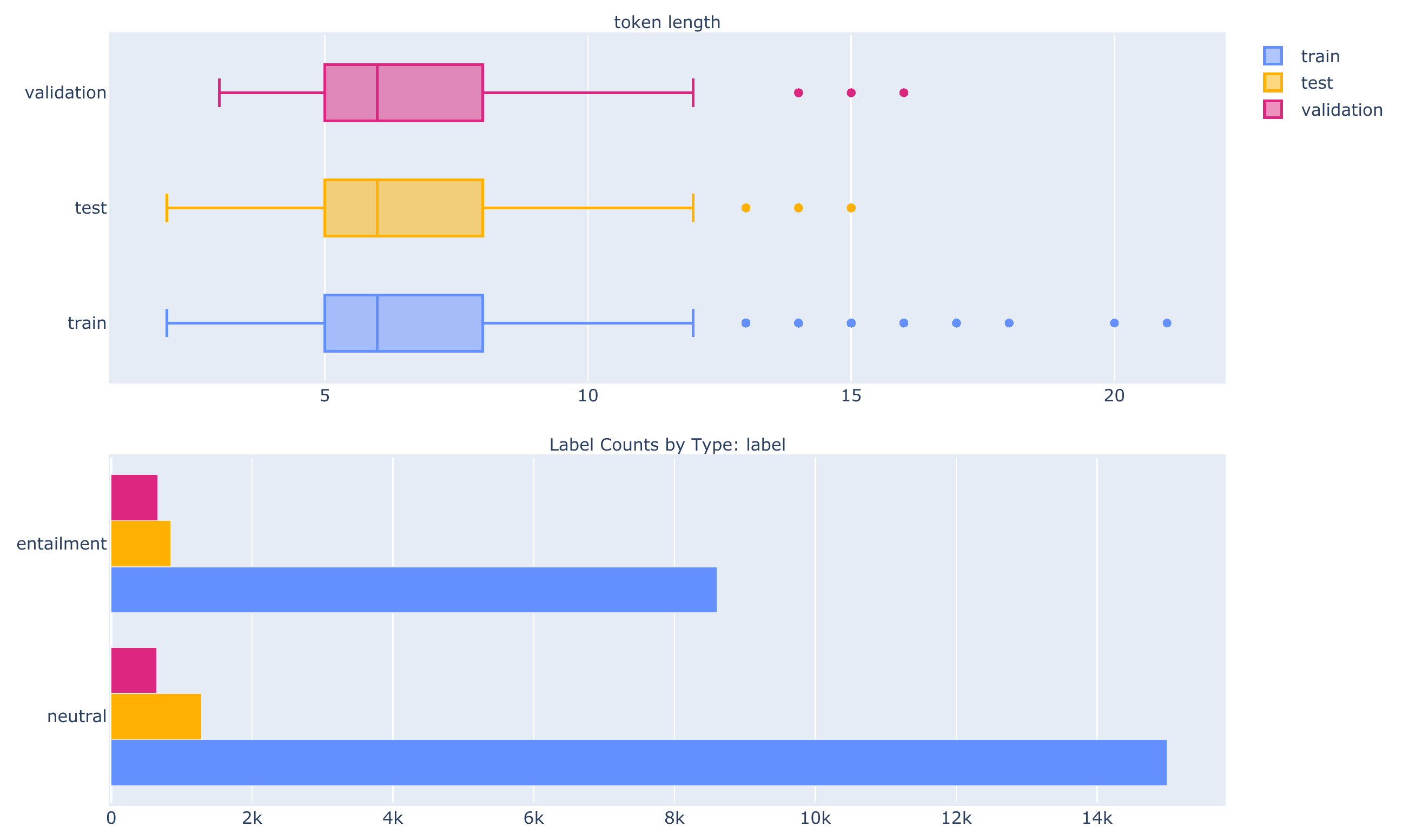}
\caption{\label{fig:scitail}Token frequency distribution by split (top) and frequency of different kind of instances (bottom).}
\end{figure}

\textbf{Dataset Description} The SciTail dataset is an entailment dataset created from multiple-choice science exams and web sentences. Each question and the correct answer choice are converted into an assertive statement to form the hypothesis. We use information retrieval to obtain relevant text from a large text corpus of web sentences, and use these sentences as a premise P. We crowdsource the annotation of such premise-hypothesis pair as supports (entails) or not (neutral), in order to create the SciTail dataset. The dataset contains 27,026 examples with 10,101 examples with entails label and 16,925 examples with neutral label.

\textbf{Homepage:} \url{https://allenai.org/data/scitail}

\textbf{URL:} \url{https://ai2-public-datasets.s3.amazonaws.com/scitail/SciTailV1.1.zip}

\textbf{Licensing:} Apache License 2.0

\textbf{Languages:} English

\textbf{Tasks:} Textual Entailment

\textbf{Schemas:} {\tt te}, {\tt source}

\textbf{Splits:} train, validation, test

%% file: mqp_mqp_bigbio_pairs.tex
\clearpage
\section*{MQP Data Card}
\begin{figure}[ht!]
\centering
\includegraphics[width=\linewidth]{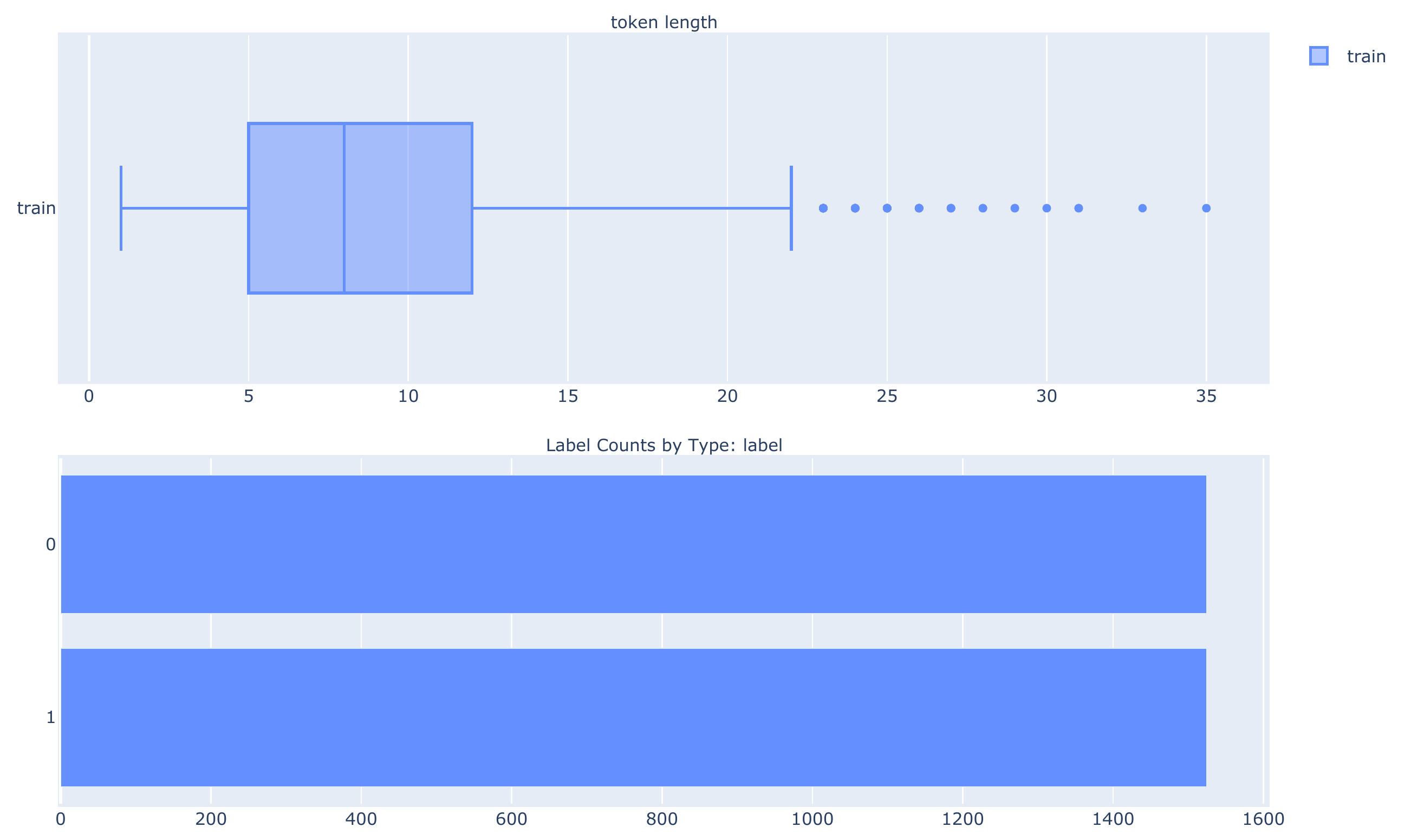}
\caption{\label{fig:mqp}Token frequency distribution by split (top) and frequency of different kind of instances (bottom).}
\end{figure}

\textbf{Dataset Description:} Medical Question Pairs dataset by McCreery et al (2020) contains pairs of medical questions and paraphrased versions of the question prepared by medical professional. Paraphrased versions were labelled as similar (syntactically dissimilar but contextually similar ) or dissimilar (syntactically may look similar but contextually dissimilar). Labels 1: similar, 0: dissimilar

\textbf{Homepage:} \url{https://github.com/curai/medical-question-pair-dataset}

\textbf{URL:} \url{https://raw.githubusercontent.com/curai/medical-question-pair-dataset/master/mqp.csv}

\textbf{Licensing:} License information unavailable

\textbf{Languages:} English

\textbf{Tasks:} Semantic Similarity

\textbf{Schemas:} {\tt pairs}, {\tt source}

\textbf{Splits:} train

%% file: 23_appx_bigbio_datasheet.tex
\clearpage
\section{\ours~Data Card}
\label{sec:bigbiodatasheet}

\textbf{Dataset Description:} 
\ours~ is a community project and meta-dataset consisting of 126+ dataset loader scripts providing programmatic access to expertly annotated biomedical natural language processing datasets. The constituent datasets support 12 tasks grouped into 6 schema types. 105 of these datasets are publicly available and can be automatically downloaded using the \ours~Python package. The remaining 21 require some level of manual action ranging from simple web forms to credentialed access and training on how to handle protected health information.

\textbf{Homepage:} \url{https://github.com/bigscience-workshop/biomedical}

\textbf{URL:} \url{https://github.com/bigscience-workshop/biomedical}

\textbf{Licensing:} \url{https://choosealicense.com/licenses/apache-2.0/}

\textbf{Languages:}  English, Spanish, French, Chinese, German, Japanese, Dutch, Portuguese, Swedish, and Vietnamese

\textbf{Tasks:} 
named entity recognition ({\tt NER}), 
named entity disambiguation/normalization ({\tt NED}), 
event extraction ({\tt EE}), 
relation extraction ({\tt RE}), 
coreference resolution ({\tt COREF}), 
question answering ({\tt QA}), 
textual entailment ({\tt TE}), 
text classification ({\tt TXTCLASS}), 
semantic similarity ({\tt STS}), 
paraphrasing ({\tt PARA}), 
translation ({\tt TRANSL}), 
summarization ({\tt SUM}). 

\textbf{Schemas:} 
Knowledge Base ({\tt KB}), 
Question Answering ({\tt QA}), 
Textual Entailment ({\tt TE}), 
Text ({\tt TEXT}), 
Text Pairs ({\tt PAIRS}), 
Text to Text ({\tt T2T}), 
source ({\tt source}).

\textbf{Splits:} train, validation, test, sample

\setlength{\LTcapwidth}{\textwidth}

\begin{scriptsize}
\begin{longtable}{L{2.1cm} L{2.5cm}  p{0.6cm}  R{0.75cm} R{0.75cm} R{0.75cm}  L{1.1cm} p{1.4cm} p{0.75cm} p{0.25cm} p{0.5cm}}
\caption{Summary statistics for all datasets included in \ours. Token counts (\# Toks) assumes white space tokenziation and example instances (\# N) correspond to the unit of text emitted by the dataloader iterable, usually a document, sentence, or text pair. Some datasets include k-folds or multiple training splits, which are noted by $k=*$. See each dataset's data card for more specific details, such as label counts by task.}\\
\toprule
Dataset Name &     \ours~Name &                Split & \# Chars & \# Toks &  \# N &                License &               Tasks & Schema &              Langs & Access \\
\midrule
\multirow{3}{2.1cm}{AnEM~\citesupp{ohta-etal-2012-open}} & \multirow{3}{2.5cm}{{\tt an\_em}} & train & 300k & 44.3k & 250 & \multirow{3}{1.1cm}{CC BY SA 3.0} & \multirow{3}{1.4cm}{RE, NER, COREF} & \multirow{3}{0.75cm}{KB} & \multirow{3}{0.25cm}{EN} & \multirow{3}{0.5cm}{Public} \\
  &   & valid & 85.6k & 11k & 50 &   &   &   &   &   \\
  &   & test & 242k & 36.2k & 200 &   &   &   &   &   \\
\midrule
\multirow{3}{2.1cm}{AnatEM~\citesupp{pyysalo2014anatomical}} & \multirow{3}{2.5cm}{{\tt anat\_em}} & train & 840k & 122k & 606 & \multirow{3}{1.1cm}{CC BY SA 3.0} & \multirow{3}{1.4cm}{NER} & \multirow{3}{0.75cm}{KB} & \multirow{3}{0.25cm}{EN} & \multirow{3}{0.5cm}{Public} \\
  &   & valid & 319k & 44.7k & 202 &   &   &   &   &   \\
  &   & test & 547k & 79.2k & 404 &   &   &   &   &   \\
\midrule
\multirow{6}{2.1cm}{AskAPatient~\citesupp{limsopatham-collier-2016-normalising}} & \multirow{6}{2.5cm}{{\tt ask\_a\_patient}} & train k=10 & 31.3k & 202k & 15665 & \multirow{6}{1.1cm}{CC BY 4.0} & \multirow{6}{1.4cm}{NER, NED} & \multirow{6}{0.75cm}{KB} & \multirow{6}{0.25cm}{EN} & \multirow{6}{0.5cm}{Public} \\
  &   & validation k=10 & 1.74k & 10.6k & 792 &   &   &   &   &   \\
  &   & test k=10 & 1.92k & 11.7k & 866 &   &   &   &   &   \\
\midrule
\multirow{3}{2.1cm}{BC5CDR~\citesupp{DBLP:journals/biodb/LiSJSWLDMWL16}} & \multirow{3}{2.5cm}{{\tt bc5cdr}} & train & 653k & 93k & 500 & \multirow{3}{1.1cm}{Public Domain Mark 1.0} & \multirow{3}{1.4cm}{RE, NER, NED} & \multirow{3}{0.75cm}{KB} & \multirow{3}{0.25cm}{EN} & \multirow{3}{0.5cm}{Public} \\
  &   & valid & 647k & 92.3k & 500 &   &   &   &   &   \\
  &   & test & 677k & 96.5k & 500 &   &   &   &   &   \\
\midrule
\multirow{3}{2.1cm}{BC7-LitCovid~\citesupp{chen2021overview}} & \multirow{3}{2.5cm}{{\tt bc7\_litcovid}} & train & 34.4M & 4.97M & 24960 & \multirow{3}{1.1cm}{\emph{Unknown}} & \multirow{3}{1.4cm}{TXTCLASS} & \multirow{3}{0.75cm}{TEXT} & \multirow{3}{0.25cm}{EN} & \multirow{3}{0.5cm}{Public} \\
  &   & valid & 3.69M & 532k & 2489 &   &   &   &   &   \\
  &   & test & 8.68M & 1.26M & 6239 &   &   &   &   &   \\
\midrule
Bio-SimVerb~\citesupp{article} & {\tt bio\_sim\_verb} & train & 14.9k & 2k & 1000 & \emph{Unknown} & STS & PAIRS & EN & Public \\
\midrule
Bio-SimLex~\citesupp{article} & {\tt bio\_simlex} & train & 16.1k & 1.98k & 988 & \emph{Unknown} & STS & PAIRS & EN & Public \\
\midrule
\multirow{2}{2.1cm}{MESINESP 2021~\citesupp{gasco2021overview}} & \multirow{2}{2.5cm}{{\tt bioasq\_2021\_mesinesp}} & valid & 256k & 38.6k & 109 & \multirow{2}{1.1cm}{CC BY 4.0} & \multirow{2}{1.4cm}{TXTCLASS} & \multirow{2}{0.75cm}{TEXT} & \multirow{2}{0.25cm}{ES} & \multirow{2}{0.5cm}{Public} \\
  &   & test & 59.4k & 9.06k & 119 &   &   &   &   &   \\
\midrule
\multirow{3}{2.1cm}{BioASQ Task B~\citesupp{tsatsaronis2015overview}} & \multirow{3}{2.5cm}{{\tt bioasq\_task\_b}} & train & 5.21M & 2.26M & 9955 & \multirow{3}{1.1cm}{NLM} & \multirow{3}{1.4cm}{QA} & \multirow{3}{0.75cm}{QA} & \multirow{3}{0.25cm}{EN} & \multirow{3}{0.5cm}{DUA} \\
  &   & valid & 573k & 249k & 1029 &   &   &   &   &   \\
  &   & test & 581k & 253k & 1041 &   &   &   &   &   \\
\midrule
\multirow{2}{2.1cm}{BioASQ Task C 2017~\citesupp{nentidis-etal-2017-results}} & \multirow{2}{2.5cm}{{\tt bioasq\_task\_c\_2017}} & train & 2.59B & 346M & 62952 & \multirow{2}{1.1cm}{NLM} & \multirow{2}{1.4cm}{TXTCLASS} & \multirow{2}{0.75cm}{TEXT} & \multirow{2}{0.25cm}{EN} & \multirow{2}{0.5cm}{DUA} \\
  &   & test & 895M & 120M & 22610 &   &   &   &   &   \\
\midrule
\multirow{2}{2.1cm}{BioInfer~\citesupp{pyysalo2007bioinfer}} & \multirow{2}{2.5cm}{{\tt bioinfer}} & train & 164k & 23.7k & 894 & \multirow{2}{1.1cm}{CC BY 2.0} & \multirow{2}{1.4cm}{RE, NER} & \multirow{2}{0.75cm}{KB} & \multirow{2}{0.25cm}{EN} & \multirow{2}{0.5cm}{Public} \\
  &   & test & 40.8k & 5.93k & 206 &   &   &   &   &   \\
\midrule
BiologyHowWhyCorpus~\citesupp{jansen-etal-2014-discourse} & {\tt biology\_how\_ \ why\_corpus} & train & 2.33M & 985k & 1269 & \emph{Unknown} & QA & QA & EN & Public \\
\midrule
BIOMRC~\citesupp{pappas-etal-2020-biomrc} & {\tt biomrc} & train & 114k & 48.6k & 30 & \emph{Unknown} & QA & QA & EN & Public \\
\midrule
\multirow{3}{2.1cm}{BioNLP 2009~\citesupp{kim-etal-2009-overview}} & \multirow{3}{2.5cm}{{\tt bionlp\_shared\_ \ task\_2009}} & train & 1.21M & 176k & 800 & \multirow{3}{1.1cm}{GENIA Project} & \multirow{3}{1.4cm}{EE, NER, COREF} & \multirow{3}{0.75cm}{KB} & \multirow{3}{0.25cm}{EN} & \multirow{3}{0.5cm}{Public} \\
  &   & valid & 234k & 33.8k & 150 &   &   &   &   &   \\
  &   & test & 397k & 57.3k & 260 &   &   &   &   &   \\
\midrule
\multirow{3}{2.1cm}{BioNLP 2011 EPI~\citesupp{ohta-etal-2011-overview}} & \multirow{3}{2.5cm}{{\tt bionlp\_st\_2011\_epi}} & train & 901k & 127k & 600 & \multirow{3}{1.1cm}{GENIA Project} & \multirow{3}{1.4cm}{EE, NER, COREF} & \multirow{3}{0.75cm}{KB} & \multirow{3}{0.25cm}{EN} & \multirow{3}{0.5cm}{Public} \\
  &   & valid & 310k & 43.5k & 200 &   &   &   &   &   \\
  &   & test & 653k & 91.9k & 440 &   &   &   &   &   \\
\midrule
\multirow{3}{2.1cm}{BioNLP 2011 GE~\citesupp{10.5555/2107691.2107693}} & \multirow{3}{2.5cm}{{\tt bionlp\_st\_2011\_ge}} & train & 1.41M & 206k & 908 & \multirow{3}{1.1cm}{CC BY 3.0} & \multirow{3}{1.4cm}{EE, NER, COREF} & \multirow{3}{0.75cm}{KB} & \multirow{3}{0.25cm}{EN} & \multirow{3}{0.5cm}{Public} \\
  &   & valid & 435k & 64.1k & 259 &   &   &   &   &   \\
  &   & test & 541k & 79k & 347 &   &   &   &   &   \\
\midrule
\multirow{3}{2.1cm}{BioNLP 2011 ID~\citesupp{pyysalo-etal-2011-overview}} & \multirow{3}{2.5cm}{{\tt bionlp\_st\_2011\_id}} & train & 438k & 64.7k & 152 & \multirow{3}{1.1cm}{GENIA Project} & \multirow{3}{1.4cm}{EE, NER, COREF} & \multirow{3}{0.75cm}{KB} & \multirow{3}{0.25cm}{EN} & \multirow{3}{0.5cm}{Public} \\
  &   & valid & 119k & 18.4k & 46 &   &   &   &   &   \\
  &   & test & 335k & 50k & 118 &   &   &   &   &   \\
\midrule
\multirow{3}{2.1cm}{BioNLP 2011 REL~\citesupp{10.5555/2107691.2107703}} & \multirow{3}{2.5cm}{{\tt bionlp\_st\_2011\_rel}} & train & 1.21M & 176k & 800 & \multirow{3}{1.1cm}{GENIA Project} & \multirow{3}{1.4cm}{RE, NER, COREF} & \multirow{3}{0.75cm}{KB} & \multirow{3}{0.25cm}{EN} & \multirow{3}{0.5cm}{Public} \\
  &   & valid & 234k & 33.8k & 150 &   &   &   &   &   \\
  &   & test & 397k & 57.3k & 260 &   &   &   &   &   \\
\midrule
\multirow{3}{2.1cm}{BioNLP 2013 CG~\citesupp{pyysalo-etal-2013-overview}} & \multirow{3}{2.5cm}{{\tt bionlp\_st\_2013\_cg}} & train & 467k & 66.1k & 300 & \multirow{3}{1.1cm}{GENIA Project} & \multirow{3}{1.4cm}{EE, NER, COREF} & \multirow{3}{0.75cm}{KB} & \multirow{3}{0.25cm}{EN} & \multirow{3}{0.5cm}{Public} \\
  &   & valid & 153k & 21.7k & 100 &   &   &   &   &   \\
  &   & test & 297k & 42.1k & 200 &   &   &   &   &   \\
\midrule
\multirow{3}{2.1cm}{BioNLP 2013 GE~\citesupp{kim-etal-2013-genia}} & \multirow{3}{2.5cm}{{\tt bionlp\_st\_2013\_ge}} & train & 371k & 54.9k & 222 & \multirow{3}{1.1cm}{GENIA Project} & \multirow{3}{1.4cm}{RE, EE, NER, COREF} & \multirow{3}{0.75cm}{KB} & \multirow{3}{0.25cm}{EN} & \multirow{3}{0.5cm}{Public} \\
  &   & valid & 391k & 57.9k & 249 &   &   &   &   &   \\
  &   & test & 506k & 75.1k & 305 &   &   &   &   &   \\
\midrule
\multirow{3}{2.1cm}{BioNLP 2013 GRO~\citesupp{kim-etal-2013-gro}} & \multirow{3}{2.5cm}{{\tt bionlp\_st\_2013\_gro}} & train & 200k & 29.4k & 150 & \multirow{3}{1.1cm}{GENIA Project} & \multirow{3}{1.4cm}{RE, EE, NER} & \multirow{3}{0.75cm}{KB} & \multirow{3}{0.25cm}{EN} & \multirow{3}{0.5cm}{Public} \\
  &   & valid & 59.8k & 8.7k & 50 &   &   &   &   &   \\
  &   & test & 132k & 19k & 100 &   &   &   &   &   \\
\midrule
\multirow{3}{2.1cm}{BioNLP 2013 PC~\citesupp{ohta-etal-2013-overview}} & \multirow{3}{2.5cm}{{\tt bionlp\_st\_2013\_pc}} & train & 378k & 53.8k & 260 & \multirow{3}{1.1cm}{GENIA Project} & \multirow{3}{1.4cm}{EE, NER, COREF} & \multirow{3}{0.75cm}{KB} & \multirow{3}{0.25cm}{EN} & \multirow{3}{0.5cm}{Public} \\
  &   & valid & 131k & 18.6k & 90 &   &   &   &   &   \\
  &   & test & 253k & 36k & 175 &   &   &   &   &   \\
\midrule
\multirow{3}{2.1cm}{BioNLP 2019 BB~\citesupp{bossy-etal-2019-bacteria}} & \multirow{3}{2.5cm}{{\tt bionlp\_st\_2019\_bb}} & train & 129k & 19k & 133 & \multirow{3}{1.1cm}{\emph{Unknown}} & \multirow{3}{1.4cm}{RE, NER, NED} & \multirow{3}{0.75cm}{KB} & \multirow{3}{0.25cm}{EN} & \multirow{3}{0.5cm}{Public} \\
  &   & valid & 66.5k & 9.71k & 66 &   &   &   &   &   \\
  &   & test & 110k & 16.2k & 96 &   &   &   &   &   \\
\midrule
\multirow{3}{2.1cm}{BioRED~\citesupp{DBLP:journals/corr/abs-2204-04263}} & \multirow{3}{2.5cm}{{\tt biored}} & train & 660k & 94.2k & 400 & \multirow{3}{1.1cm}{\emph{Unknown}} & \multirow{3}{1.4cm}{RE, NER} & \multirow{3}{0.75cm}{KB} & \multirow{3}{0.25cm}{EN} & \multirow{3}{0.5cm}{Public} \\
  &   & valid & 173k & 24.9k & 100 &   &   &   &   &   \\
  &   & test & 168k & 24.1k & 100 &   &   &   &   &   \\
\midrule
\multirow{2}{2.1cm}{BioRelEx~\citesupp{khachatrian2019biorelex}} & \multirow{2}{2.5cm}{{\tt biorelex}} & train & 237k & 37.8k & 1405 & \multirow{2}{1.1cm}{\emph{Unknown}} & \multirow{2}{1.4cm}{RE, NER, NED, COREF} & \multirow{2}{0.75cm}{KB} & \multirow{2}{0.25cm}{EN} & \multirow{2}{0.5cm}{Public} \\
  &   & valid & 33.1k & 5.29k & 201 &   &   &   &   &   \\
\midrule
BioScope~\citesupp{vincze2008bioscope} & {\tt bioscope} & train & 171k & 42k & 6383 & CC BY 2.0 & NER & KB & EN & Public \\
\midrule
\multirow{3}{2.1cm}{BIOSSES~\citesupp{souganciouglu2017biosses}} & \multirow{3}{2.5cm}{{\tt biosses}} & train & 20.1k & 2.94k & 64 & \multirow{3}{1.1cm}{GPL 3.0} & \multirow{3}{1.4cm}{STS} & \multirow{3}{0.75cm}{PAIRS} & \multirow{3}{0.25cm}{EN} & \multirow{3}{0.5cm}{Public} \\
  &   & valid & 5.09k & 733 & 16 &   &   &   &   &   \\
  &   & test & 6.44k & 925 & 20 &   &   &   &   &   \\
\midrule
CADEC~\citesupp{karimi2015cadec} & {\tt cadec} & train & 575k & 104k & 1250 & Custom & NER, NED & KB & EN & Public \\
\midrule
\multirow{3}{2.1cm}{CANTEMIST~\citesupp{miranda2020named}} & \multirow{3}{2.5cm}{{\tt cantemist}} & train & 2.6M & 382k & 501 & \multirow{3}{1.1cm}{CC BY 4.0} & \multirow{3}{1.4cm}{NER, NED, TXTCLASS} & \multirow{3}{0.75cm}{KB, TEXT} & \multirow{3}{0.25cm}{ES} & \multirow{3}{0.5cm}{Public} \\
  &   & valid & 2.33M & 341k & 500 &   &   &   &   &   \\
  &   & test & 1.41M & 206k & 300 &   &   &   &   &   \\
\midrule
CAS~\citesupp{grabar-etal-2018-cas} & {\tt cas} & train & 972k & 175k & 7580 & DUA & TXTCLASS & TEXT, KB & FR & DUA \\
\midrule
\multirow{2}{2.1cm}{CellFinder~\citesupp{neves2012annotating}} & \multirow{2}{2.5cm}{{\tt cellfinder}} & train & 171k & 25.2k & 26 & \multirow{2}{1.1cm}{CC BY SA 3.0} & \multirow{2}{1.4cm}{NER} & \multirow{2}{0.75cm}{KB} & \multirow{2}{0.25cm}{EN} & \multirow{2}{0.5cm}{Public} \\
  &   & test & 205k & 30.3k & 27 &   &   &   &   &   \\
\midrule
CHEBI Corpus~\citesupp{Shardlow2018} & {\tt chebi\_nactem} & train & 1.95M & 306k & 100 & CC BY 4.0 & RE, NER & KB & EN & Public \\
\midrule
\multirow{3}{2.1cm}{CHEMDNER~\citesupp{krallinger2015}} & \multirow{3}{2.5cm}{{\tt chemdner}} & train & 4.88M & 687k & 3500 & \multirow{3}{1.1cm}{\emph{Unknown}} & \multirow{3}{1.4cm}{NER, TXTCLASS} & \multirow{3}{0.75cm}{KB, TEXT} & \multirow{3}{0.25cm}{EN} & \multirow{3}{0.5cm}{Public} \\
  &   & valid & 4.86M & 683k & 3500 &   &   &   &   &   \\
  &   & test & 4.19M & 591k & 3000 &   &   &   &   &   \\
\midrule
\multirow{3}{2.1cm}{ChemProt~\citesupp{DBLP:journals/biodb/LiSJSWLDMWL16}} & \multirow{3}{2.5cm}{{\tt chemprot}} & train & 1.64M & 230k & 1020 & \multirow{3}{1.1cm}{Public Domain Mark 1.0} & \multirow{3}{1.4cm}{RE, NER} & \multirow{3}{0.75cm}{KB} & \multirow{3}{0.25cm}{EN} & \multirow{3}{0.5cm}{Public} \\
  &   & valid & 990k & 139k & 612 &   &   &   &   &   \\
  &   & test & 1.3M & 182k & 800 &   &   &   &   &   \\
\midrule
CHIA~\citesupp{kury2020chia} & {\tt chia} & train & 1.04M & 151k & 2000 & CC BY 4.0 & RE, NER & KB & EN & Public \\
\midrule
Citation GIA Test Collection~\citesupp{Wei2015} & {\tt citation\_gia\_ \ test\_collection} & test & 230k & 33.4k & 151 & \emph{Unknown} & NER, NED & KB & EN & Public \\
\midrule
\multirow{4}{2.1cm}{CodiEsp~\citesupp{miranda2020overview}} & \multirow{4}{2.5cm}{{\tt codiesp}} & train & 193M & 29.1M & 176294 & \multirow{4}{1.1cm}{CC BY 4.0} & \multirow{4}{1.4cm}{TXTCLASS} & \multirow{4}{0.75cm}{TEXT} & \multirow{4}{0.25cm}{ES} & \multirow{4}{0.5cm}{Public} \\
  &   & train & 0 & 0 & 500 &   &   &   &   &   \\
  &   & valid & 0 & 0 & 250 &   &   &   &   &   \\
  &   & test & 0 & 0 & 250 &   &   &   &   &   \\
\midrule
CORD-NER~\citesupp{DBLP:journals/corr/abs-2003-12218} & {\tt cord\_ner} & train & 407M & 62.5M & 29500 & Custom & NER & KB & EN & Public \\
\midrule
\multirow{3}{2.1cm}{CT-EBM-SP~\citesupp{CampillosLlanos2021}} & \multirow{3}{2.5cm}{{\tt ctebmsp}} & train & 625k & 90.3k & 420 & \multirow{3}{1.1cm}{CC BY NC 4.0} & \multirow{3}{1.4cm}{NER} & \multirow{3}{0.75cm}{KB} & \multirow{3}{0.25cm}{ES} & \multirow{3}{0.5cm}{Public} \\
  &   & valid & 212k & 30.7k & 140 &   &   &   &   &   \\
  &   & test & 206k & 29.9k & 140 &   &   &   &   &   \\
\midrule
\multirow{2}{2.1cm}{DDI Corpus~\citesupp{HERREROZAZO2013914}} & \multirow{2}{2.5cm}{{\tt ddi\_corpus}} & train & 928k & 128k & 714 & \multirow{2}{1.1cm}{CC BY NC 4.0} & \multirow{2}{1.4cm}{RE, NER} & \multirow{2}{0.75cm}{KB} & \multirow{2}{0.25cm}{EN} & \multirow{2}{0.5cm}{Public} \\
  &   & test & 281k & 38.7k & 303 &   &   &   &   &   \\
\midrule
\multirow{4}{2.1cm}{DIANN~\citesupp{DBLP:conf/sepln/2018ibereval}} & \multirow{4}{2.5cm}{{\tt diann\_iber\_eval}} & train & 548k & 81.8k & 400 & \multirow{4}{1.1cm}{\emph{Unknown}} & \multirow{4}{1.4cm}{NER} & \multirow{4}{0.75cm}{KB} & \multirow{4}{0.25cm}{EN, ES} & \multirow{4}{0.5cm}{Public} \\
  &   & test & 144k & 21.5k & 100 &   &   &   &   &   \\
  &   & train & 1.06M & 156k & 400 &   &   &   &   &   \\
  &   & test & 275k & 40.9k & 100 &   &   &   &   &   \\
\midrule
DisTEMIST~\citesupp{luis_gasco_2022_6458455} & {\tt distemist} & train & 1.76M & 264k & 750 & CC BY 4.0 & NER & KB & EN & Public \\
\midrule
\multirow{2}{2.1cm}{EBM NLP~\citesupp{nye-etal-2018-corpus}} & \multirow{2}{2.5cm}{{\tt ebm\_pico}} & train & 7.68M & 1.29M & 4746 & \multirow{2}{1.1cm}{\emph{Unknown}} & \multirow{2}{1.4cm}{NER} & \multirow{2}{0.75cm}{KB} & \multirow{2}{0.25cm}{EN} & \multirow{2}{0.5cm}{Public} \\
  &   & test & 306k & 50.9k & 187 &   &   &   &   &   \\
\midrule
EHR-Rel~\citesupp{schulz-etal-2020-biomedical} & {\tt ehr\_rel} & train & 174k & 23.4k & 3741 & Apache 2.0 & STS & PAIRS & EN & Public \\
\midrule
ESSAI~\citesupp{Dalloux} & {\tt essai} & train & 1.83M & 314k & 13848 & DUA & TXTCLASS & TEXT, KB & FR & DUA \\
\midrule
EU-ADR~\citesupp{VANMULLIGEN2012879} & {\tt euadr} & train & 452k & 64.3k & 300 & \emph{Unknown} & RE, NER & KB & EN & Public \\
\midrule
\multirow{3}{2.1cm}{Evidence Inference 2.0~\citesupp{deyoung-etal-2020-evidence}} & \multirow{3}{2.5cm}{{\tt evidence\_inference}} & train & 2.91M & 446k & 10150 & \multirow{3}{1.1cm}{MIT} & \multirow{3}{1.4cm}{TE} & \multirow{3}{0.75cm}{TE} & \multirow{3}{0.25cm}{EN} & \multirow{3}{0.5cm}{Public} \\
  &   & valid & 352k & 53.6k & 1238 &   &   &   &   &   \\
  &   & test & 358k & 54.9k & 1228 &   &   &   &   &   \\
\midrule
\multirow{3}{2.1cm}{GAD~\citesupp{Bravo2015}} & \multirow{3}{2.5cm}{{\tt gad}} & train & 740k & 113k & 4261 & \multirow{3}{1.1cm}{CC BY 4.0} & \multirow{3}{1.4cm}{TXTCLASS} & \multirow{3}{0.75cm}{TEXT} & \multirow{3}{0.25cm}{EN} & \multirow{3}{0.5cm}{Public} \\
  &   & valid & 91k & 13.9k & 535 &   &   &   &   &   \\
  &   & test & 98.6k & 14.9k & 534 &   &   &   &   &   \\
\midrule
\multirow{3}{2.1cm}{GENETAG~\citesupp{Tanabe2005}} & \multirow{3}{2.5cm}{{\tt genetag}} & train & 1.16M & 197k & 7500 & \multirow{3}{1.1cm}{NCBI} & \multirow{3}{1.4cm}{NER} & \multirow{3}{0.75cm}{KB} & \multirow{3}{0.25cm}{EN} & \multirow{3}{0.5cm}{Public} \\
  &   & valid & 783k & 133k & 5000 &   &   &   &   &   \\
  &   & test & 387k & 65.5k & 2500 &   &   &   &   &   \\
\midrule
PTM Events~\citesupp{ohta-etal-2010-event} & {\tt genia\_ptm\_ \ event\_corpus} & train & 145k & 20.8k & 112 & GENIA Project & EE, NER, COREF & KB & EN & Public \\
\midrule
\multirow{3}{2.1cm}{GENIA Relation Corpus~\citesupp{pyysalo-etal-2009-static}} & \multirow{3}{2.5cm}{{\tt genia\_relation\_ \ corpus}} & train & 1.21M & 176k & 800 & \multirow{3}{1.1cm}{GENIA Project} & \multirow{3}{1.4cm}{RE} & \multirow{3}{0.75cm}{KB} & \multirow{3}{0.25cm}{EN} & \multirow{3}{0.5cm}{Public} \\
  &   & valid & 234k & 33.8k & 150 &   &   &   &   &   \\
  &   & test & 397k & 57.3k & 260 &   &   &   &   &   \\
\midrule
GENIA Term Corpus~\citesupp{10.5555/1289189.1289260} & {\tt genia\_term\_corpus} & train & 2.99M & 435k & 2000 & GENIA Project & NER & KB & EN & Public \\
\midrule
\multirow{2}{2.1cm}{GEOkhoj v1~\citesupp{geokhoj_v1}} & \multirow{2}{2.5cm}{{\tt geokhoj\_v1}} & train & 4.25M & 554k & 25000 & \multirow{2}{1.1cm}{CC BY NC 4.0} & \multirow{2}{1.4cm}{TXTCLASS} & \multirow{2}{0.75cm}{TEXT} & \multirow{2}{0.25cm}{EN} & \multirow{2}{0.5cm}{Public} \\
  &   & test & 848k & 111k & 5000 &   &   &   &   &   \\
\midrule
\multirow{2}{2.1cm}{GNormPlus~\citesupp{Wei2015}} & \multirow{2}{2.5cm}{{\tt gnormplus}} & train & 379k & 55.7k & 281 & \multirow{2}{1.1cm}{\emph{Unknown}} & \multirow{2}{1.4cm}{NER, NED} & \multirow{2}{0.75cm}{KB} & \multirow{2}{0.25cm}{EN} & \multirow{2}{0.5cm}{Public} \\
  &   & test & 359k & 52.5k & 262 &   &   &   &   &   \\
\midrule
\multirow{3}{2.1cm}{Hallmarks of Cancer~\citesupp{DBLP:journals/bioinformatics/BakerSGAHSk16}} & \multirow{3}{2.5cm}{{\tt hallmarks\_of\_cancer}} & train & 1.96M & 312k & 12119 & \multirow{3}{1.1cm}{GPL 3.0} & \multirow{3}{1.4cm}{TXTCLASS} & \multirow{3}{0.75cm}{TEXT} & \multirow{3}{0.25cm}{EN} & \multirow{3}{0.5cm}{Public} \\
  &   & valid & 296k & 47.1k & 1798 &   &   &   &   &   \\
  &   & test & 573k & 91.8k & 3547 &   &   &   &   &   \\
\midrule
\multirow{2}{2.1cm}{HPRD50~\citesupp{fundel2007relex}} & \multirow{2}{2.5cm}{{\tt hprd50}} & train & 18.1k & 2.67k & 34 & \multirow{2}{1.1cm}{\emph{Unknown}} & \multirow{2}{1.4cm}{RE, NER} & \multirow{2}{0.75cm}{KB} & \multirow{2}{0.25cm}{EN} & \multirow{2}{0.5cm}{Public} \\
  &   & test & 4.94k & 710 & 9 &   &   &   &   &   \\
\midrule
\multirow{2}{2.1cm}{IEPA~\citesupp{ding2001mining}} & \multirow{2}{2.5cm}{{\tt iepa}} & train & 75.1k & 10.9k & 160 & \multirow{2}{1.1cm}{\emph{Unknown}} & \multirow{2}{1.4cm}{RE} & \multirow{2}{0.75cm}{KB} & \multirow{2}{0.25cm}{EN} & \multirow{2}{0.5cm}{Public} \\
  &   & test & 18.6k & 2.68k & 40 &   &   &   &   &   \\
\midrule
\multirow{2}{2.1cm}{JNLPBA~\citesupp{collier-kim-2004-introduction}} & \multirow{2}{2.5cm}{{\tt jnlpba}} & train & 0 & 0 & 37094 & \multirow{2}{1.1cm}{CC BY 3.0} & \multirow{2}{1.4cm}{NER} & \multirow{2}{0.75cm}{KB} & \multirow{2}{0.25cm}{EN} & \multirow{2}{0.5cm}{Public} \\
  &   & valid & 0 & 0 & 7714 &   &   &   &   &   \\
\midrule
LINNAEUS~\citesupp{gerner2010linnaeus} & {\tt linnaeus} & train & 2.46M & 373k & 84 & CC BY 4.0 & NER, NED & KB & EN & Public \\
\midrule
\multirow{2}{2.1cm}{LLL05~\citesupp{article}} & \multirow{2}{2.5cm}{{\tt lll}} & train & 13.2k & 1.99k & 77 & \multirow{2}{1.1cm}{\emph{Unknown}} & \multirow{2}{1.4cm}{RE} & \multirow{2}{0.75cm}{KB} & \multirow{2}{0.25cm}{EN} & \multirow{2}{0.5cm}{Public} \\
  &   & test & 13.1k & 2.07k & 87 &   &   &   &   &   \\
\midrule
Mantra GSC~\citesupp{10.1093/jamia/ocv037} & {\tt mantra\_gsc} & train & 16k & 2.12k & 50 & CC BY 4.0 & NER, NED & KB & EN, FR, DE, NL, ES & Public \\
\midrule
MayoSRS~\citesupp{pedersen2007measures} & {\tt mayosrs} & train & 2.69k & 314 & 101 & CC0 1.0 & STS & PAIRS & EN & Public \\
\midrule
\multirow{3}{2.1cm}{MedQA~\citesupp{jin2021disease}} & \multirow{3}{2.5cm}{{\tt med\_qa}} & train & 1.84M & 890k & 11298 & \multirow{3}{1.1cm}{\emph{Unknown}} & \multirow{3}{1.4cm}{QA} & \multirow{3}{0.75cm}{QA} & \multirow{3}{0.25cm}{EN} & \multirow{3}{0.5cm}{Public} \\
  &   & valid & 229k & 111k & 1412 &   &   &   &   &   \\
  &   & test & 234k & 114k & 1413 &   &   &   &   &   \\
\midrule
\multirow{3}{2.1cm}{MedDialog~\citesupp{DBLP:journals/corr/abs-2004-03329}} & \multirow{3}{2.5cm}{{\tt meddialog}} & train & 290k & 51k & 981 & \multirow{3}{1.1cm}{\emph{Unknown}} & \multirow{3}{1.4cm}{TXTCLASS} & \multirow{3}{0.75cm}{TEXT} & \multirow{3}{0.25cm}{EN, ZH} & \multirow{3}{0.5cm}{Public} \\
  &   & valid & 41.8k & 7.35k & 126 &   &   &   &   &   \\
  &   & test & 35.5k & 6.31k & 122 &   &   &   &   &   \\
\midrule
\multirow{3}{2.1cm}{MEDDOCAN~\citesupp{marimon2019automatic}} & \multirow{3}{2.5cm}{{\tt meddocan}} & train & 1.42M & 208k & 500 & \multirow{3}{1.1cm}{CC BY 4.0} & \multirow{3}{1.4cm}{NER} & \multirow{3}{0.75cm}{KB} & \multirow{3}{0.25cm}{ES} & \multirow{3}{0.5cm}{Public} \\
  &   & valid & 755k & 111k & 250 &   &   &   &   &   \\
  &   & test & 711k & 105k & 250 &   &   &   &   &   \\
\midrule
\multirow{2}{2.1cm}{MedHop~\citesupp{welbl-etal-2018-constructing}} & \multirow{2}{2.5cm}{{\tt medhop}} & train & 187M & 78.7M & 1620 & \multirow{2}{1.1cm}{CC BY SA 3.0} & \multirow{2}{1.4cm}{QA} & \multirow{2}{0.75cm}{QA} & \multirow{2}{0.25cm}{EN} & \multirow{2}{0.5cm}{Public} \\
  &   & valid & 32.8M & 13.8M & 342 &   &   &   &   &   \\
\midrule
\multirow{2}{2.1cm}{Medical Data~\citesupp{ask9medicaldata}} & \multirow{2}{2.5cm}{{\tt medical\_data}} & train & 11M & 1.81M & 5279 & \multirow{2}{1.1cm}{\emph{Unknown}} & \multirow{2}{1.4cm}{TE} & \multirow{2}{0.75cm}{TE} & \multirow{2}{0.25cm}{EN} & \multirow{2}{0.5cm}{DUA} \\
  &   & test & 7.12M & 1.16M & 2924 &   &   &   &   &   \\
\midrule
MEDIQA NLI~\citesupp{https://doi.org/10.13026/gtv4-g455} & {\tt mediqa\_nli} & test & 49.6k & 8.37k & 405 & PhysioNet 1.5 & TE & TE & EN & DUA \\
\midrule
\multirow{4}{2.1cm}{MEDIQA QA~\citesupp{MEDIQA2019}} & \multirow{4}{2.5cm}{{\tt mediqa\_qa}} & train k=2 & 1.91M & 4.56M & 104 & \multirow{4}{1.1cm}{\emph{Unknown}} & \multirow{4}{1.4cm}{QA} & \multirow{4}{0.75cm}{QA} & \multirow{4}{0.25cm}{EN} & \multirow{4}{0.5cm}{Public} \\
  &   & valid & 1.24M & 519k & 25 &   &   &   &   &   \\
  &   & test & 5.78M & 2.42M & 150 &   &   &   &   &   \\
\midrule
\multirow{3}{2.1cm}{MEDIQA RQE~\citesupp{MEDIQA2019}} & \multirow{3}{2.5cm}{{\tt mediqa\_rqe}} & train & 1.69M & 262k & 8588 & \multirow{3}{1.1cm}{\emph{Unknown}} & \multirow{3}{1.4cm}{TE} & \multirow{3}{0.75cm}{TE} & \multirow{3}{0.25cm}{EN} & \multirow{3}{0.5cm}{Public} \\
  &   & valid & 86.5k & 15.6k & 302 &   &   &   &   &   \\
  &   & test & 68.1k & 12.1k & 230 &   &   &   &   &   \\
\midrule
\multirow{3}{2.1cm}{MedMentions~\citesupp{mohan2019medmentions}} & \multirow{3}{2.5cm}{{\tt medmentions}} & train & 4.16M & 606k & 2635 & \multirow{3}{1.1cm}{CC0 1.0} & \multirow{3}{1.4cm}{NER, NED} & \multirow{3}{0.75cm}{KB} & \multirow{3}{0.25cm}{EN} & \multirow{3}{0.5cm}{Public} \\
  &   & valid & 1.4M & 204k & 878 &   &   &   &   &   \\
  &   & test & 1.39M & 203k & 879 &   &   &   &   &   \\
\midrule
\multirow{3}{2.1cm}{MedNLI~\citesupp{https://doi.org/10.13026/c2rs98}} & \multirow{3}{2.5cm}{{\tt mednli}} & train & 1.51M & 240k & 11232 & \multirow{3}{1.1cm}{PhysioNet 1.5} & \multirow{3}{1.4cm}{TE} & \multirow{3}{0.75cm}{TE} & \multirow{3}{0.25cm}{EN} & \multirow{3}{0.5cm}{DUA} \\
  &   & valid & 196k & 31.1k & 1395 &   &   &   &   &   \\
  &   & test & 187k & 29.6k & 1422 &   &   &   &   &   \\
\midrule
MeQSum~\citesupp{ben-abacha-demner-fushman-2019-summarization} & {\tt meqsum} & train & 405k & 70.8k & 1000 & \emph{Unknown} & SUM & T2T & EN & Public \\
\midrule
MiniMayoSRS~\citesupp{pedersen2007measures} & {\tt minimayosrs} & train & 803 & 92 & 29 & CC0 1.0 & STS & PAIRS & EN & Public \\
\midrule
\multirow{2}{2.1cm}{miRNA~\citesupp{Bagewadi2014}} & \multirow{2}{2.5cm}{{\tt mirna}} & train & 272k & 38.2k & 201 & \multirow{2}{1.1cm}{CC BY NC 3.0} & \multirow{2}{1.4cm}{NER, NED} & \multirow{2}{0.75cm}{KB} & \multirow{2}{0.25cm}{EN} & \multirow{2}{0.5cm}{Public} \\
  &   & test & 115k & 16k & 100 &   &   &   &   &   \\
\midrule
\multirow{3}{2.1cm}{MLEE~\citesupp{pyysalo2012event}} & \multirow{3}{2.5cm}{{\tt mlee}} & train & 199k & 27.9k & 131 & \multirow{3}{1.1cm}{CC BY NC SA 3.0} & \multirow{3}{1.4cm}{RE, EE, NER, COREF} & \multirow{3}{0.75cm}{KB} & \multirow{3}{0.25cm}{EN} & \multirow{3}{0.5cm}{Public} \\
  &   & valid & 68.1k & 9.61k & 44 &   &   &   &   &   \\
  &   & test & 135k & 19.1k & 87 &   &   &   &   &   \\
\midrule
MQP~\citesupp{DBLP:journals/biodb/LiSJSWLDMWL16} & {\tt mqp} & train & 644k & 120k & 3048 & \emph{Unknown} & STS & PAIRS & EN & Public \\
\midrule
MSH WSD~\citesupp{jimeno2011exploiting} & {\tt msh\_wsd} & train & 52.8M & 7.59M & 37888 & UMLS & NED & KB & EN & DUA \\
\midrule
\multirow{2}{2.1cm}{MuchMore~\citesupp{buitelaar2003multi}} & \multirow{2}{2.5cm}{{\tt muchmore}} & train & 8.43M & 1.11M & 7808 & \multirow{2}{1.1cm}{\emph{Unknown}} & \multirow{2}{1.4cm}{NER} & \multirow{2}{0.75cm}{KB} & \multirow{2}{0.25cm}{EN, DE} & \multirow{2}{0.5cm}{Public} \\
  &   & train & 12.7M & 1.69M & 6374 &   &   &   &   &   \\
\midrule
\multirow{3}{2.1cm}{Multi-XScience~\citesupp{https://doi.org/10.48550/arxiv.2010.14235}} & \multirow{3}{2.5cm}{{\tt multi\_xscience}} & train & 143M & 21.3M & 30369 & \multirow{3}{1.1cm}{MIT} & \multirow{3}{1.4cm}{SUM, PARA} & \multirow{3}{0.75cm}{T2T} & \multirow{3}{0.25cm}{EN} & \multirow{3}{0.5cm}{Public} \\
  &   & valid & 23.9M & 3.54M & 5066 &   &   &   &   &   \\
  &   & test & 23.6M & 3.51M & 5093 &   &   &   &   &   \\
\midrule
\multirow{2}{2.1cm}{MutationFinder~\citesupp{Caporaso2007}} & \multirow{2}{2.5cm}{{\tt mutation\_finder}} & valid & 416k & 61.4k & 305 & \multirow{2}{1.1cm}{Custom} & \multirow{2}{1.4cm}{NER} & \multirow{2}{0.75cm}{KB} & \multirow{2}{0.25cm}{EN} & \multirow{2}{0.5cm}{Public} \\
  &   & test & 726k & 107k & 508 &   &   &   &   &   \\
\midrule
\multirow{2}{2.1cm}{n2c2 2006 De-identification~\citesupp{uzuner2007evaluating}} & \multirow{2}{2.5cm}{{\tt n2c2\_2006\_deid}} & train & 2.25M & 340k & 669 & \multirow{2}{1.1cm}{DUA} & \multirow{2}{1.4cm}{NER} & \multirow{2}{0.75cm}{KB} & \multirow{2}{0.25cm}{EN} & \multirow{2}{0.5cm}{DUA} \\
  &   & test & 952k & 146k & 220 &   &   &   &   &   \\
\midrule
\multirow{2}{2.1cm}{n2c2 2006 Smoking Status~\citesupp{uzuner2008identifying}} & \multirow{2}{2.5cm}{{\tt n2c2\_2006\_smokers}} & train & 1.72M & 304k & 398 & \multirow{2}{1.1cm}{DUA} & \multirow{2}{1.4cm}{TXTCLASS} & \multirow{2}{0.75cm}{TEXT} & \multirow{2}{0.25cm}{EN} & \multirow{2}{0.5cm}{DUA} \\
  &   & test & 479k & 85.1k & 104 &   &   &   &   &   \\
\midrule
\multirow{2}{2.1cm}{n2c2 2008 \ Obesity~\citesupp{uzuner2009recognizing}} & \multirow{2}{2.5cm}{{\tt n2c2\_2008}} & train & 5M & 852k & 730 & \multirow{2}{1.1cm}{DUA} & \multirow{2}{1.4cm}{TXTCLASS} & \multirow{2}{0.75cm}{TEXT} & \multirow{2}{0.25cm}{EN} & \multirow{2}{0.5cm}{DUA} \\
  &   & test & 3.5M & 595k & 507 &   &   &   &   &   \\
\midrule
\multirow{2}{2.1cm}{n2c2 2009 Medication~\citesupp{DBLP:journals/jamia/UzunerSC10}} & \multirow{2}{2.5cm}{{\tt n2c2\_2009}} & train & 4.86M & 824k & 696 & \multirow{2}{1.1cm}{DUA} & \multirow{2}{1.4cm}{NER} & \multirow{2}{0.75cm}{KB} & \multirow{2}{0.25cm}{EN} & \multirow{2}{0.5cm}{DUA} \\
  &   & test & 3.75M & 637k & 553 &   &   &   &   &   \\
\midrule
\multirow{2}{2.1cm}{n2c2 2010 Relations~\citesupp{DBLP:journals/jamia/UzunerSSD11}} & \multirow{2}{2.5cm}{{\tt n2c2\_2010}} & train & 827k & 150k & 170 & \multirow{2}{1.1cm}{DUA} & \multirow{2}{1.4cm}{RE, NER} & \multirow{2}{0.75cm}{KB} & \multirow{2}{0.25cm}{EN} & \multirow{2}{0.5cm}{DUA} \\
  &   & test & 1.48M & 267k & 256 &   &   &   &   &   \\
\midrule
\multirow{2}{2.1cm}{n2c2 2011 Coreference~\citesupp{uzuner2012evaluating}} & \multirow{2}{2.5cm}{{\tt n2c2\_2011}} & train & 1.37M & 247k & 251 & \multirow{2}{1.1cm}{DUA} & \multirow{2}{1.4cm}{COREF} & \multirow{2}{0.75cm}{KB} & \multirow{2}{0.25cm}{EN} & \multirow{2}{0.5cm}{DUA} \\
  &   & test & 916k & 167k & 173 &   &   &   &   &   \\
\midrule
\multirow{2}{2.1cm}{n2c2 2014 De-identification~\citesupp{stubbs2015automated}} & \multirow{2}{2.5cm}{{\tt n2c2\_2014\_deid}} & train & 3.4M & 489k & 790 & \multirow{2}{1.1cm}{DUA} & \multirow{2}{1.4cm}{NER} & \multirow{2}{0.75cm}{KB} & \multirow{2}{0.25cm}{EN} & \multirow{2}{0.5cm}{DUA} \\
  &   & test & 2.19M & 316k & 514 &   &   &   &   &   \\
\midrule
\multirow{2}{2.1cm}{n2c2 2014 Cardiac Risk Factors~\citesupp{kUMAR2015S6}} & \multirow{2}{2.5cm}{{\tt n2c2\_2014\_risk\_factors}} & train & 3.4M & 489k & 790 & \multirow{2}{1.1cm}{DUA} & \multirow{2}{1.4cm}{TXTCLASS} & \multirow{2}{0.75cm}{TEXT} & \multirow{2}{0.25cm}{EN} & \multirow{2}{0.5cm}{DUA} \\
  &   & test & 2.19M & 316k & 514 &   &   &   &   &   \\
\midrule
\multirow{2}{2.1cm}{n2c2 2018 Selection Criteria~\citesupp{DBLP:journals/jamia/StubbsFSHU19}} & \multirow{2}{2.5cm}{{\tt n2c2\_2018\_track1}} & train & 3.91M & 550k & 202 & \multirow{2}{1.1cm}{DUA} & \multirow{2}{1.4cm}{TXTCLASS} & \multirow{2}{0.75cm}{TEXT} & \multirow{2}{0.25cm}{EN} & \multirow{2}{0.5cm}{DUA} \\
  &   & test & 1.64M & 231k & 86 &   &   &   &   &   \\
\midrule
\multirow{2}{2.1cm}{n2c2 2018 ADE~\citesupp{DBLP:journals/jamia/HenryBFSU20}} & \multirow{2}{2.5cm}{{\tt n2c2\_2018\_track2}} & train & 3.84M & 574k & 303 & \multirow{2}{1.1cm}{DUA} & \multirow{2}{1.4cm}{RE, NER} & \multirow{2}{0.75cm}{KB} & \multirow{2}{0.25cm}{EN} & \multirow{2}{0.5cm}{DUA} \\
  &   & test & 2.54M & 377k & 202 &   &   &   &   &   \\
\midrule
\multirow{3}{2.1cm}{NCBI Disease~\citesupp{Dogan2014NCBIDC}} & \multirow{3}{2.5cm}{{\tt ncbi\_disease}} & train & 747k & 113k & 592 & \multirow{3}{1.1cm}{CC0 1.0} & \multirow{3}{1.4cm}{NER, NED} & \multirow{3}{0.75cm}{KB} & \multirow{3}{0.25cm}{EN} & \multirow{3}{0.5cm}{Public} \\
  &   & valid & 133k & 20.1k & 100 &   &   &   &   &   \\
  &   & test & 135k & 20.4k & 100 &   &   &   &   &   \\
\midrule
\multirow{2}{2.1cm}{NLM-Gene~\citesupp{islamaj2021nlm}} & \multirow{2}{2.5cm}{{\tt nlm\_gene}} & train & 812k & 114k & 450 & \multirow{2}{1.1cm}{CC0 1.0} & \multirow{2}{1.4cm}{NER, NED} & \multirow{2}{0.75cm}{KB} & \multirow{2}{0.25cm}{EN} & \multirow{2}{0.5cm}{Public} \\
  &   & test & 180k & 25.2k & 100 &   &   &   &   &   \\
\midrule
NLM WSD~\citesupp{weeber2001developing} & {\tt nlm\_wsd} & train & 8.37M & 1.22M & 5000 & UMLS & NED & KB & EN & DUA \\
\midrule
\multirow{3}{2.1cm}{NLM-Chem~\citesupp{islamaj2021nlm}} & \multirow{3}{2.5cm}{{\tt nlmchem}} & train & 2.69M & 408k & 80 & \multirow{3}{1.1cm}{CC0 1.0} & \multirow{3}{1.4cm}{NER, NED, TXTCLASS} & \multirow{3}{0.75cm}{KB, TEXT} & \multirow{3}{0.25cm}{EN} & \multirow{3}{0.5cm}{Public} \\
  &   & valid & 663k & 100k & 20 &   &   &   &   &   \\
  &   & test & 1.52M & 229k & 50 &   &   &   &   &   \\
\midrule
\multirow{4}{2.1cm}{NTCIR-13 MedWeb~\citesupp{wakamiya2017overview}} & \multirow{4}{2.5cm}{{\tt ntcir\_13\_medweb}} & train & 79.4M & 3.71M & 1920 & \multirow{4}{1.1cm}{CC BY 4.0} & \multirow{4}{1.4cm}{TXTCLASS} & \multirow{4}{0.75cm}{TEXT} & \multirow{4}{0.25cm}{EN, ZH, JA} & \multirow{4}{0.5cm}{DUA} \\
  &   & test & 8.38M & 412k & 640 &   &   &   &   &   \\
  &   & train & 163k & 27.2k & 1920 &   &   &   &   &   \\
  &   & test & 50.7k & 8.47k & 640 &   &   &   &   &   \\
\midrule
OSIRIS~\citesupp{Furlong2008} & {\tt osiris} & train & 172k & 25.7k & 105 & CC BY 3.0 & NER, NED & KB & EN & Public \\
\midrule
\multirow{3}{2.1cm}{ParaMed~\citesupp{liu2021paramed}} & \multirow{3}{2.5cm}{{\tt paramed}} & train & 16.2M & 3.74M & 62127 & \multirow{3}{1.1cm}{CC BY 4.0} & \multirow{3}{1.4cm}{TRANSL} & \multirow{3}{0.75cm}{T2T} & \multirow{3}{0.25cm}{EN, ZH} & \multirow{3}{0.5cm}{Public} \\
  &   & valid & 552k & 128k & 2036 &   &   &   &   &   \\
  &   & test & 564k & 130k & 2102 &   &   &   &   &   \\
\midrule
PDR~\citesupp{kim2019corpus} & {\tt pdr} & train & 274k & 40.5k & 179 & \emph{Unknown} & EE, NER, COREF & KB & EN & Public \\
\midrule
\multirow{3}{2.1cm}{PharmaCoNER~\citesupp{gonzalez2019pharmaconer}} & \multirow{3}{2.5cm}{{\tt pharmaconer}} & train & 1.18M & 177k & 500 & \multirow{3}{1.1cm}{CC BY 4.0} & \multirow{3}{1.4cm}{NER, TXTCLASS} & \multirow{3}{0.75cm}{KB, TEXT} & \multirow{3}{0.25cm}{ES} & \multirow{3}{0.5cm}{Public} \\
  &   & valid & 567k & 85.1k & 250 &   &   &   &   &   \\
  &   & test & 587k & 88.2k & 250 &   &   &   &   &   \\
\midrule
\multirow{3}{2.1cm}{PhoNER\_COVID19~\citesupp{PhoNER_COVID19}} & \multirow{3}{2.5cm}{{\tt pho\_ner}} & train & 671k & 168k & 5027 & \multirow{3}{1.1cm}{Custom} & \multirow{3}{1.4cm}{NER} & \multirow{3}{0.75cm}{KB} & \multirow{3}{0.25cm}{VI} & \multirow{3}{0.5cm}{Public} \\
  &   & valid & 286k & 71.3k & 2000 &   &   &   &   &   \\
  &   & test & 433k & 108k & 3000 &   &   &   &   &   \\
\midrule
PICO Annotation~\citesupp{zlabinger-etal-2020-effective} & {\tt pico\_extraction} & train & 60.4k & 10.2k & 421 & \emph{Unknown} & NER & KB & EN & Public \\
\midrule
\multirow{3}{2.1cm}{PMC-Patients~\citesupp{zhao2022pmcpatients}} & \multirow{3}{2.5cm}{{\tt pmc\_patients}} & train & 1.22B & 184M & 257366 & \multirow{3}{1.1cm}{CC BY NC SA 4.0} & \multirow{3}{1.4cm}{STS} & \multirow{3}{0.75cm}{PAIRS} & \multirow{3}{0.25cm}{EN} & \multirow{3}{0.5cm}{Public} \\
  &   & valid & 6.72M & 1.02M & 2144 &   &   &   &   &   \\
  &   & test & 7.67M & 1.17M & 2366 &   &   &   &   &   \\
\midrule
\multirow{6}{2.1cm}{ProGene~\citesupp{faessler-etal-2020-progene}} & \multirow{6}{2.5cm}{{\tt progene}} & split k=10 & 821k & 4.76M & 30926 & \multirow{6}{1.1cm}{CC BY 4.0} & \multirow{6}{1.4cm}{NER} & \multirow{6}{0.75cm}{KB} & \multirow{6}{0.25cm}{EN} & \multirow{6}{0.5cm}{Public} \\
  &   & split k=10 & 43.3k & 251k & 1676 &   &   &   &   &   \\
  &   & split k=10 & 96.1k & 557k & 3623 &   &   &   &   &   \\
\midrule
\multirow{2}{2.1cm}{PsyTAR~\citesupp{Zolnoori2019}} & \multirow{2}{2.5cm}{{\tt psytar}} & train & 319k & 56.4k & 3398 & \multirow{2}{1.1cm}{CC BY 4.0} & \multirow{2}{1.4cm}{NER} & \multirow{2}{0.75cm}{KB} & \multirow{2}{0.25cm}{EN} & \multirow{2}{0.5cm}{DUA} \\
  &   & train & 57k & 7.56k & 6003 &   &   &   &   &   \\
\midrule
\multirow{3}{2.1cm}{PUBHEALTH~\citesupp{kotonya2020explainable}} & \multirow{3}{2.5cm}{{\tt pubhealth}} & train & 5.61M & 899k & 9804 & \multirow{3}{1.1cm}{MIT} & \multirow{3}{1.4cm}{TXTCLASS} & \multirow{3}{0.75cm}{PAIRS} & \multirow{3}{0.25cm}{EN} & \multirow{3}{0.5cm}{Public} \\
  &   & valid & 683k & 110k & 1223 &   &   &   &   &   \\
  &   & test & 692k & 111k & 1231 &   &   &   &   &   \\
\midrule
\multirow{3}{2.1cm}{PubMedQA~\citesupp{jin2019pubmedqa}} & \multirow{3}{2.5cm}{{\tt pubmed\_qa}} & train & 1.28M & 549k & 450 & \multirow{3}{1.1cm}{MIT} & \multirow{3}{1.4cm}{QA} & \multirow{3}{0.75cm}{QA} & \multirow{3}{0.25cm}{EN} & \multirow{3}{0.5cm}{Public} \\
  &   & valid & 141k & 60.3k & 50 &   &   &   &   &   \\
  &   & test & 1.45M & 618k & 500 &   &   &   &   &   \\
\midrule
PubTator Central~\citesupp{10.1093/nar/gkz389} & {\tt pubtator\_central} & train & 19.5k & 2.91k & 4 & NCBI & NER, NED & KB & EN & Public \\
\midrule
\multirow{3}{2.1cm}{QUAERO~\citesupp{neveol14quaero}} & \multirow{3}{2.5cm}{{\tt quaero}} & train & 67.7k & 10.6k & 833 & \multirow{3}{1.1cm}{GFDL 1.3} & \multirow{3}{1.4cm}{NER} & \multirow{3}{0.75cm}{KB} & \multirow{3}{0.25cm}{FR} & \multirow{3}{0.5cm}{Public} \\
  &   & valid & 68.2k & 10.5k & 832 &   &   &   &   &   \\
  &   & test & 70k & 10.9k & 832 &   &   &   &   &   \\
\midrule
SCAI Chemical~\citesupp{kolarik:lrec-ws08} & {\tt scai\_chemical} & train & 155k & 20.9k & 100 & \emph{Unknown} & NER & KB & EN & Public \\
\midrule
SCAI Disease~\citesupp{gurulingappa:lrec-ws10} & {\tt scai\_disease} & train & 630k & 90.4k & 400 & \emph{Unknown} & NER & KB & EN & Public \\
\midrule
\multirow{3}{2.1cm}{SciCite~\citesupp{cohan:naacl19}} & \multirow{3}{2.5cm}{{\tt scicite}} & train & 1.82M & 280k & 8243 & \multirow{3}{1.1cm}{\emph{Unknown}} & \multirow{3}{1.4cm}{TXTCLASS} & \multirow{3}{0.75cm}{TEXT} & \multirow{3}{0.25cm}{EN} & \multirow{3}{0.5cm}{Public} \\
  &   & valid & 203k & 31.3k & 916 &   &   &   &   &   \\
  &   & test & 413k & 63.4k & 1861 &   &   &   &   &   \\
\midrule
SciELO~\citesupp{soares2018large} & {\tt scielo} & train & 995M & 153M & 2828917 & CC BY 4.0 & TRANSL & T2T & EN, ES, PT & Public \\
\midrule
\multirow{3}{2.1cm}{SciFact~\citesupp{wadden2020fact}} & \multirow{3}{2.5cm}{{\tt scifact}} & train & 787k & 112k & 919 & \multirow{3}{1.1cm}{CC BY NC 2.0} & \multirow{3}{1.4cm}{TE} & \multirow{3}{0.75cm}{TE} & \multirow{3}{0.25cm}{EN} & \multirow{3}{0.5cm}{Public} \\
  &   & valid & 280k & 39.6k & 339 &   &   &   &   &   \\
  &   & test & 26.4k & 3.62k & 300 &   &   &   &   &   \\
\midrule
\multirow{3}{2.1cm}{SciQ~\citesupp{welbl-etal-2017-crowdsourcing}} & \multirow{3}{2.5cm}{{\tt sciq}} & train & 11.8M & 4.96M & 11679 & \multirow{3}{1.1cm}{CC BY NC 3.0} & \multirow{3}{1.4cm}{QA} & \multirow{3}{0.75cm}{QA} & \multirow{3}{0.25cm}{EN} & \multirow{3}{0.5cm}{Public} \\
  &   & valid & 993k & 418k & 1000 &   &   &   &   &   \\
  &   & test & 1.02M & 428k & 1000 &   &   &   &   &   \\
\midrule
\multirow{3}{2.1cm}{SciTail~\citesupp{scitail}} & \multirow{3}{2.5cm}{{\tt scitail}} & train & 4.19M & 681k & 23596 & \multirow{3}{1.1cm}{Apache 2.0} & \multirow{3}{1.4cm}{TE} & \multirow{3}{0.75cm}{TE} & \multirow{3}{0.25cm}{EN} & \multirow{3}{0.5cm}{Public} \\
  &   & valid & 237k & 38.8k & 1304 &   &   &   &   &   \\
  &   & test & 372k & 62.3k & 2126 &   &   &   &   &   \\
\midrule
SETH Corpus~\citesupp{SETH2016} & {\tt seth\_corpus} & train & 760k & 111k & 630 & Apache 2.0 & RE, NER & KB & EN & Public \\
\midrule
SPL ADR~\citesupp{demner2018dataset} & {\tt spl\_adr\_200db} & train & 29M & 3.46M & 2208 & CC0 1.0 & RE, NER, NED & KB & EN & Public \\
\midrule
Swedish Medical NER~\citesupp{almgren-etal-2016-named} & {\tt swedish\_medical\_ner} & train & 85k & 14.1k & 926 & CC BY SA 4.0 & NER & KB & SV & Public \\
\midrule
SNP Corpus~\citesupp{thomas2011challenges} & {\tt thomas2011} & test & 0 & 0 & 296 & Custom & NER, NED & KB & EN & Public \\
\midrule
\multirow{2}{2.1cm}{tmVar v1~\citesupp{wei2013tmvar}} & \multirow{2}{2.5cm}{{\tt tmvar\_v1}} & train & 547k & 80.2k & 334 & \multirow{2}{1.1cm}{\emph{Unknown}} & \multirow{2}{1.4cm}{NER} & \multirow{2}{0.75cm}{KB} & \multirow{2}{0.25cm}{EN} & \multirow{2}{0.5cm}{Public} \\
  &   & test & 265k & 38.8k & 166 &   &   &   &   &   \\
\midrule
tmVar v2~\citesupp{wei2018tmvar} & {\tt tmvar\_v2} & train & 259k & 38k & 158 & \emph{Unknown} & NER, NED & KB & EN & Public \\
\midrule
tmVar v3~\citesupp{https://doi.org/10.48550/arxiv.2204.03637} & {\tt tmvar\_v3} & test & 812k & 119k & 500 & \emph{Unknown} & NER, NED & KB & EN & Public \\
\midrule
\multirow{3}{2.1cm}{TwADR-L~\citesupp{limsopatham-collier-2016-normalising}} & \multirow{3}{2.5cm}{{\tt twadrl}} & train k=10 & 10k & 76.1k & 4805 & \multirow{3}{1.1cm}{CC BY 4.0} & \multirow{3}{1.4cm}{NER, NED} & \multirow{3}{0.75cm}{KB} & \multirow{3}{0.25cm}{EN} & \multirow{3}{0.5cm}{Public} \\
  &   & validation k=10 & 327 & 1.84k & 125 &   &   &   &   &   \\
  &   & test k=10 & 361 & 2.03k & 142 &   &   &   &   &   \\
\midrule
UMNSRS~\citesupp{pakhomov2010semantic} & {\tt umnsrs} & train & 11.3k & 1.2k & 587 & CC0 1.0 & STS & PAIRS & EN & Public \\
\midrule
Verspoor 2013~\citesupp{verspoor2013annotating} & {\tt verspoor\_2013} & train & 279k & 42.9k & 120 & \emph{Unknown} & RE, NER & KB & EN & Public \\
\bottomrule
\end{longtable}
\end{scriptsize}